\documentclass{article}
\pdfoutput=1
\usepackage[preprint]{nips_2018}

\usepackage{graphicx}
\usepackage{url}
\usepackage[utf8]{inputenc}
\usepackage[T1]{fontenc}
\usepackage[english]{babel}
\usepackage{xcolor}
\usepackage{booktabs}
\usepackage{makecell}
\usepackage{appendix}
\usepackage{amsmath}
\usepackage{amssymb}
\usepackage{subfig}
\usepackage{multirow}
\usepackage{indentfirst}

\usepackage{tikz}
\usetikzlibrary{decorations.pathmorphing}


\begin{document}

\title{Online Multi-target regression trees with stacked leaf models}


\author{Saulo Martiello Mastelini\\
        Institute of Mathematics and Computer Sciences\\
        University of São Paulo\\
        São Carlos, BR 13566-590\\
        \texttt{mastelini@usp.br}\\
        \And
        Sylvio Barbon Jr.\\
        Computer Science Department\\
        State University of Londrina\\
        Londrina, BR 86057-970\\
        \texttt{barbon@uel.br}\\
        \AND
        André Carlos Ponce de Leon Ferreira de Carvalho\\
        Institute of Mathematics and Computer Sciences\\
        University of São Paulo\\
        São Carlos, BR 13566-590\\
        \texttt{andre@icmc.usp.br}
}





\maketitle

\begin{abstract}

One of the current challenges in machine learning is how to deal with data coming at increasing rates in data streams.
New predictive learning strategies are needed to cope with the high throughput data and concept drift.
One of the data stream mining tasks where new learning strategies are needed is multi-target regression, due to its applicability in a high number of real world problems.
While reliable and effective learning strategies have been proposed for batch multi-target regression, few have been proposed for multi-target online learning in data streams. 
Besides, most of the existing solutions do not consider the occurrence of inter-target correlations when making predictions.
In this work, we propose a novel online learning strategy for multi-target regression in data streams.
The proposed strategy extends existing online decision tree learning algorithm to explore inter-target dependencies while making predictions.
For such, the proposed strategy, called Stacked Single-target Hoeffding Tree (SST-HT), uses the inter-target dependencies as an additional information source to enhance predictive accuracy. Throughout an extensive experimental setup, we evaluate our proposal against state-of-the-art decision tree-based algorithms for online multi-target regression.
According to the experimental results, SST-HT presents superior predictive accuracy, with a small increase in the processing time and memory requirements.

\end{abstract}

\section{Introduction}
\label{sec_introduction}

Recent advances of computing technologies have increased the amount of data being produced, resulting in data streams of potentially unbounded size. These advances also boosted the speed in which computers process and exchange data. While previous generations of machine learning (ML) algorithms were concerned with processing relatively small amounts of data (in batches), without time restrictions, the new challenges brought by big data changed the needs and shifted the research efforts to other directions. 

As data can arrive fast and in large volumes, the data stream algorithms must be able to process each incoming example just once (since all data cannot be indefinitely stored)~\citep{read2012scalable}. Moreover, the induced models must be ready to predict new cases at any point and expect an infinite data stream (despite using finite and limited resources regarding time and memory)~\citep{read2012scalable,kocev2013tree,sousa2018multilabel}. 

The continuous data flow may present novel characteristics and bring new challenges for which traditional ML algorithms were not designed to deal with. They include concept drift (CD), novelty detection, among other aspects~\citep{gama2010knowledge,krempl2014open}.
Nonetheless, in this work we focus on stationary streams.
Besides, data streams are varied and can come from many sources, ranging from sensor networks and manufacturing processes to video streams and user operations in a web browser~\cite{MOA-Book-2018}.
Data streams impose new requirements to ML algorithms, such as fast and incremental learning, robustness to noise, and low memory need.

In the data stream mining literature, most of the efforts have been devoted to dealing with single-target (ST) tasks, mostly for classification~\citep{gama2010knowledge,nguyen2015survey,gomes2017survey,krawczyk2017ensemble}.
A small number of studies addresses other tasks, for instance single-target regression (STR)~\citep{ikonomovska2011learning,duarte2016adaptive,gouk2019}. Moreover, little attention has been given to structured output tasks~\citep{kocev2013tree,borchani2015survey,waegeman2018}, i.e., when multiple target variables are related to the same set of input features.
Notwithstanding, this type of prediction tasks reflects many aspects of real-world problems, including several problems associated with data streams, such as predicting river flow properties, multiple product sales and airline ticket prices~\citep{read2012scalable,spyromitros2016multi,sousa2018multilabel}.
In this work, we focus on multi-target regression (MTR) tasks, which are concerned, as the name implies, with the simultaneous prediction of multiple continuous target values.
These targets can be correlated, since they are explained by the same set of predictive feature values or represent correlated quantities in real-world problems. As a consequence, this information can be used by ML algorithms to improve the overall predictive performance.

ML research in MTR is a relatively new research area, even for batch data mining problems~\citep{borchani2015survey,spyromitros2016multi,mastelini2017dstars,melki2017multi,mastelini2018multi,santana2018predicting}.
There is still room for improvement in the few existing online solutions~\citep{ikonomovska2011incremental,duarte2015multi,osojnik2018tree}, e.g., by improving the way in which inter-target dependencies are explored.
In fact, some of the ideas implemented for batch data can be adapted to online scenarios, without largely impacting the necessary computational resources. However, as online MTR algorithms must cope with the requirements of learning from data streams, the search for a balance between performance and feasible MTR solutions is essential.

We propose a new MTR algorithm, called Stacked Single-Target Hoeffding Tree (SST-HT), which extends existing incremental decision tree induction algorithms.
SST-HT combines the simultaneous prediction of multiple targets to explore their inter-dependencies.
SST-HT follows the common trend in batch MTR literature of using stacked single-target (SST) regression models to improve predictive performance~\citep{spyromitros2016multi,mastelini2017dstars,mastelini2018multi,melki2017multi,santana2018predicting}.
Nevertheless, 
SST-HT does not change the way the decision trees are built, i.e., how the decision splits are performed, nor it highly impacts the required computational resources.
Therefore, we expect the same tree structure and improved predictive performance in comparison with traditional tree-based algorithms for MTR in data streams. Two variations of SST-HT are evaluated: the first only uses the stacked regressors as predictors, and the second dynamically selects between stacked regressors, the target mean, or linear predictors.

Experimental results show the superior predictive performance of SST-HT, when compared with variations of the iSOUP-Tree algorithm in sixteen MTR datasets.
To the best of our knowledge, this is the highest number of datasets used so far as benchmarks for MTR in data streams.
Six of the evaluated datasets were first employed as MTR resources in this research.
Among these new datasets, SCFP was specially tailored for this work by adapting an existing dataset and adding textual (provided by a word embedding model~\citep{pennington2014glove}) and geolocated information (see Appendix~\ref{sec_appendix_datasets} for more details).
The remaining newly introduced datasets are derived from well-known and publicly available real-world data, commonly used in other data analysis tasks.
In the experiments carried out for this study,
SST-HT obtained the best predictive performance, while adding small extra memory consumption, and a processing time linearly comparable to the other algorithms.

The remainder of this work is organized as follows. Section~\ref{sec_background_related} presents a brief background on MTR solutions for data streams, as well as a literature review on the subject. Section~\ref{sec_mtr_trees} provides the theoretical foundation of traditional Hoeffding Tree (HT) algorithms, the basis of SST-HT and related algorithms. SST-HT is described in detail in Section~\ref{sec_sst_ht}. Section~\ref{sec_experimental_setup} presents our experimental setup, including the datasets, evaluation strategy, and metrics, as well as the configurations used for the decision tree algorithms. The obtained results are discussed in Section~\ref{sec_results}, and our final considerations presented in Section~\ref{sec_conclusion}. Finally, detailed information concerning the datasets used and the obtained results are presented in Appendices~\ref{sec_appendix_datasets} and \ref{sec_appendix_line_plots}. 
\section{Background and Related Work}\label{sec_background_related}

MTR deals with the prediction of multiple continuous target variables, using the same set of predictive variables.
This task can be seen as an extension of STR tasks~\citep{borchani2015survey, osojnik2018tree}.
Nevertheless, MTR aims not only at modelling the input to output relations, but also possible inter-output dependencies.
This can improve the representation of the problem to be solved, and, as a result, the predictive performance.
On the other hand, this additional effort demands solutions specially tailored for MTR tasks, which are often more complex than using a separate STR model for each target variable.

Formally, a MTR task can be described as the search for a function $f$, able to model the relation between a set $\mathbf{X}$ of $m$ input variables (real, ordinal or nominal values) and a set $\mathbf{Y} \subset \mathbb{R}^d$ of $d$ output variables. Therefore, a MTR task can be represented by the expression

$$f\colon \mathbf{X}\to \mathbf{Y}.$$

Function $f$ can be used to predict $\hat{\mathbf{y}}$ for an instance $\mathbf{x} \in \mathbf{X}$. When inducing $f$, it is expected that the predicted values, $\hat{\mathbf{y}}$, are as close as possible to the true values, $\mathbf{y}$.

According to~\cite{kocev2013tree}, MTR algorithms follow two main approaches: global and local.
Global algorithms use a single model to predict all target variables at once. These algorithms implicitly model the inter-target dependencies, and offer more compact and less computationally costly solutions, which are more suited to online scenarios.
Local algorithms combine traditional STR solutions and often manipulate or modify the input space to insert inter-target dependency information within the modelling process.
Thus, local algorithms use multiple ST regressors to solve an MTR task, often more than one regressor for each target variable~\citep{spyromitros2016multi,mastelini2017dstars,santana2018predicting,mastelini2018multi}.
As a result, they have a higher cost than global algorithms. 
The simplest local solution for MTR tasks, as previously mentioned, is the induction of an STR for each target,
overlooking inter-target dependencies.
In this work we apply local-based techniques within a global tree-based MTR algorithm.

MTR have been widely used in batch learning applications~\citep{borchani2015survey,spyromitros2016multi,mastelini2017dstars,melki2017multi,mastelini2018multi,santana2018predicting}, since they are related to several real-life problems, such as prediction of river flow properties, online sales, airline ticket prices and poultry meat properties~\citep{spyromitros2016multi,santana2018predicting}. 
However, there are few papers investigating the use of MTR (and even STR) problems in online learning tasks.
Applications of MTR in data streams not only have the same computational constraints as in online ST classification applications, but they also bring the additional challenge of simultaneously producing multiple predictions~\citep{borchani2015survey,waegeman2018}.

One of the key works where regression algorithms were used for data stream mining was~\cite{ikonomovska2011learning}.
In this paper, the authors proposed an online and incremental algorithm to induce regression trees in the presence of CD.
The proposed algorithm, called FIMT-DD (Fast Incremental Model Tree with Drift Detection), adopts the Hoeffding's bound theorem to decide whether a split decision must be made~\citep{domingos2000mining,gama2010knowledge}.
The main aspects of their algorithm are very similar to VFDT (Very Fast Decision Tree)~\citep{domingos2000mining}.
FIMT-DD uses perceptrons with linear activation function at the tree's leaves to provide the responses.
As it is one of the first works dealing with regression problems on data streams, the authors mostly evaluated their approach against traditional batch regression algorithms.
Notwithstanding, their research pioneered the research on STR and MTR for data streams.
However, FIMT-DD was designed to deal only with numerical attributes.
This limits its application to numerical data only, unless some data transformation technique, e.g., one hot encoding, is used.

The same authors also proposed the FIMT-MT (Fast Incremental Model Tree - Multi-target)~\citep{ikonomovska2011incremental}, an extension of the FIMT-DD algorithm for MTR settings.
This proposal, a global approach, uses aspects of predictive clustering trees~\citep{kocev2013tree} to make decision splits on multiple targets.  FIMT-MT considers each split as the induction of a cluster.
Thus, the root node corresponds to the cluster that contains all the data.
Each new split tries to reduce the intra-cluster variance of the new created partitions, while maximizing the inter-cluster variance. Similarly to the FIMT-DD algorithm, FIMT-MT has perceptrons in its leaves, one model per target. On the other hand, no mechanism for dealing with CD is inherited from the original STR algorithm.
FIMT-MT also supports only numerical attributes.

Deviating from tree-based strategies, \cite{almeida2013adaptive} proposed the Adaptive Model Rules (AMRules) algorithm for online STR tasks.
This algorithm was expanded by \cite{duarte2016adaptive}.
AMRules also uses linear perceptrons as the consequent of the rules.
Moreover, it employs a built-in mechanism for dealing with CD based on the Page-Hinkley (PH) test~\citep{ikonomovska2011learning}.
When detecting a CD, AMRules simply drops outdated decision rules.
In addition, the decision rule algorithm also has a routine to detect anomalous examples, e.g., noisy data.
These examples are not used to update the decision models.

\cite{duarte2015multi} also expanded the AMRules framework to allow their use in MTR tasks. This expanded version is also based on the principle that the created partitions must reduce the variance in the output space.
However, different from the previous solution, AMRules-MTR does not lie in the global/local categorization, since it can specialize in subsets of targets.
When executing a rule expansion test, if the variance in the target space is reduced only for some targets, a new decision rule encompassing the targets benefited by the split is created.
In a complementary manner, a rule without the expansion is also created for the remaining targets. Therefore, AMRules-MTR creates decision rules which can encompass all the targets, some of them, or even a single target.
Hence this algorithm should be characterized as a hybrid of a local and global approach.
More recently, AMRules was also adapted to deal with multi-label classification tasks~\citep{sousa2018multilabel}.
Tree-based solutions do not have similar mechanism to explored inter-target dependencies.

Following the trend of applying tree-based algorithms to data streams, \citep{osojnik2015comparison,osojnik2018tree} proposed an extension for the FIMT-MT algorithm, called iSOUP-Tree (incremental Structured Output Prediction Tree).
This algorithm builds upon the research of \cite{ikonomovska2011incremental} by adding support to categorical features and using an adaptive prediction model in the leaves.
Instead of using only perceptrons, iSOUP-Tree also maintains a mean predictor for each target.
Besides, it selects the best current model by monitoring a faded error metric for each model.
The authors adapted the iSOUP-Tree algorithm for multiple settings, including ensembles (Bagging and Random Forest) and Option Trees~\citep{osojnik2018tree}. Moreover, they investigated the application of all the variations of the MTR algorithm to multi-label classification tasks~\citep{osojnik2015multilabel,osojnik2017multilabel}.

Neither of the previous tree-based solutions effectively take advantage of inter-target dependencies when making predictions.
In all of them, individual models are created for each target.
Thus, they ignore how the targets relate to each other.
Inspired by \cite{spyromitros2016multi}, we propose the Single-target Hoeffding Tree (SST-HT) algorithm, which is based on iSOUP-Tree and use Stacked Single-target (SST) predictors in the tree's leaves.
SST-HT can deal with the mutable characteristics of streaming tasks by automatically selecting the best current predictor for each target, i.e., whether to use SST, the standard perceptron, or the most straightforward mean predictor.
Since SST-HT is based on iSOUP-Tree algorithm, and, consequently FIMT-MT, we will first present the base algorithm for building incremental MTR decision trees and later describe how SST-HT works. 
\section{Online Multi-target regression Trees}\label{sec_mtr_trees}

This section presents the traditional strategies for inducing decision tree algorithms for data streams. First, the general Hoeffding Tree algorithm is presented, followed by its application for MTR tasks. This variant will be from here onward referred to as the Multi-target regression Hoeffding Tree (MTR-HT).

\subsection{Hoeffding Tree algorithm}

The previous tree-based solutions for online STR and MTR use the Hoeffding bound (HB)~\citep{hoeffding1963probability} for performing decision splits.
This idea was first proposed by \cite{domingos2000mining} in their well-known work proposing the VFDT algorithm.
HB provides statistical evidence that, given enough observations, the current most promising split decision is the best one.
Therefore, splits are only performed when enough statistical evidence is gathered by the decision tree induction algorithm.
Thus, the split decisions have statistical guarantees to deviate from the expected value by at most a value $\xi$. 

Suppose a heuristic measure $h$ that provides a score for each attempted split decision for a predictive feature.
At time step or instance $n$, the current heuristic value is denoted by $h^n$. The higher the $h$, the better the candidate input space partitioning is.
Let $x_b$ be the input feature with the current best split candidate, with a score of $h_b$.
Besides, let $x_{sb}$ be the feature with the second best split heuristic score $h_{sb}$.
By monitoring the ratio $\frac{h_{sb}}{h_b}$ over the time, a new random variable $r \in [0, 1]$ can be derived by using

\begin{equation*}
    r \in \left\{\frac{h_{sb}^{1}}{h_b^{1}}, \ldots, \frac{h_{sb}^{n}}{h_b^{n}}, \frac{h_{sb}^{n+1}}{h_b^{n+1}}, \ldots\right\}.
\end{equation*}

Considering that a stream can be potentially unbounded, calculating the expected value of $r$ is not trivial.
On the other hand, the variable mean value in time step $n$, $\bar{r}^n$, can be easily calculated as

\begin{equation*}
    \bar{r}^{n} = \frac{1}{n}\left(\frac{h_{sb}^{1}}{h_b^{1}} + \frac{h_{sb}^{2}}{h_b^{2}} + \ldots + \frac{h_{sb}^{n}}{h_b^{n}}\right) = \frac{1}{n}\sum_{i=1}^{n} \frac{h_{sb}^{i}}{h_b^{i}}.
\end{equation*}

Using Hoeffding's inequality~\citep{hoeffding1963probability}, we can state that the expected value of $E(r)$ will not deviate from its sample mean at time step $n$ by more than a factor $\xi$, with a confidence level $1 - \delta$. For brevity, herein the time/instance indicator $n$ will be omitted from the mathematical expressions.
Equation~\ref{eq_hoeff_ineq} gives the simplified form of the Hoeffding's inequality (considering the range of $r$) subjected to $\delta$.

\begin{equation}
    P(|\bar{r} - E(r)| > \xi) \le 2e^{-2n\xi^{2}}
    \label{eq_hoeff_ineq}
\end{equation}

From Equation~\ref{eq_hoeff_ineq}, we can isolate $\xi$ in terms of $\delta$ in the following form

\begin{equation}
    \xi = \sqrt{\frac{1}{2n}\ln{\frac{2}{\delta}}}.
    \label{eq_hb}
\end{equation}

The value of $\xi$ obtained from Equation~\ref{eq_hb} enables us to bound a deviation interval in the form $E(r) \in [\bar{r} - \xi, \bar{r} + \xi]$, with confidence level $1 - \delta$. Thus, if $\bar{r} + \xi < 1$ then $E(r) < 1$. Hence, we can assume that the split decision in the predictive feature that generated $h_b$ is indeed the best choice for making a new partition.
Nonetheless, in some cases, two decision splits may achieve almost equal heuristic scores.
This implies that they are equally good choices.
An extreme example of that situation is the presence of repeated or redundant input features.
In these cases, if the value of $\xi$ is substantially shrunk, no split decision will be made.
To avoid this situation, an additional threshold or tie-break parameter $\tau$ is added.
Hence, a split is performed if $\bar{r} + \xi < 1$ or $\xi < \tau$.

It worth mentioning that, though some issues with the statistical guarantees of the HB usage in HTs have been identified, still, these algorithms present consistent performance in practice~\citep{ikonomovska2015online}.
Corrections to these problems have been proposed in the literature, e.g., the works of \cite{rutkowski2013decision} and \cite{matuszyk2013correcting}. Nonetheless, the choice for HT is still a trend for practitioners.
Independently of the raised concerns, our proposal can be easily adapted to work with any heuristic strategy to control tree growth.

\subsection{Multi-target regression Hoeffding Trees}

Both the FIMT-MT and iSOUP-Tree employ the intra-cluster variance reduction (ICVarR) as the heuristic score, following the steps of the predictive clustering framework~\citep{kocev2013tree}.
In this proposal, the variance measures the dispersion of the objects in the partition (i.e., a cluster) from their center of mass (the centroid)~\citep{ikonomovska2011incremental}.
The ICVarR calculation for a set of partitions $P$ over $\mathbf{Y}$ is given by

\begin{equation*}
    \text{ICVarR}(P) = \text{ICVar}(P) - \sum_{p \in P}\frac{|p|}{|P|}\text{ICVar}(p),
\end{equation*}

\noindent where the intra-cluster variance (ICVar) is calculated for an example $\mathbf{y} \in \mathbf{Y}$ as follows:

\begin{equation*}
    \text{ICVar}(\mathbf{y}) = \frac{1}{d}\sum_{t=1}^d\text{Var}(y_t).
\end{equation*}

Sufficient statistics must be stored to incrementally estimate the variance for each target. 
These variances are used to evaluate the split candidates.
As shown in recent MTR literature in data streams~\citep{ikonomovska2011incremental,osojnik2018tree}, maintaining a counter of the number of elements seen ($n$), the sum of each target $y_t$ ($\sum y_t$), $t \in \{1, ..., d\}$, and the sum of their square values ($\sum y_t^2$) for each leaf is enough to calculate the required measures.
Besides, we can also standardize the features $\mathbf{X}$ and targets by maintaining the same set of statistics for the inputs.
This action is especially relevant to avoid possible different scales for the features and targets having an negative impact on the linear prediction models and in the obtained ICVar.
The input features and targets are standardized using the z-score approach, i.e., they are centered by their mean values and scaled by their standard deviation~\citep{osojnik2018tree}.

Numerical attributes are monitored using the Extended Binary Search Tree (E-BST), as proposed by \cite{ikonomovska2011learning,ikonomovska2011incremental} and later expanded by \cite{osojnik2018tree}.
Often, this structure is also referred to in the literature as \textit{attribute observer}.
The FIMT-MT algorithm does not support categorical features, as previously mentioned, but this functionality was added by the iSOUP-Tree.
This algorithm creates a tree branch for each possible value in the nominal feature after splitting.

The original proposal of FIMT-MT only uses perceptron models with linear activation functions at the leaves; one predictor for each target. These models are incrementally trained using the Delta rule~\citep{ikonomovska2011learning} for linear weight updating.
The iSOUP-Tree algorithm introduces the use of adaptive models for each target, i.e., it decides between using the perceptrons or a more straightforward mean predictor for each new incoming instance.
To this end, a fading metric of error is monitored to assess which is the current best performer for each target.

As previously mentioned, neither of the employed predicting strategies for the leaf nodes effectively take into consideration the existence of inter-target correlations.
We reason that this possibility should be considered when making predictions.
We believe it can increase the accuracy of the whole tree model, as well as leverage intrinsic characteristics of the dataset for making predictions.
Next, we describe how SST-HT is capable of considering this aspect during the tree construction.
\section{Online Multi-target regression Tree with stacked leaf models}\label{sec_sst_ht}

\begin{figure}[!htb]
    \centering
    \includegraphics[width=\textwidth]{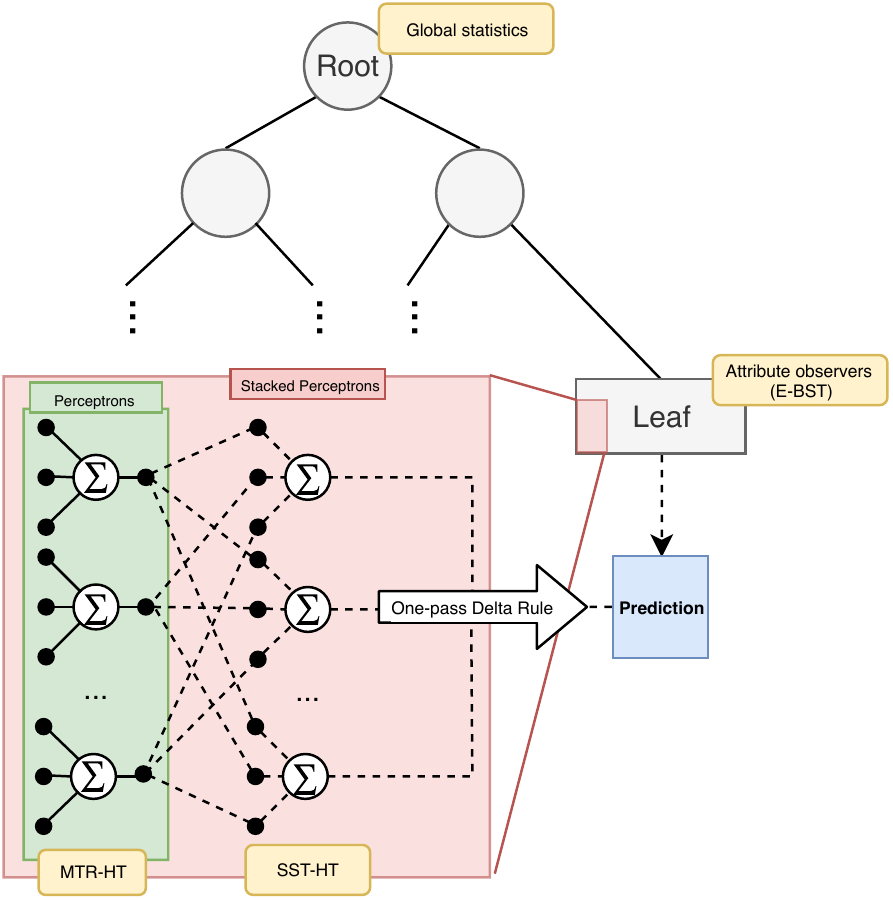}
    \caption{Overview of the SST-HT algorithm}
    \label{fig_sst_ht_overview}
\end{figure}

The algorithm proposed in this paper, the Stacked Single-target Hoeffding Tree (SST-HT), was tailored to encompass the best aspects of the existing MTR tree-based solutions, while increasing the prediction performance.
Our reasoning is that if there was a partition in the target space, the targets in this space must be inter-correlated.
By using stacked models~\citep{gama2000cascade,spyromitros2016multi}
for the leaves, these inter-correlations can be explored to decrease the prediction error.
Figure\ref{fig_sst_ht_overview} illustrates our proposal. SST-HT builds upon the original MTR-HT tree algorithm by using an additional layer of prediction models at the leaf nodes.
The other structures, such as attribute observers and statistics, are directly inherited from MTR-HT, as indicated in the figure.

The traditional use of linear models in the leaf nodes is the creation of as many perceptrons as the number of targets $d$.
Therefore, the predictions $\hat{\mathbf{y}}$ are computed separately, only considering the original problem's features and a bias term.
As previously mentioned, the input features for each instance are standardized using the z-score strategy, resulting in a normalized instance $\tilde{\mathbf{x}}$.
The normalized prediction $\tilde{y}_t$ for the $t$-th target is calculated as follows:

\begin{equation*}
    \tilde{y}_t = \beta_{0,t} + \sum_{j=1}^{m} \beta_{j,t}\tilde{x}_{j},
\end{equation*}

\noindent where $\beta_j$, $j \in \{0, 1, ..., m\}$ are the weights of the linear model.
Given the standardized value of the expected response $\breve{y}$, the linear predictor's weights are updated with the Delta rule

\begin{equation*}
    \beta_{j,t} \gets \beta_{j,t} + \eta(\tilde{y}_t - \breve{y}_t)x_j,
\end{equation*}

\noindent where $\eta$ represents the learning rate.
In the standard MTR-HT models, the final predictions are computed by transforming the $\tilde{y}$ values back to their original scales and ranges. Our proposal, in turn, adds another layer of linear models to combine and enhance the predictions from MTR-HT models.
We call these newly added predictors meta models, whereas the regressors at the first layer are referred to as base models. 

It is worth mentioning that since both base and meta predictors are linear transformations, they could be merged into a single transformation matrix.
This could be done by multiplying the neurons’ weight matrices to obtain a more compact representation.
Nonetheless, this operation is costly to perform for each new sample, and thus was avoided in our implementation.
Stacking multiple linear predictors is redundant in batch scenarios, as pointed out by \cite{borchani2015survey}.
A single linear transformation can achieve the same results.
However, as our online predictors process each incoming sample just once, i.e., they solve the linear system incrementally, we experimentally observed that the stacking of linear regressors improved the predictive performance.
In the future, we intend to investigate non-linear alternatives for incremental regression.
They need to be fast, robust, and flexible to deal with arbitrary regression regions.
Moreover, regularization strategies can also be investigated.

After applying the meta models, the new normalized responses are computed as follows:

\begin{equation*}
    \tilde{y}_{t}^{'} = \gamma_{0,t} + \sum_{k=1}^{d} \gamma_{k,t}\tilde{y}_{k,t},
\end{equation*}

\noindent and their corresponding weights $\gamma_{k,t}$, $k \in \{0, 1, ..., d\}$ are updated using the delta rule as well:

\begin{equation*}
    \gamma_{k,t} \gets \gamma_{k,t} + \eta(\tilde{y}_{t}^{'} - \breve{y}_t)\tilde{y}_{t}.
\end{equation*}

It must be observed that in these weight update expressions there are no input values for the bias terms. This value is typically set to the unit value, as we did.
Besides, both expressions use the same learning factor ($\eta$).
In the experiments carried out in this paper with SST-HT, the same $\eta$ value was used for both the base and meta models.
The use of different learning rates for the base and meta layers will be investigated in future research.
Another possibility is to use decaying factors for the learning rates.
However, fine-tuning the prediction models is out-of-the-scope of the current work.

Similar to the iSOUP-Tree, SST-HT uses an adaptive model to continuously select one among a set of predictors.
While the iSOUP-Tree chooses between the perceptron and mean predictors, SST-HT adds a third model: a stacked perceptron predictor.
In the same way as the preceding algorithm, SST-HT uses a fading error metric for online predictor selection.
As in the iSOUP-Tree, we monitor prediction performance using the faded Mean Absolute Error (fMAE).
This metric, presented in Equation~\ref{eq_fmae}\footnote{Note that the decaying factor is fixed in the expression. We intend to evaluate strategies for dynamically choosing the decaying factors of this metric in the future.}, uses an exponential decay to assign more importance to the most recent cases.
In this equation, $\text{\textit{learner}} \in $ \{\textit{mean}, \textit{perceptron}, \textit{stacked perceptron}\}.

\begin{equation}
    \text{\textit{fMAE}}_{\text{\textit{learner}}}(t) = \frac{\sum_{j=1}^{n} 0.95^{n-j}|y_t-\hat{y}_t|}{\sum_{j=1}^{n} 0.95^{n-j}}
    \label{eq_fmae}
\end{equation}

SST-HT allows the selection of either a specific prediction model to use or a dynamic selection.
Note that the automatic selection of predictors does not impact the tree structure, since the splits only consider the increase in partitions' homogeneity, regardless of the predictive errors.
Therefore, we expect the same tree structures for SST-HT and traditional MTR HTs.
On the other hand, due to the  additional set of predictors required at each leaf node, we expect a small increase in the memory use and increased training times.
It is worth mentioning that, as the number of input features is typically much smaller than the number of targets, i.e., $d \ll m$, the meta models have less adjustable parameters (weights) than their base counterparts.
\section{Experimental Setup}\label{sec_experimental_setup}
This section describes how the experiments were carried out, including datasets, settings for the tree predictors, performance metrics and evaluation strategy. All the experiments were executed using the \texttt{scikit-multiflow}\footnote{Available in: \url{https://github.com/scikit-multiflow/scikit-multiflow}} framework~\citep{montiel2018scikit}, which is an open and free platform for data stream mining.
Besides, our scripts were executed in a CentOS 7.2 Linux server using a single processing blade which contained 128 GB of DDR3 1866MHz memory, and two ten-core processors Intel Xeon E5-2680v2 at 2.8 GHz.

\subsection{Datasets}

A total of 16 datasets, synthetic and real, and from different application domains were used in the experiments.
The main characteristics of these datasets can be seen in Table~\ref{tab_datasets}.
Some of these have been used in previous MTR studies~\citep{duarte2015multi,osojnik2018tree}, whereas the datasets \textit{CPU, NPSDecay, SCFP, Sulfur}, and \textit{Wine} are used for the first time, in the context of online MTR, in this work.

All these new datasets, except for NPSDecay, are derived from real-world data. They are originally used for batch or online ST classification and regression. NPSDecay is built upon synthetic data collected from a vessel simulator~\citep{cipollini2018condition}.
Moreover, the version of SCFP used was specially tailored for this work.
A description of the datasets can be seen in Appendix~\ref{sec_appendix_datasets}.

\begin{table}[!htbp]
    \caption{Datasets used in the experiments. Datasets marked with $\star$ in the Source column were first proposed or adapted to MTR tasks in this research (please refer to Appendix~\ref{sec_appendix_datasets} for more details)}
    \centering
    \renewcommand{\tabcolsep}{4pt}
    \resizebox{\textwidth}{!}{
    \begin{tabular}{lrrrrl}
        \toprule    
        \thead[l]{\textbf{Dataset}} & \thead{\textbf{\#Examples}} & \thead{\textbf{\#Numeric}\\\textbf{Inputs}} & \thead{\textbf{\#Categorical}\\\textbf{Inputs}} & \thead{\textbf{\#Outputs}} & \thead{\textbf{Source}} \\ \midrule
        2Dplanes & 256,000 & 20 & 0 & 8 &  \cite{duarte2015multi}\\
        Bicycles & 17,379 & 4 & 9 & 3 & \cite{duarte2015multi}\\
        CPU & 8,192 & 22 & 0 & 4 & -, $\star$ \\
        Electricity & 45,312 & 6 & 0 & 2 & \cite{gama2004learning}, $\star$ \\
        Eunite03 & 8,064 & 29 & 0 & 5 & \cite{duarte2015multi}\\
        FriedD & 256,000 & 10 & 0 & 4 & \cite{duarte2015multi}\\
        FriedAsyncD & 256,000 & 10 & 0 & 4 & \cite{duarte2015multi}\\
        MV & 256,000 & 16 & 4 & 9 & \cite{duarte2015multi}\\
        NPSDecay & 455,109 & 25 & 0 & 5 & \cite{cipollini2018condition}, $\star$ \\
        RF1 & 9,005 & 64 & 0 & 8 & \cite{spyromitros2016multi}\\
        RF2 & 7,679 & 576 & 0 & 8 & \cite{spyromitros2016multi}\\
        SCFP & 223,129 & 54 & 3 & 3 & -, $\star$ \\
        SCM1d & 9,803 & 280 & 0 & 16 & \cite{spyromitros2016multi}\\
        SCM20d & 8,966 & 61 & 0 & 16 & \cite{spyromitros2016multi}\\
        Sulfur & 10,081 & 5 & 0 & 2 & \cite{fortuna2007soft}, $\star$\\
        Wine & 6,497 & 8 & 0 & 4 & \cite{cortez2009modeling}, $\star$\\ \bottomrule
    \end{tabular}}
    \label{tab_datasets}
\end{table}

\subsection{Settings used in the tree predictors}

During the experiments, we fixed some hyper-parameters for the tree predictors according to values proposed in the literature~\citep{domingos2000mining,duarte2015multi,osojnik2018tree}. Split attempts were performed at intervals of $200$ examples.
We set the significance level for the HB calculation to $\delta=10^{-7}$, and the tie-break parameter to $\tau=0.05$.

Additionally, in all cases, $200$ examples were used to initiate the tree predictors, providing a `warm' start for the evaluations.
Finally, the perceptron weights were started with uniform random values in the range $[-1,1]$.
In case of a split, new leaf nodes inherit their ancestors' weights.

Regarding the decision tree induction algorithms, we compared two variants of our proposal with iSOUP-Tree and two variants of the MTR-HT algorithm. Table~\ref{tab_methods_variants} summarizes the variants used in the comparisons, including their main characteristics and acronyms.

\begin{table}[!htb]
    \centering
    \caption{Description of the algorithms used in the comparisons}
    \begin{tabular}{lp{0.7\textwidth}}
        \toprule
        \textbf{Acronym} & \textbf{Description} \\ \midrule
        \multirow{2}{*}{$\text{MTR-HT}_{\text{Mean}}$} & MTR-HT variant that uses the mean of the targets as responses at the leaf nodes \\ \hline
        \multirow{2}{*}{$\text{MTR-HT}_\text{Perceptron}$} & MTR-HT variant that uses a perceptron model per target at the leaf nodes \\ \hline
        \multirow{2}{*}{$\text{iSOUP-Tree}$} & ISOUP-Tree algorithm that dynamically selects between the two previous prediction variants \\ \hline
        \multirow{2}{*}{$\text{SST-HT}$} & Variant of MTR-HT (proposed algorithm) that always use the stacked regressors for making predictions \\ \hline
        \multirow{2}{*}{$\text{SST-HT}_\text{Adaptive}$} & Variant of MTR-HT (proposed algorithm) that dynamically selects between the mean, perceptron, and stacked perceptron predictors for each target \\
        \bottomrule
    \end{tabular}
    \label{tab_methods_variants}
\end{table}

\subsection{Evaluation strategy}\label{sec_evaluation_strategy}

To compare the predictive performance of the algorithms, we used the prequential strategy~\citep{gama2010knowledge}.
In this strategy, after an example is evaluated by a predictive model, it is used to update the model.
For all the metrics used, we computed their mean value and also considered windowed measurements.
For such, we employed a non-overlapping sliding window of size $200$.
All MTR algorithms were applied thirty times to each dataset with varying seeds for the pseudo-random generators.
In order to reduce effects of randomness and operational system external influences, we used the average of the thirty results.
These effects relate mostly to the running time measurements, but the perceptron models in some of the tree variants ought to be also affected.

In particular, we used the Average Root Mean Square Error ($\overline{\text{RMSE}}$) as the error metric.
Both errors per sliding window and an overall measurement using all the seen data were considered in our analysis.
Equation~\ref{eq_armse} shows how the $\overline{\text{RMSE}}$ is calculated.

\begin{equation}
    \overline{\text{RMSE}} = \frac{1}{d}\sum_{t=1}^{d}\sqrt{\frac{\sum_{i=1}^{N} (y_i^{t} - \hat{y}_i^{t})^ {2}}{N}}
    \label{eq_armse}
\end{equation}

The average amount of time spent by each algorithm (in seconds) and the total of memory resources consumed by the predictors (in MB) are also reported.
We performed statistical tests to verify whether the differences in the predictive performance of the models are statistically significant, regarding the evaluation metrics.
The Friedman test and post-hoc Nemenyi test were used (with $\alpha=0.05$), as described in \cite{demvsar2006statistical}.
We considered the windowed evaluations to perform the statistical tests.
These measurements describe how the algorithms evolved over the time.
We summed the error metric values accounted for each window to perform the tests.
\section{Results and Discussion}\label{sec_results}

This section presents and discuss the main experimental results observed for the compared MTR algorithms. The results are discussed regarding predictive performance, running time and model size. We also highlight some cases in details, while presenting detailed information about all the evaluated datasets in the supplementary material (see Appendix~\ref{sec_appendix_line_plots}).

\subsection{Predictive Performance}

Table~\ref{tab_armse} summarizes the predictive performance of the investigated algorithms considering the mean measured errors, i.e., considering the average of the errors after processing the whole stream. The smallest $\overline{\text{RMSE}}$ value observed for each dataset is highlighted in bold.

\begin{table}[!htbp]
	\caption{Mean $\overline{\text{RMSE}}$ values observed (after processing the whole stream) }
    \label{tab_armse}
	\centering
	\setlength{\tabcolsep}{3pt}
	\resizebox{\textwidth}{!}{
	\begin{tabular}{lrrrrr}
		\toprule
		Dataset & $\text{MTR-HT}_{\text{Mean}}$ & $\text{MTR-HT}_\text{Perceptron}$ & $\text{iSOUP-Tree}$ & $\text{SST-HT}$ & $\text{SST-HT}_\text{Adaptive}$\\
		\midrule
		2Dplanes & $2.7507 \pm 0.00$ & $4.5075 \pm 0.00$ & $\mathbf{2.7340 \pm 0.00}$ & $4.8736 \pm 0.00$ & $2.7372 \pm 0.00$\\
		Bicycles & $\mathbf{87.3947 \pm 0.00}$ & $139.7503 \pm 0.01$ & $101.0994 \pm 0.00$ & $135.6794 \pm 0.02$ & $102.4765 \pm 0.00$\\
		CPU & $5.1473 \pm 0.00$ & $4.3383 \pm 0.00$ & $\mathbf{3.4078 \pm 0.00}$ & $4.8185 \pm 0.00$ & $3.4265 \pm 0.00$\\
		Electricity & $0.0242 \pm 0.00$ & $0.0313 \pm 0.00$ & $0.0207 \pm 0.00$ & $0.0264 \pm 0.00$ & $\mathbf{0.0206 \pm 0.00}$\\
		Eunite03 & $24.0007 \pm 0.00$ & $26.3034 \pm 0.00$ & $26.0206 \pm 0.00$ & $\mathbf{22.0181 \pm 0.00}$ & $22.9048 \pm 0.00$\\
		FriedD & $10.4575 \pm 0.00$ & $8.5829 \pm 0.00$ & $7.9927 \pm 0.00$ & $8.2212 \pm 0.00$ & $\mathbf{7.3887 \pm 0.00}$\\
		FriedAsyncD & $9.0928 \pm 0.00$ & $8.4728 \pm 0.00$ & $7.4457 \pm 0.00$ & $8.1805 \pm 0.00$ & $\mathbf{6.8746 \pm 0.00}$\\
		MV & $23.7614 \pm 0.00$ & $32.6842 \pm 0.00$ & $\mathbf{23.7486 \pm 0.00}$ & $33.3511 \pm 0.00$ & $23.7659 \pm 0.00$\\
		NPSDecay & $0.0195 \pm 0.00$ & $0.0251 \pm 0.00$ & $0.0147 \pm 0.00$ & $0.0205 \pm 0.00$ & $\mathbf{0.0137 \pm 0.00}$\\
		RF1 & $23.3911 \pm 0.00$ & $28.0100 \pm 0.00$ & $12.5794 \pm 0.00$ & $19.5451 \pm 0.00$ & $\mathbf{9.1290 \pm 0.00}$\\
		RF2 & $23.7385 \pm 0.00$ & $59.2034 \pm 0.00$ & $21.2524 \pm 0.00$ & $55.0323 \pm 0.00$ & $\mathbf{18.9755 \pm 0.00}$\\
		SCFP & $10.1314 \pm 0.00$ & $10.0563 \pm 0.00$ & $9.4049 \pm 0.00$ & $9.8438 \pm 0.00$ & $\mathbf{9.3349 \pm 0.00}$\\
		SCM1d & $234.3647 \pm 0.00$ & $355.7236 \pm 0.00$ & $215.4752 \pm 0.00$ & $298.3532 \pm 0.00$ & $\mathbf{198.0668 \pm 0.00}$\\
		SCM20d & $246.2019 \pm 0.00$ & $194.4146 \pm 0.00$ & $145.7449 \pm 0.00$ & $176.4975 \pm 0.00$ & $\mathbf{135.0029 \pm 0.00}$\\
		Sulfur & $0.0580 \pm 0.00$ & $0.0441 \pm 0.00$ & $0.0439 \pm 0.00$ & $0.0445 \pm 0.00$ & $\mathbf{0.0438 \pm 0.00}$\\
		Wine & $0.5958 \pm 0.00$ & $\mathbf{0.4424 \pm 0.00}$ & $0.4445 \pm 0.00$ & $0.4605 \pm 0.00$ & $0.4438 \pm 0.00$\\
		\midrule
		\textbf{{Average rank}} & $3.75$ & $4.12$ & $2.00$ & $3.62$ & $\mathbf{1.50}$\\
		\bottomrule
	\end{tabular}}
\end{table}

As shown in the table, SST-HT$_{\text{Adaptive}}$ presented the best predictive performance in the majority of datasets ($10$ out of $16$ datasets). The simplest, most straightforward SST-HT version presented the smallest $\overline{\text{RMSE}}$ for only one dataset (Eunite03). The same observation holds for MTR-HT$_{\text{Mean}}$ (Bicycles) and MTR-HT$_{\text{Perceptron}}$ (Wine). The second best performer in this analysis was the iSOUP-Tree, which obtained the best predictive performance in $3$ out of $16$ datasets. In general, as expected, the adaptive variants of the tree predictors obtained the best predictive performance most of the time.

\begin{figure}[!htbp]
    \centering
    \subfloat[Bicycles] {
        \includegraphics[width=0.95\textwidth]{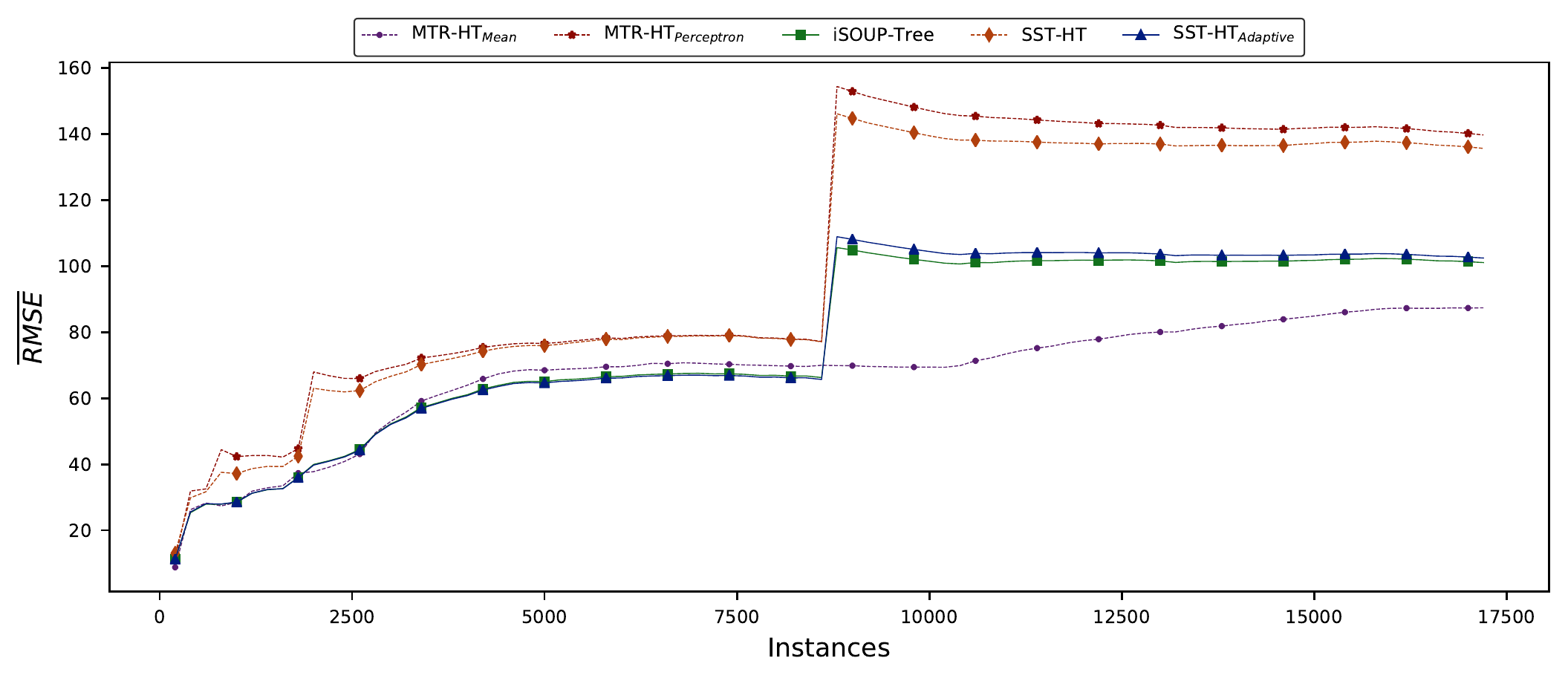}
        \label{fig_error_bicycles}
    }
    
    \subfloat[SCM1d] {
        \includegraphics[width=0.95\textwidth]{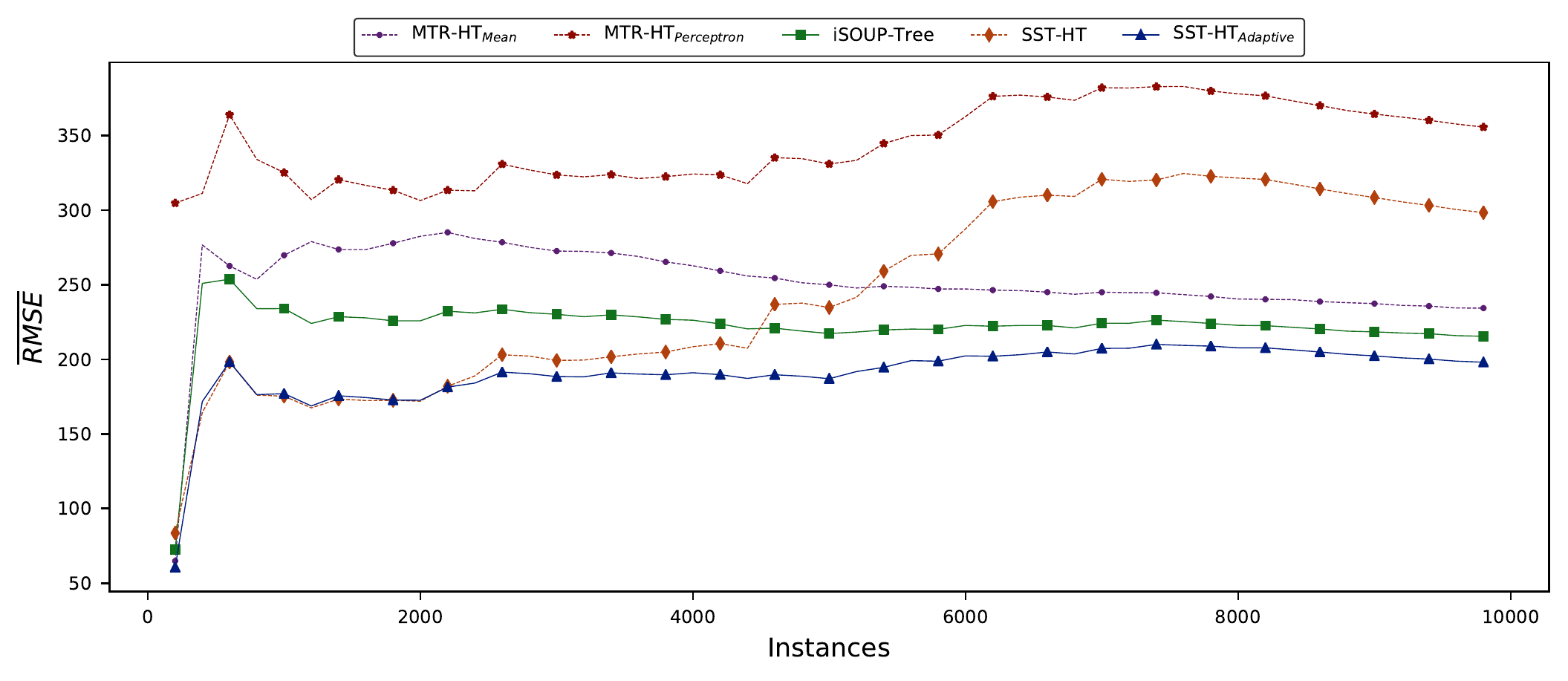}
        \label{fig_error_scm1d}
    }
    \caption{Example of the different patterns observed for the compared algorithms in the datasets Bicycles and SCM1d, regarding the $\overline{\text{RMSE}}$}
    \label{fig_error_lines_examples}
\end{figure}

The evolution of the observed error over time was also considered in our analysis. We observed different patterns, depending on the dataset being considered. To illustrate what occurs, we present line plots for two of the evaluated datasets, Bicycles and SCM1d, in Figures \ref{fig_error_bicycles} and \ref{fig_error_scm1d}, respectively. 

In the first case, almost all algorithms presented the same behavior regarding the $\overline{\text{RMSE}}$ values. Until around 8000 examples, the SST-HT$_{\text{Adaptive}}$ presented the smallest $\overline{\text{RMSE}}$ values. However, from this point until the end of the stream, the simplest MTR-HT$_\text{Mean}$ presented the smallest $\overline{\text{RMSE}}$. These results show that the underlying concepts of the data became stable. Hence, the mean value of the targets provided the best responses for the new cases. Interestingly, the model selection procedure did not appear to be effective in detecting this phenomenon, nor did the linear models behave well with the new samples. An alternative to overcome this problem is to use decaying factors different from $0.95$ to the faded error (refer to Section~\ref{sec_sst_ht}). This alternative would give less importance to the past cases, giving more attention to the current state. Ideally, the decaying factor could be set dynamically based on the characteristics of the incoming data.

In the second case, SST-HT$_{\text{Adaptive}}$ maintained the most accurate predictions for the whole stream. On the other hand, MTR-HT$_\text{Perceptron}$ presented the worst predictive performance. SST-HT presented the second best predictive performance until just after $4000$ examples, when there was a sudden increase in its $\overline{\text{RMSE}}$ values, presenting the second worst performance at the end of the stream. SST-HT and MTR-HT$_\text{Perceptron}$ presented similar error curves. The adaptive methods, differently, seemed to identify that the mean predictor became the best option, presenting error curves whose slopes were close to those presented by MTR-HT$_\text{Mean}$.
Different from what occurred with the Bicycles dataset,  the model selection mechanism worked well for the SCM1d dataset. This, again, reinforces the hypothesis that the tree must dynamically define the level of importance it gives to the current and past data.

Another possibility for these results is that in these previous cases the input to output and inter-output relations were not linear in the data partitions used. This could result in a better performance of the mean predictor when compared with the linear regressors. As previously presented, the investigation of non-linear regression methods for the leaf models is a promising venue for future research. Similar plots for all the considered datasets are presented in Appendix~\ref{sec_appendix_error_plots}.

\subsection{Running time}

As expected, the most simple algorithm (MTR-HT$_\text{Mean}$) was also the fastest in almost all the cases, as shown in Table~\ref{tab_time}. Again, the smallest running time per dataset is in bold. The MTR-HT$_\text{Perceptron}$ variant presented the second fastest running times in most of the cases. SST-HT, in general, performed as fast as the iSOUP-Tree for the majority of the datasets. As expected, SST-HT$_{Adaptive}$ presented the longest running time. This variant, in addition to performing a dynamic selection between predictors,  maintains and updates three different prediction models per leaf node. However, apart from MTR-HT$_\text{Mean}$, the difference between the SST-HT$_{Adaptive}$ and the other algorithms is is not very significant, regarding the running times. Moreover, this difference is compensated by the gains in predictive performance of our method. Concerning the cases where MTR-HT$_\text{Mean}$ was not the fastest algorithm, this occurred probably due to factors external to the algorithms, given the standard deviation of the observations.

\begin{table}[!htbp]
	\caption{Running time for each dataset (in seconds)}
    \label{tab_time}
	\centering
	\setlength{\tabcolsep}{3pt}
	\resizebox{\textwidth}{!}{
	\begin{tabular}{lrrrrr}
		\toprule
		Dataset & $\text{MTR-HT}_{\text{Mean}}$ & $\text{MTR-HT}_\text{Perceptron}$ & $\text{iSOUP-Tree}$ & $\text{SST-HT}$ & $\text{SST-HT}_\text{Adaptive}$\\
		\midrule
		2Dplanes & $\mathbf{81.2263 \pm 10.31}$ & $179.8277 \pm 28.29$ & $203.8099 \pm 27.35$ & $213.5590 \pm 26.37$ & $274.0997 \pm 42.55$\\
		Bicycles & $\mathbf{6.3643 \pm 1.56}$ & $11.5208 \pm 3.00$ & $11.8662 \pm 2.29$ & $11.9889 \pm 2.60$ & $14.8656 \pm 3.36$\\
		CPU & $\mathbf{23.7227 \pm 3.42}$ & $29.0668 \pm 8.23$ & $27.1192 \pm 4.18$ & $30.8000 \pm 8.89$ & $30.2394 \pm 8.23$\\
		Electricity & $\mathbf{19.2018 \pm 3.51}$ & $31.1544 \pm 5.64$ & $34.2767 \pm 5.52$ & $34.1326 \pm 6.12$ & $37.2195 \pm 2.39$\\
		Eunite03 & $\mathbf{9.6498 \pm 2.00}$ & $12.3035 \pm 2.99$ & $12.1218 \pm 1.40$ & $13.8561 \pm 3.49$ & $14.0753 \pm 2.57$\\
		FriedD & $\mathbf{389.1604 \pm 20.73}$ & $460.1738 \pm 29.18$ & $485.8485 \pm 28.41$ & $479.3183 \pm 34.06$ & $518.8329 \pm 32.35$\\
		FriedAsyncD & $\mathbf{387.3880 \pm 18.42}$ & $455.2480 \pm 28.32$ & $478.9144 \pm 25.80$ & $482.7184 \pm 31.05$ & $526.5462 \pm 32.65$\\
		MV & $\mathbf{672.7499 \pm 26.85}$ & $753.1689 \pm 33.84$ & $789.6671 \pm 38.35$ & $798.6512 \pm 40.73$ & $862.2696 \pm 51.03$\\
		NPSDecay & $\mathbf{1790.9435 \pm 72.67}$ & $1939.6547 \pm 68.31$ & $2002.5621 \pm 70.72$ & $1993.3690 \pm 70.17$ & $2084.4066 \pm 93.22$\\
		RF1 & $\mathbf{154.3603 \pm 21.47}$ & $174.3120 \pm 45.29$ & $161.3790 \pm 28.66$ & $187.6669 \pm 46.69$ & $175.2095 \pm 34.45$\\
		RF2 & $136.3630 \pm 19.10$ & $146.9837 \pm 29.91$ & $151.8701 \pm 36.37$ & $153.9039 \pm 30.42$ & $\mathbf{130.7061 \pm 9.33}$\\
		SCFP & $\mathbf{506.8263 \pm 24.40}$ & $587.0308 \pm 39.92$ & $616.7821 \pm 54.29$ & $595.3749 \pm 44.09$ & $610.9264 \pm 46.10$\\
		SCM1d & $713.6154 \pm 73.12$ & $726.5279 \pm 82.65$ & $726.7681 \pm 73.21$ & $738.1430 \pm 91.63$ & $\mathbf{683.5466 \pm 53.69}$\\
		SCM20d & $\mathbf{79.9444 \pm 11.90}$ & $83.3493 \pm 8.96$ & $86.5577 \pm 12.65$ & $92.6893 \pm 19.69$ & $90.2755 \pm 12.45$\\
		Sulfur & $\mathbf{9.5295 \pm 2.48}$ & $13.1352 \pm 3.85$ & $12.8251 \pm 3.19$ & $13.1586 \pm 3.44$ & $13.8657 \pm 3.46$\\
		Wine & $\mathbf{2.5600 \pm 0.53}$ & $4.6482 \pm 1.12$ & $5.0032 \pm 1.19$ & $5.2934 \pm 1.35$ & $5.4540 \pm 0.64$\\
		\midrule
		\textbf{{Average rank}} & $\mathbf{1.12}$ & $2.38$ & $3.19$ & $4.06$ & $4.25$\\
		\bottomrule
	\end{tabular}}
\end{table}

\begin{figure}[!htbp]
    \centering
    \includegraphics[width=\textwidth]{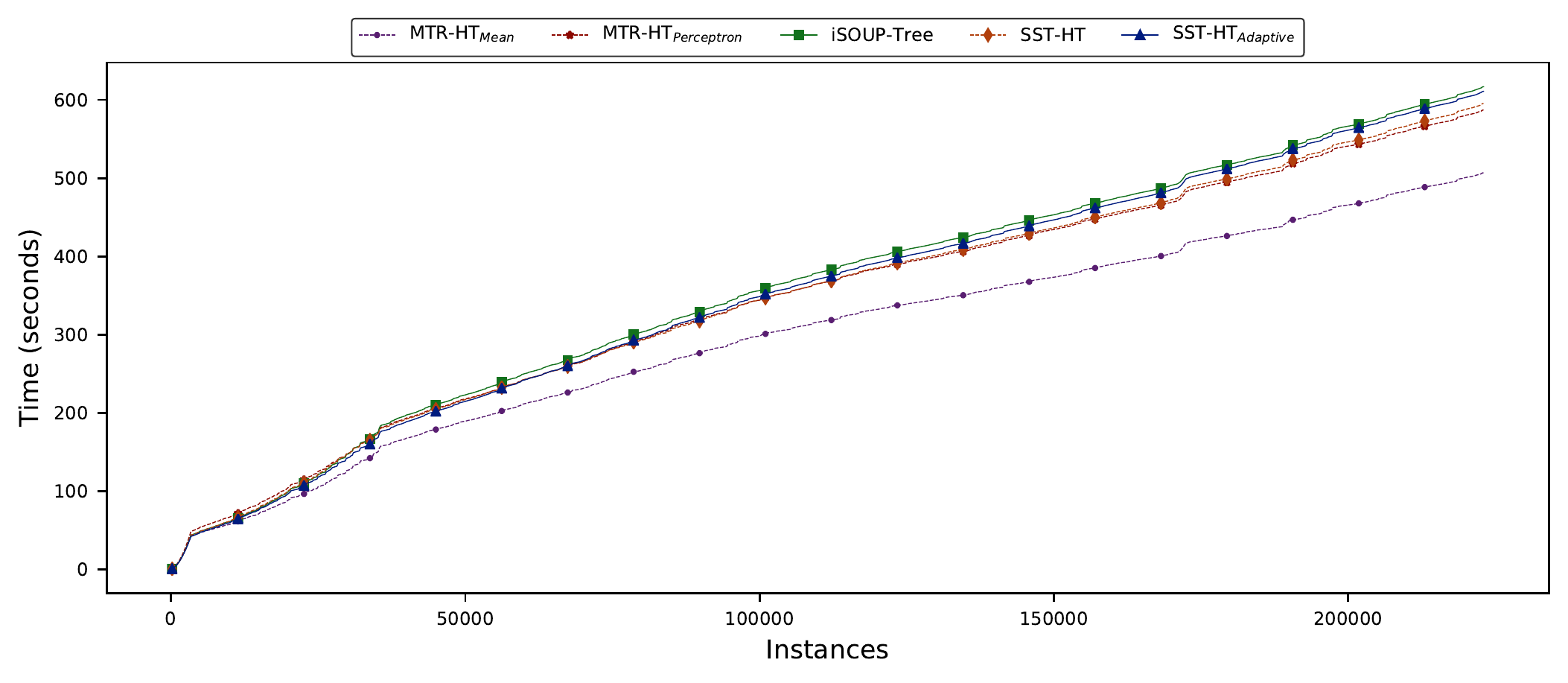}
    \caption{Running time for the SCFP dataset}
    \label{fig_running_time_scfp}
\end{figure}

Considering that all the compared  algorithms only differ in the strategy that the leaf nodes use to generate predictions, an approximately linear relationship between the running times of the algorithms was observed. This was expected, since the amount of extra processing performed by the different algorithms is leaf-wise constant.
This comparison for the dataset SCFP is illustrated in Figure~\ref{fig_running_time_scfp}.

Similar behaviors were observed for all the datasets. Detailed results for running time are presented in Appendix~\ref{sec_appendix_time_plots}.

\subsection{Model size}

When considering the size of the generated models, we obtained  results very similar  to those observed in the running times. This, again, was expected, given the use of an extra layer of predictors by SST-HT. The total sizes of the trained models at the end of the data streams are summarized in Table~\ref{tab_size}. The results are presented in Megabytes (MB). Excluding the variations of SST-HT (SST-HT and SST-HT$_{\text{Adaptive}}$), the sizes of the investigated algorithms only differed in small amounts, regardless of the dataset used (with $8$, $7$, and $1$ wins for the MTR-HT$_\text{Mean}$, iSOUP-Tree, and MTR-HT$_\text{Perceptron}$, respectively). The differences were very small in the majority of the cases.

The two versions of SST-HT spent more memory resources than the others. However, the additional amount of memory needed was minimal in almost all the cases.
This difference was negligible for most of the real-world application datasets.

\begin{table}[!htbp]
	\caption{Total model size for each dataset (in MB)}
    \label{tab_size}
	\centering
	\setlength{\tabcolsep}{3pt}
	\resizebox{\textwidth}{!}{
	\begin{tabular}{lrrrrr}
		\toprule
		Dataset & $\text{MTR-HT}_{\text{Mean}}$ & $\text{MTR-HT}_\text{Perceptron}$ & $\text{iSOUP-Tree}$ & $\text{SST-HT}$ & $\text{SST-HT}_\text{Adaptive}$\\
		\midrule
		2Dplanes & $19.3818 \pm 0.01$ & $19.4535 \pm 0.01$ & $\mathbf{19.2903 \pm 0.01}$ & $19.6516 \pm 0.00$ & $19.5243 \pm 0.01$\\
		Bicycles & $\mathbf{3.5271 \pm 0.00}$ & $3.5320 \pm 0.00$ & $3.5491 \pm 0.00$ & $3.5916 \pm 0.01$ & $3.5977 \pm 0.00$\\
		CPU & $22.0923 \pm 0.01$ & $22.0974 \pm 0.01$ & $\mathbf{22.0127 \pm 0.01}$ & $22.3349 \pm 0.00$ & $22.2608 \pm 0.01$\\
		Electricity & $\mathbf{12.4277 \pm 0.00}$ & $12.4434 \pm 0.00$ & $12.4639 \pm 0.00$ & $12.6361 \pm 0.00$ & $12.6669 \pm 0.00$\\
		Eunite03 & $8.3868 \pm 0.01$ & $8.3782 \pm 0.05$ & $\mathbf{8.1995 \pm 0.06}$ & $8.7168 \pm 0.01$ & $8.7260 \pm 0.01$\\
		FriedD & $993.7507 \pm 0.00$ & $993.8051 \pm 0.00$ & $\mathbf{993.5337 \pm 0.00}$ & $1005.1122 \pm 0.00$ & $1004.6132 \pm 0.00$\\
		FriedAsyncD & $\mathbf{967.7357 \pm 0.01}$ & $967.7916 \pm 0.00$ & $968.1268 \pm 0.01$ & $978.8781 \pm 0.00$ & $978.0164 \pm 0.00$\\
		MV & $1021.7871 \pm 0.01$ & $1021.8641 \pm 0.00$ & $\mathbf{1020.5312 \pm 0.02}$ & $1032.7374 \pm 0.07$ & $1031.3735 \pm 0.06$\\
		NPSDecay & $861.0028 \pm 0.04$ & $860.9132 \pm 0.04$ & $\mathbf{860.0335 \pm 0.02}$ & $870.7448 \pm 0.18$ & $869.4314 \pm 0.16$\\
		RF1 & $\mathbf{11.3013 \pm 0.01}$ & $11.3637 \pm 0.02$ & $11.3241 \pm 0.04$ & $12.9458 \pm 0.04$ & $12.8036 \pm 0.20$\\
		RF2 & $\mathbf{34.4144 \pm 0.02}$ & $34.5717 \pm 0.02$ & $34.5742 \pm 0.02$ & $36.5764 \pm 0.03$ & $36.6312 \pm 0.03$\\
		SCFP & $332.7387 \pm 0.02$ & $332.8372 \pm 0.02$ & $\mathbf{332.2960 \pm 0.09}$ & $337.0597 \pm 0.10$ & $335.9908 \pm 0.08$\\
		SCM1d & $660.6949 \pm 0.08$ & $\mathbf{660.5839 \pm 0.08}$ & $660.6439 \pm 0.29$ & $675.4183 \pm 0.00$ & $675.4258 \pm 0.01$\\
		SCM20d & $\mathbf{275.8278 \pm 0.01}$ & $275.8724 \pm 0.02$ & $275.8520 \pm 0.01$ & $276.6837 \pm 0.04$ & $276.6777 \pm 0.01$\\
		Sulfur & $\mathbf{9.5019 \pm 0.00}$ & $9.5055 \pm 0.00$ & $9.5099 \pm 0.00$ & $9.6541 \pm 0.00$ & $9.6615 \pm 0.00$\\
		Wine & $\mathbf{3.4698 \pm 0.00}$ & $3.4738 \pm 0.01$ & $3.4722 \pm 0.01$ & $3.5140 \pm 0.01$ & $3.5166 \pm 0.01$\\
		\midrule
		\textbf{{Average rank}} & $\mathbf{1.69}$ & $2.44$ & $1.88$ & $4.56$ & $4.44$\\
		\bottomrule
	\end{tabular}}
\end{table}

\begin{figure}[!htbp]
    \centering
    \includegraphics[width=\textwidth]{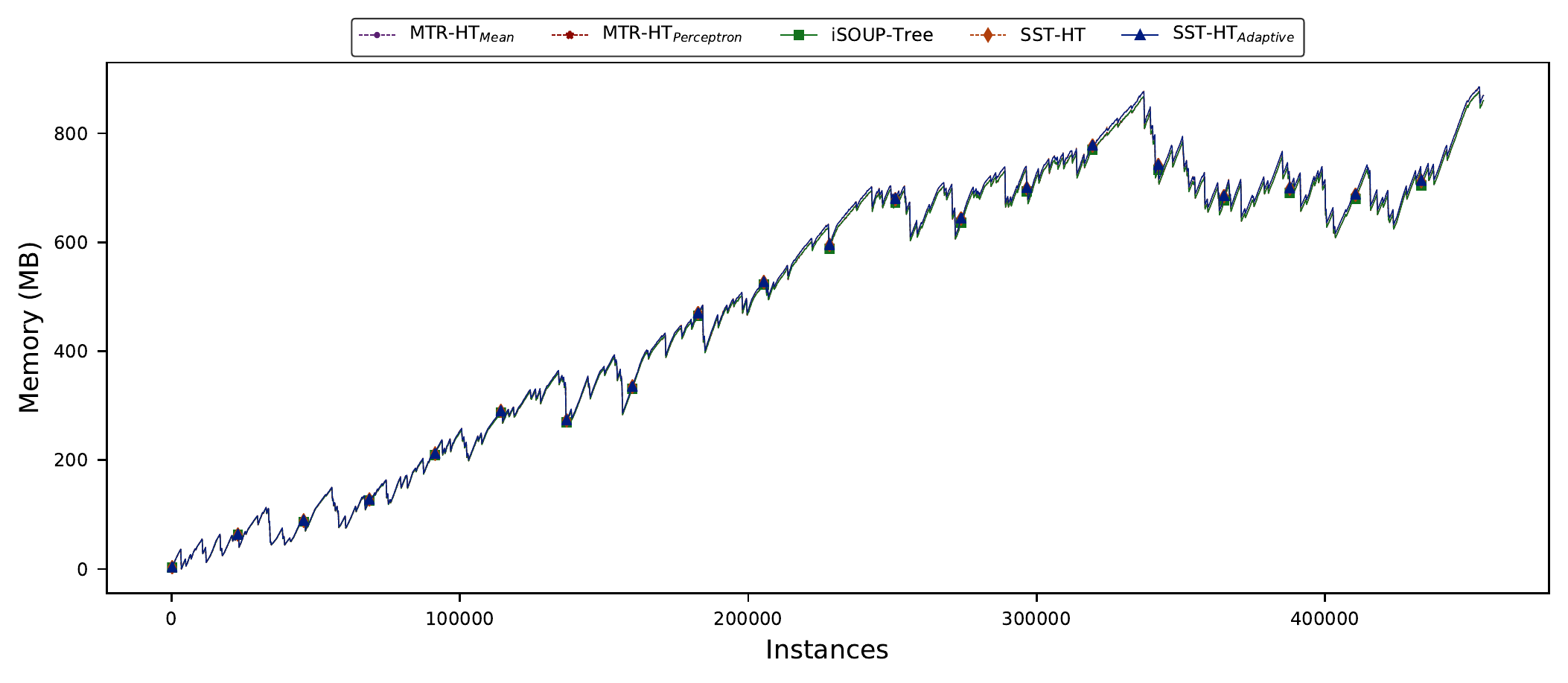}
    \caption{Memory usage by the compared algorithms on the NPSDecay dataset}
    \label{fig_line_memory_npsdecay}
\end{figure}

The relation between the amount of memory used by the different algorithms over time was also linear for nearly all cases. Considering that SST-HT does not impact the tree growth characteristics, the extra memory usage is constant in all the cases. This can be verified in detail for each dataset in Appendix~\ref{sec_appendix_memory_plots}. As a matter of illustration, we present the memory usage varying on time for the NPSDecay dataset in Figure~\ref{fig_line_memory_npsdecay}. This dataset has the highest number of examples among all the sets considered in our experiments.

\subsection{Statistical test analysis}

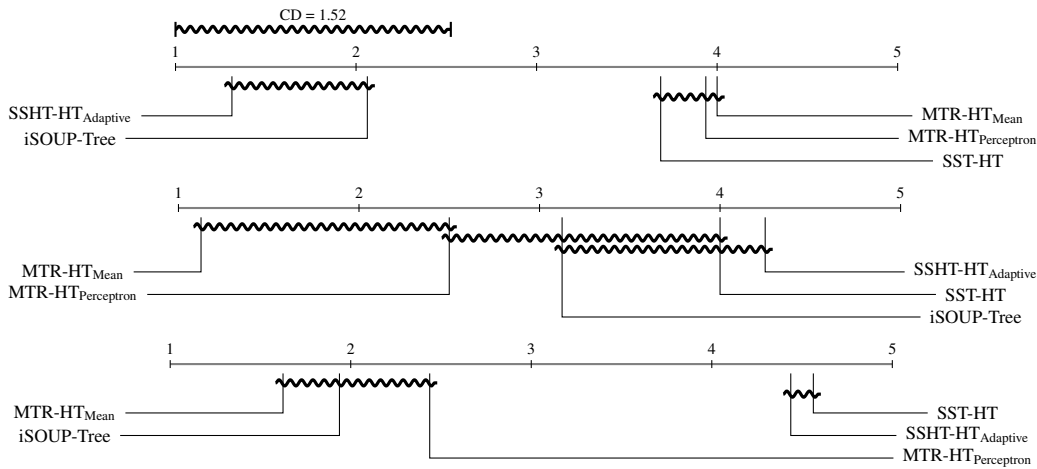
\begin{figure}
    \centering
    \begin{tikzpicture}[xscale=2]
        \node (Label) at (2.1149990163929138, 0.7){\tiny{CD = 1.52}}; 
        \draw[decorate,decoration={snake,amplitude=.4mm,segment length=1.5mm,post length=0mm},very thick, color = black] (1.2,0.5) -- (3.029998032785828,0.5);
        \foreach \x in {1.2, 3.029998032785828} \draw[thick,color = black] (\x, 0.4) -- (\x, 0.6);
         
        \draw[gray, thick](1.2,0) -- (6.0,0);
        \foreach \x in {1.2,2.4,3.6,4.8,6.0} \draw (\x cm,1.5pt) -- (\x cm, -1.5pt);
        \node (Label) at (1.2,0.2){\tiny{1}};
        \node (Label) at (2.4,0.2){\tiny{2}};
        \node (Label) at (3.6,0.2){\tiny{3}};
        \node (Label) at (4.8,0.2){\tiny{4}};
        \node (Label) at (6.0,0.2){\tiny{5}};
        \draw[decorate,decoration={snake,amplitude=.4mm,segment length=1.5mm,post length=0mm},very thick, color = black](1.525,-0.25) -- (2.525,-0.25);
        \draw[decorate,decoration={snake,amplitude=.4mm,segment length=1.5mm,post length=0mm},very thick, color = black](4.375,-0.4) -- (4.85,-0.4);
        \node (Point) at (1.575, 0){};\node (Label) at (0.5,-0.65){\scriptsize{$\text{SSHT-HT}_{\text{Adaptive}}$}}; \draw (Point) |- (Label);
        \node (Point) at (2.475, 0){};\node (Label) at (0.5,-0.95){\scriptsize{iSOUP-Tree}}; \draw (Point) |- (Label);
        \node (Point) at (4.8, 0){};\node (Label) at (6.5,-0.65){\scriptsize{$\text{MTR-HT}_{\text{Mean}}$}}; \draw (Point) |- (Label);
        \node (Point) at (4.725, 0){};\node (Label) at (6.5,-0.95){\scriptsize{$\text{MTR-HT}_{\text{Perceptron}}$}}; \draw (Point) |- (Label);
        \node (Point) at (4.425, 0){};\node (Label) at (6.5,-1.25){\scriptsize{SST-HT}}; \draw (Point) |- (Label);
    \end{tikzpicture}
    \begin{tikzpicture}[xscale=2]
         
        \draw[gray, thick](1.2,0) -- (6.0,0);
        \foreach \x in {1.2,2.4,3.6,4.8,6.0} \draw (\x cm,1.5pt) -- (\x cm, -1.5pt);
        \node (Label) at (1.2,0.2){\tiny{1}};
        \node (Label) at (2.4,0.2){\tiny{2}};
        \node (Label) at (3.6,0.2){\tiny{3}};
        \node (Label) at (4.8,0.2){\tiny{4}};
        \node (Label) at (6.0,0.2){\tiny{5}};
        \draw[decorate,decoration={snake,amplitude=.4mm,segment length=1.5mm,post length=0mm},very thick, color = black](1.3,-0.25) -- (3.05,-0.25);
        \draw[decorate,decoration={snake,amplitude=.4mm,segment length=1.5mm,post length=0mm},very thick, color = black](2.95,-0.4) -- (4.85,-0.4);
        \draw[decorate,decoration={snake,amplitude=.4mm,segment length=1.5mm,post length=0mm},very thick, color = black](3.7,-0.55) -- (5.1499999999999995,-0.55);
        \node (Point) at (1.35, 0){};\node (Label) at (0.5,-0.8500000000000001){\scriptsize{$\text{MTR-HT}_{\text{Mean}}$}}; \draw (Point) |- (Label);
        \node (Point) at (3.0, 0){};\node (Label) at (0.5,-1.1500000000000001){\scriptsize{$\text{MTR-HT}_{\text{Perceptron}}$}}; \draw (Point) |- (Label);
        \node (Point) at (5.1, 0){};\node (Label) at (6.5,-0.8500000000000001){\scriptsize{$\text{SSHT-HT}_{\text{Adaptive}}$}}; \draw (Point) |- (Label);
        \node (Point) at (4.8, 0){};\node (Label) at (6.5,-1.1500000000000001){\scriptsize{SST-HT}}; \draw (Point) |- (Label);
        \node (Point) at (3.75, 0){};\node (Label) at (6.5,-1.4500000000000002){\scriptsize{iSOUP-Tree}}; \draw (Point) |- (Label);
    \end{tikzpicture}
    \begin{tikzpicture}[xscale=2]
         
        \draw[gray, thick](1.2,0) -- (6.0,0);
        \foreach \x in {1.2,2.4,3.6,4.8,6.0} \draw (\x cm,1.5pt) -- (\x cm, -1.5pt);
        \node (Label) at (1.2,0.2){\tiny{1}};
        \node (Label) at (2.4,0.2){\tiny{2}};
        \node (Label) at (3.6,0.2){\tiny{3}};
        \node (Label) at (4.8,0.2){\tiny{4}};
        \node (Label) at (6.0,0.2){\tiny{5}};
        \draw[decorate,decoration={snake,amplitude=.4mm,segment length=1.5mm,post length=0mm},very thick, color = black](1.9,-0.25) -- (2.9749999999999996,-0.25);
        \draw[decorate,decoration={snake,amplitude=.4mm,segment length=1.5mm,post length=0mm},very thick, color = black](5.275,-0.4) -- (5.5249999999999995,-0.4);
        \node (Point) at (1.95, 0){};\node (Label) at (0.5,-0.65){\scriptsize{$\text{MTR-HT}_{\text{Mean}}$}}; \draw (Point) |- (Label);
        \node (Point) at (2.325, 0){};\node (Label) at (0.5,-0.95){\scriptsize{iSOUP-Tree}}; \draw (Point) |- (Label);
        \node (Point) at (5.475, 0){};\node (Label) at (6.5,-0.65){\scriptsize{SST-HT}}; \draw (Point) |- (Label);
        \node (Point) at (5.325, 0){};\node (Label) at (6.5,-0.95){\scriptsize{$\text{SSHT-HT}_{\text{Adaptive}}$}}; \draw (Point) |- (Label);
        \node (Point) at (2.925, 0){};\node (Label) at (6.5,-1.25){\scriptsize{$\text{MTR-HT}_{\text{Perceptron}}$}}; \draw (Point) |- (Label);
    \end{tikzpicture}

    \caption{Friedman test and Nemenyi post-hoc test results. Tests performed for $\overline{\text{RMSE}}$ (top), Running time (middle), and Model size (bottom)}
    \label{fig_nemenyi}
\end{figure}

Given that we observed different performance behaviors among the algorithms and along the streams, we applied a statistical significance test to assess the significance of the observed differences.
For such, we considered the metrics measured in the sliding windows, as presented in Section~\ref{sec_evaluation_strategy}.
The Friedman statistical test and the Nemenyi post-hoc test were used to compare the algorithms.
The results of these tests are graphically illustrated, as recommended in \cite{demvsar2006statistical}.
The main findings of this analysis can be seen in Figure~\ref{fig_nemenyi}. 

Concerning $\overline{\text{RMSE}}$, both SST-HT$_\text{Adaptive}$ and iSOUP-Tree appeared in the first group.
We did not observe statistically significant differences among their predictive performance.
Nonetheless, we firmly believe that given more observation, i.e., datasets, our algorithm's variant would perform significantly better than iSOUP-Tree in the statistical analysis.
The remaining algorithms appear in a second group, including the non-adaptive version of SST-HT.
These algorithms are statistically equivalent regarding their predictive performance.
Therefore, the adaptive choice of predictors proved itself to be one of the most relevant aspects to determine the final performance of the HT algorithms for regression.

When we consider the running time, as the prediction strategy becomes more sophisticated, the rankings become worse.
We expected this behavior.
Nonetheless, most of the considered algorithms did not present statistically significant differences in our analysis.
It is noteworthy that the variants of SST-HT were statistically equivalent to iSOUP-Tree in this test.

Lastly, regarding memory usage, our proposal has a disadvantage when faced with its competitors.
This fact was again expected since SST-HT needs to train and monitor and an additional set of predictors.
Hence, it ends up having an approximately constant increase in its size when compared with iSOUP-Tree.

In summary, SST-HT compared favorably to its competitors concerning its prediction error.
This observation is especially true when we take into account the dynamic mechanism for leaf predictor choice.
SST-HT adds extra memory and running time burden, but its prediction performance compensates for this fact.

\section{Final Considerations}\label{sec_conclusion}

In this work, we presented an extension for online MTR decision tree algorithms that better explores the characteristics of these kinds of problems. Our proposal, called SST-HT, improves the prediction performance without affecting the structure of the tree models. The main idea behind SST-HT is to use stacked linear models at the leaf nodes to capture and model the possible existing inter-target dependencies. Thus, the split decisions are made in the same way as those from the traditional online MTR tree algorithm. Similarly to existing solutions, SST-HT is also able to dynamically select the most adequate predictor for each instance. SST-HT, however, selects between three predictors: mean, perceptron or stacked perceptron predictors.

We evaluated two variations of SST-HT, experimentally comparing them with three well-known tree algorithms for dealing with MTR tasks in data streams. A large set of 16 benchmark datasets was used in the experimental evaluation. To the best of our knowledge, this is the most extensive set of online MTR datasets used so far. In the experiments carried out, the proposed algorithm obtained the most accurate predictions in the majority of the cases without demanding large increases in the amounts of computational resources. 

As future work, we intend to verify the possibility of extending our ideas to ensembles of decision rules, like those in AMRules. In this sense, the modeling of inter-target dependencies could be further improved, since AMRules creates rules which encompass subsets of targets with the highest inter-correlation. Moreover, we intend to evaluate other possibilities of stacked regression models for the leaves. Our goal is to find fast, robust, and flexible non-linear alternatives to the linear models. Using SST-HT as the base model for traditional online ensemble algorithms is another possibility for future research. Besides, we also want to evaluate alternatives for monitoring the necessary statistics for splitting numerical attributes, reducing the cost of this procedure. Finally, the application of our proposal to correlated tasks, e.g., online multi-label classification, could also be investigated.

\subsubsection*{Acknowledgments}
The authors would like to thank FAPESP (S{\~a}o Paulo Research Foundation) for its financial support (grants \#2018/07319-6, \#2016/18615-0 and \#2013/07375-0) and Intel Inc. for providing equipment for some of the experiments. 
The authors would also like to give our special thanks to Ricardo Sousa and Professor João Gama for kindly providing some of the datasets used in our experiments.

\bibliographystyle{apalike}

\clearpage

\begin{appendices}
\section{Used Datasets}\label{sec_appendix_datasets}

This appendix describes the datasets that were used in the experiments. Firstly, the datasets already reported as online MTR tasks in the literature are described. Next, the datasets that were used for the first time in this work are presented.

\subsection{Existing datasets}
This section briefly describes the datasets used in the experiments that were already reported in the literature~\citep{duarte2015multi,spyromitros2016multi,osojnik2018tree}.



\subsubsection*{Bicycles}

The \textit{Bicycles} dataset has already been used in multiple online MTR research~\citep{duarte2015multi,osojnik2018tree}. It describes the hourly count of rental bikes, considering the period between 2011 and 2012 in the \textit{Capital bikeshare system}~\citep{duarte2015multi}. The data contains weather and seasonal information for each rent event. The task consists of predicting the count of casual (non-registered), registered and total users.

\subsubsection*{Eunite03}
The Eunite03 dataset was used during the competition of the \textit{3rd European Symposium on Intelligent Technologies, Hybrid Systems and their implementation on Smart Adaptive Systems} (2003). The dataset describes a process of continuous production of manufactured glasses~\citep{duarte2015multi}. The input features describe the parameters used when producing the glass products, while the outputs refers to the glass quality. 

\subsubsection*{2Dplanes, FriedD, FriedAsyncD, and MV}

2Dplanes, FriedD, and FriedAsyncD are MTR artificial datasets generated by \cite{duarte2015multi}. They are modifications of well-known artificial ST regression tasks~\citep{breiman2017classification}. The FriedD and FriedAsyncD datasets contain one CD for each of the output targets. In FriedD the CD occur simultaneously for all the target variables in the middle of the data stream, while in FriedAsyncD the CDs occur asynchronously~\citep{duarte2015multi}. Lastly, MV was also constructed by \cite{duarte2015multi} based on an ST regression artificial problem.

\subsubsection*{RF1 and RF2}

The RF1 and RF2 (River Flow) datasets were firstly reported by \cite{spyromitros2016multi} and ever since then they have been used in MTR data streaming tasks~\citep{duarte2015multi,osojnik2018tree}. The datasets concern the prediction of river network flows considering a time window of 48 h in the future, at specific locations. Hourly flow observations were registered for 8 sites in the Mississippi River network (US) considering a period of one year (from September 2011 to September 2012). The data was obtained from the US National Weather Service. Each observation includes the most recent data, as well as delayed measurements considering intervals ranging from $[6,12,18,24,36,48,60]$ hours in the past. The first dataset, RF1, uses only the sensor data, whereas the second one, RF2, adds precipitation forecast information (expected rainfall) for each of the measurement sites.

\subsubsection*{SCM1d and SCM20d}

The SCM (Supply Chain Management) was extracted \textit{from the Trading Agent Competition in Supply Chain Management} (TACSCM) tournament in 2010. Again, these datasets were firstly proposed by \cite{spyromitros2016multi}, and were applied in data stream problems of MTR~\citep{duarte2015multi,osojnik2018tree}. Each example corresponds to an observation day in the tournament (from a total of 220 days in each game and 18 games during the whole tournament). The input variables correspond to the observed prices considering a specific tournament day. Additionally, four time-delayed observations are added for each observed product and component (delays of 1, 2, 4 and 8 days) aiming at facilitating the anticipation of trends. Each dataset has 16 targets, which correspond to the predictions for the next day mean price (SCM1d) or mean price for 20-days in the future (SCM20d), concerning each product in the simulation.

\subsection{\textbf{New datasets}}

This section describes the datasets that were firstly evaluated as MTR tasks in streaming scenarios in this work.

\subsubsection*{CPU}

The Computer Activity database\footnote{\url{https://www.cs.toronto.edu/~delve/data/comp-activ/desc.html}}, collected around 1996 at the University of Toronto, records multiple performance measures, such as the number of bytes read and written from the system memory. All data was collected from a computer Sun Sparctation model 20/712, which had 2 CPUs (Central Processing Unit) and 128 MegaBytes of main memory. The records concern the monitoring of normal computer usage, for example, browsing through the web or using text editors. The records were gathered at intervals of five seconds. Originally, the tasks related to this dataset concerned predicting the percent of the time the CPU ran in user mode. However, taking into consideration that the data also contains the amount of time the CPU ran in system mode, and the period it stayed in idle due to waiting for block IO or any other circumstances, the task was tackled as an incremental MTR problem.

\subsubsection*{Electricity}
The Electricity dataset is an adapted version of the well-known ELEC2 dataset~\citep{gama2004learning}, which is commonly used in online classification tasks. The original task corresponds to identifying the change of the price (up or down) in the Australian New South Wales Electricity Market. In this market the prices are not fixed, and are affected by aspects of demand and supply, and set every five minutes. The data comprehends an interval between 1996 and 1998, and each example in the dataset refers to a period of $30$ minutes. It is a scenario with a potential to multiples changes, given that transfers to/from the neighboring state of Victoria are performed to alleviate fluctuations. In this adapted version of the task, the original label property was discarded, and the prices for the New South Wales and Victoria states were set as the new targets to be predicted. As input features, we selected the remaining data properties: the measured electricity demands for those markets, the measurement time stamp, the day of the week, and the scheduled electricity transfer between the two states.

\subsubsection*{NPSDecay}
The NPSDecay dataset~\citep{cipollini2018condition} concerns the prediction of performance decay in a  Naval Propulsion System (NPS) over time. The data comes from a vessel (frigate) simulator which was specially tailored and fine-tuned over the years to represent the components of a possible real vessel. The simulated vessel has a combined diesel, electric, and gas propulsion plant. The targets correspond to decay coefficients for the main components of the NPS system, namely: the gas turbine, the gas turbine compressor, the hull, and the propeller. Hence, in this task the following coefficients must be predicted:

\begin{itemize}
    \item Propeller Thrust decay state coefficient (Kkt);
    \item Propeller Torque decay state coefficient (Kkq);
    \item Hull decay state coefficient (Khull);
    \item Gas Turbine Compressor decay state coefficient (KMcompr);
    \item Gas Turbine decay state coefficient (KMturb).
\end{itemize}

A total of 25 features related to parameters that indirectly represent the system state are available for each measurement of the performance decay coefficients. The dataset is available in OPenML, as well as in a website made available by its authors\footnote{\url{https://sites.google.com/view/cbm/home}}.

\subsubsection*{SCFP}

The See Click Fix Prediction (SCFP) competition\footnote{\url{https://www.kaggle.com/c/see-click-predict-fix}} was firstly held by \textit{Kaggle} as a \textit{hackathon}. Later on, the dataset adopted in that competition was used in a new competition promoted by the same organization. The dataset concerns registers of issues subjected by the population to the Open311\footnote{\url{http://www.open311.org/}} service. The original task consists of predicting the number of views, comments, and votes an issue would receive. The original dataset contains textual information about a summary and description of the issue, as well as geolocated data, the publication source (mobile, desktop, etc.), and a tag type for the publication. The original dataset contains missing data in many fields. A random $1\%$ sample of the mentioned dataset was used by \cite{spyromitros2016multi} in batch scenarios. However, their version simply overlooked the textual information contained in the examples, using only the other fields, as well as some hand-engineered features.

In our processed version of the original dataset, the categorical values were encoded using numeric values. The missing fields were encoded with $-1$. Following the approach of \cite{spyromitros2016multi}, in addition to the latitude and longitude fields, an additional attribute concerning the distance of the published issue to its city downtown (in meters) was also added. Besides, another field denoting the time interval (in hours) since the last registered issue was included in the dataset. Moreover, our main contribution to the previous and reduced version of SCFP was taking into account the textual information of the dataset. To this end, the summary of the issues was considered. We adopted a pre-trained word embedding~\citep{pennington2014glove} model\footnote{\url{https://nlp.stanford.edu/projects/glove/}} with an array of $50$ positions to encode each of the non-stopwords in the summary field of the issues. The mean vector among all the considered words was then taken as a representation of the issue's summary. Therefore, $50$ additional features were added to our version of SCFP.

\subsubsection*{Sulfur}

The Sulfur dataset concerns the prediction of pollutants concentration ($H_{2}S$ and $SO_2$), given air and gas flows as inputs. The dataset is available at OPenML~\citep{Vanschoren2014}, and corresponds to the data described in \cite{fortuna2007soft}. In the Sulfur dataset, no pre-processing step was performed.

\subsubsection*{Wine}

The Wine dataset~\citep{cortez2009modeling} describes the chemical properties of red and white wine examples. The input features correspond to objective tests, for instance, acidity and pH tests. Originally, the only target was the sensory data (a human-based score, given by the median of three evaluations made by wine experts). Notwithstanding, for the purposes of evaluating a multi-output scenario, the fixed and volatile acidity, as well as the citric acid amounts were joined along with the quality score as new targets. Thus, the new task consists of predicting acidity levels and a quality score, modeling how those quantities relate to each other.

\newpage
\section{Time-varying observations for error, running time and model size}\label{sec_appendix_line_plots}

This appendix presents line plots for the observed errors, running time, and model size considering all the evaluated datasets.

\subsection{Measured error ($\overline{\text{RMSE}}$)}\label{sec_appendix_error_plots}

\begin{figure}[!htbp]
    \centering
    \subfloat[2DPlanes]{\includegraphics[width=0.5\textwidth]{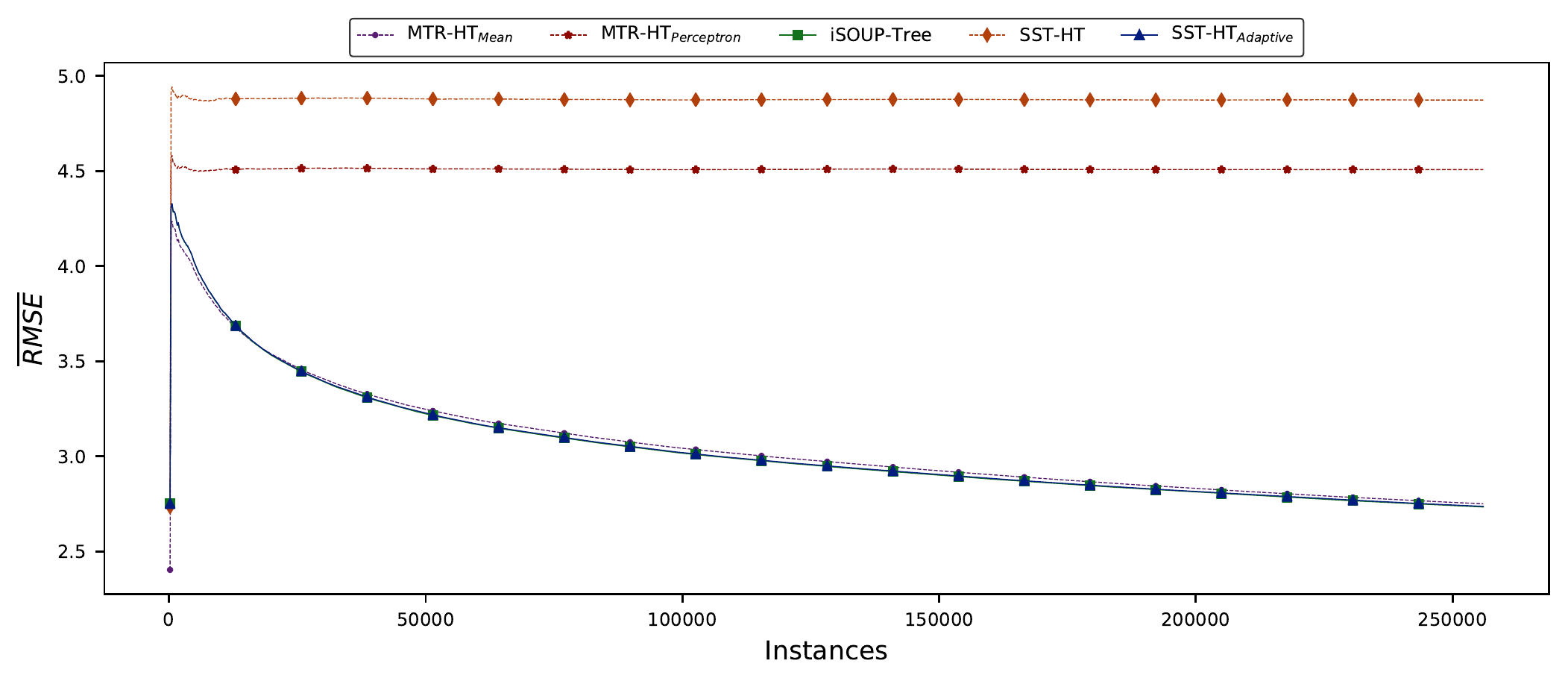}}
    \subfloat[Bicycles]{\includegraphics[width=0.5\textwidth]{line_bicycles_mean_armse_M0}}
    
    \subfloat[CPU]{\includegraphics[width=0.5\textwidth]{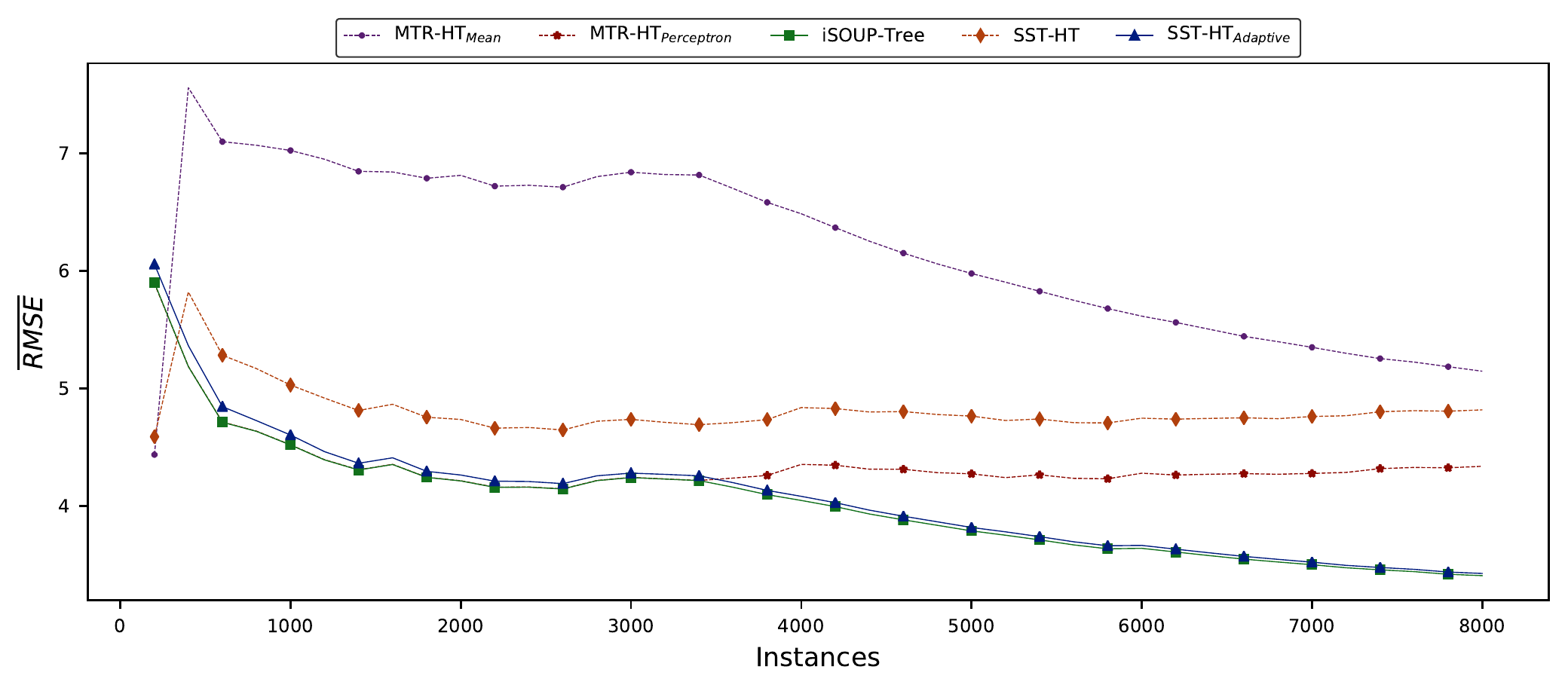}}
    \subfloat[Electricity]{\includegraphics[width=0.5\textwidth]{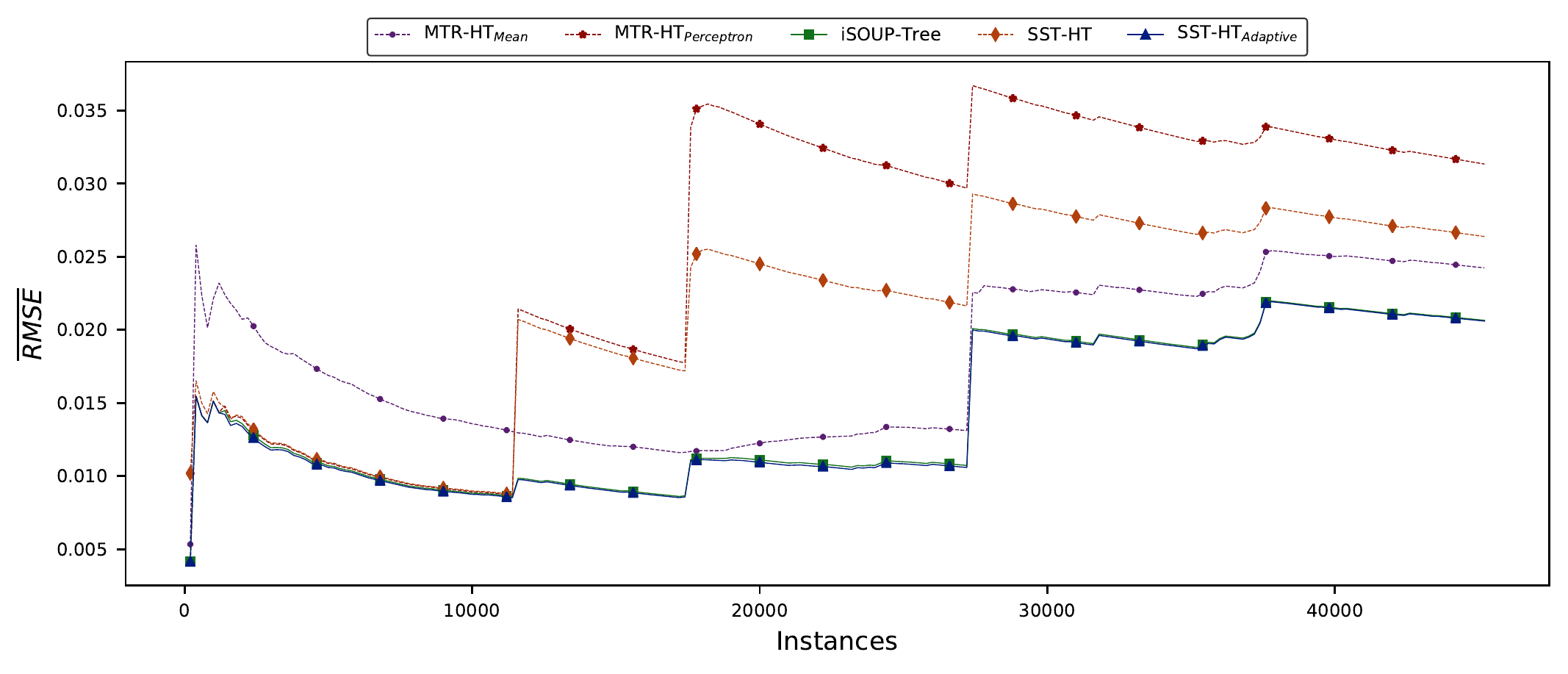}}
    
    \subfloat[Eunite03]{\includegraphics[width=0.5\textwidth]{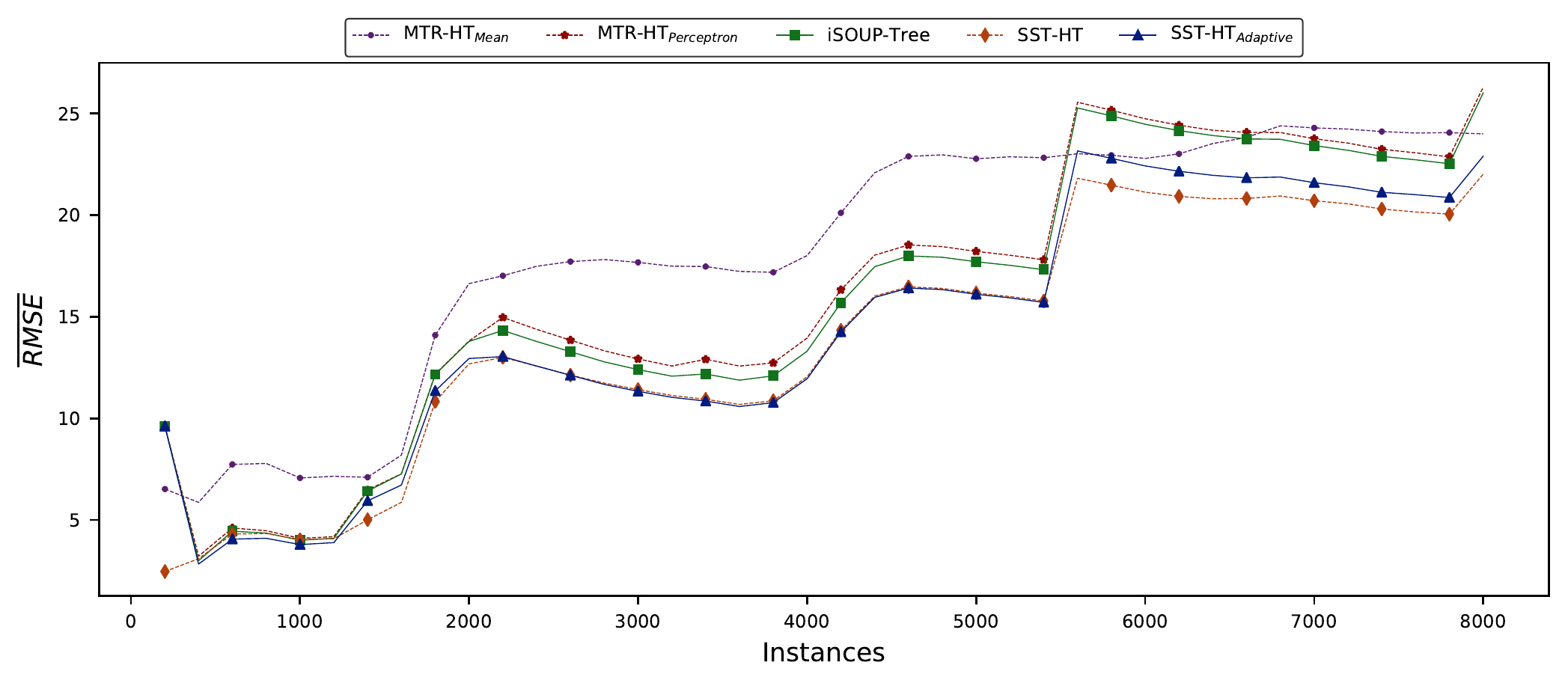}}
    \subfloat[FriedD]{\includegraphics[width=0.5\textwidth]{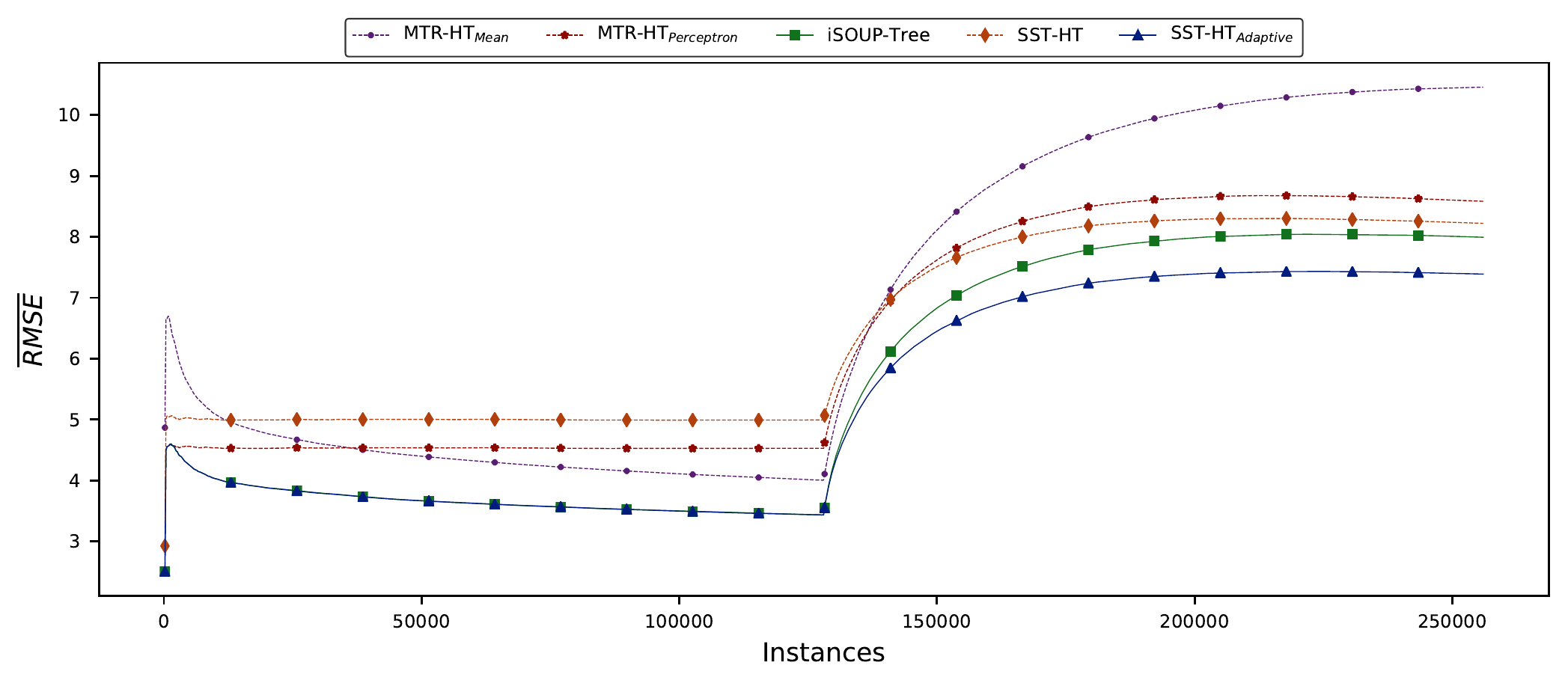}}
    
    \subfloat[FriedAsyncD]{\includegraphics[width=0.5\textwidth]{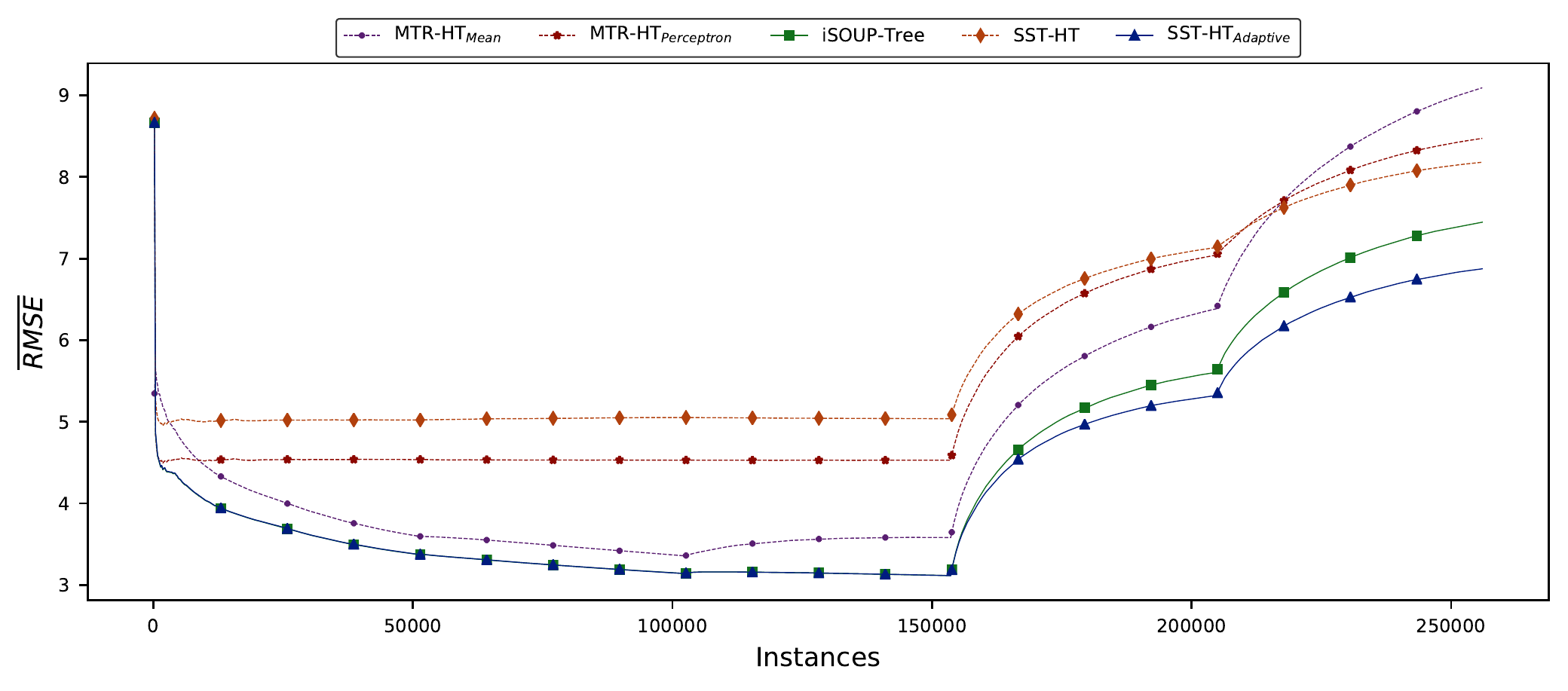}}
    \subfloat[MV]{\includegraphics[width=0.5\textwidth]{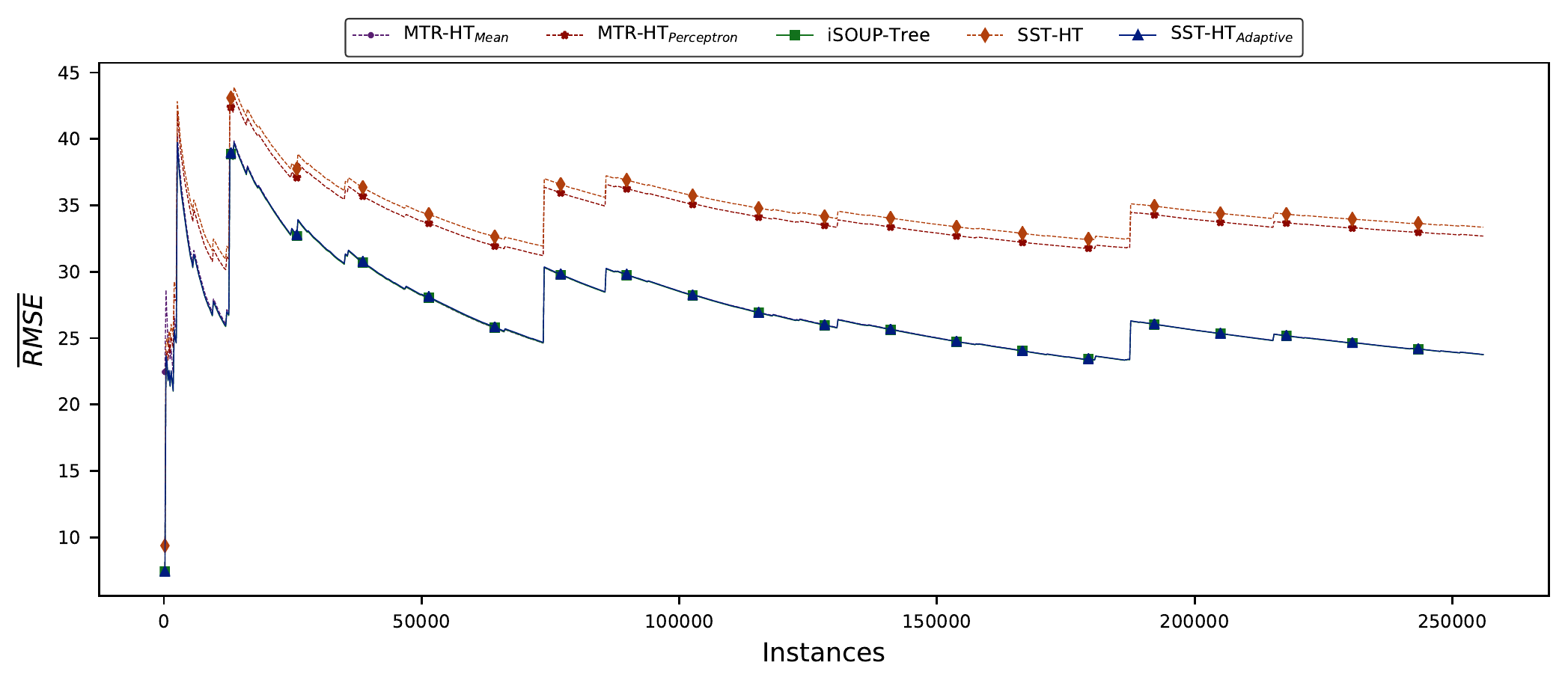}}
    \caption{Time varying results for the measured $\overline{\text{RMSE}}$ values}
\end{figure}

\begin{figure}[!htbp]\ContinuedFloat
    \centering
    \subfloat[NPSDecay]{\includegraphics[width=0.5\textwidth]{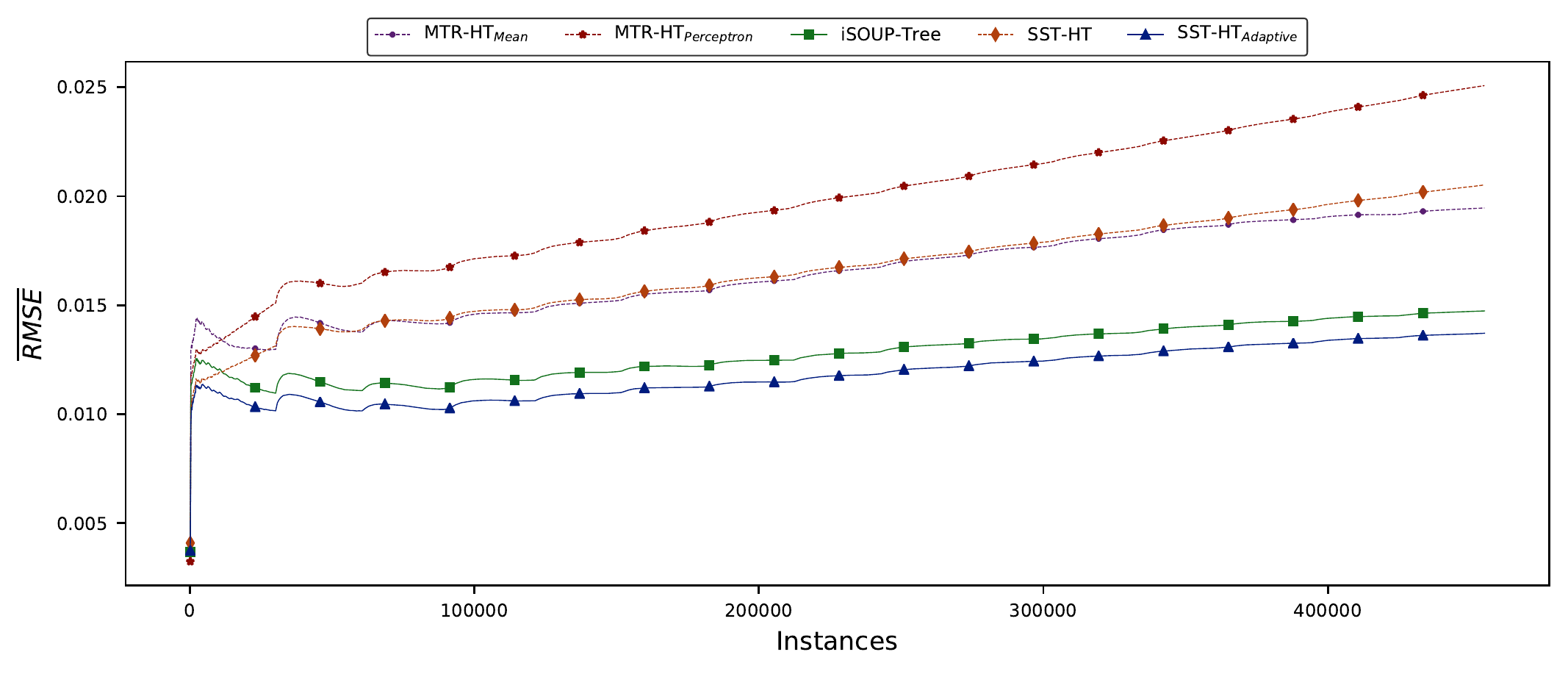}}
    \subfloat[RF1]{\includegraphics[width=0.5\textwidth]{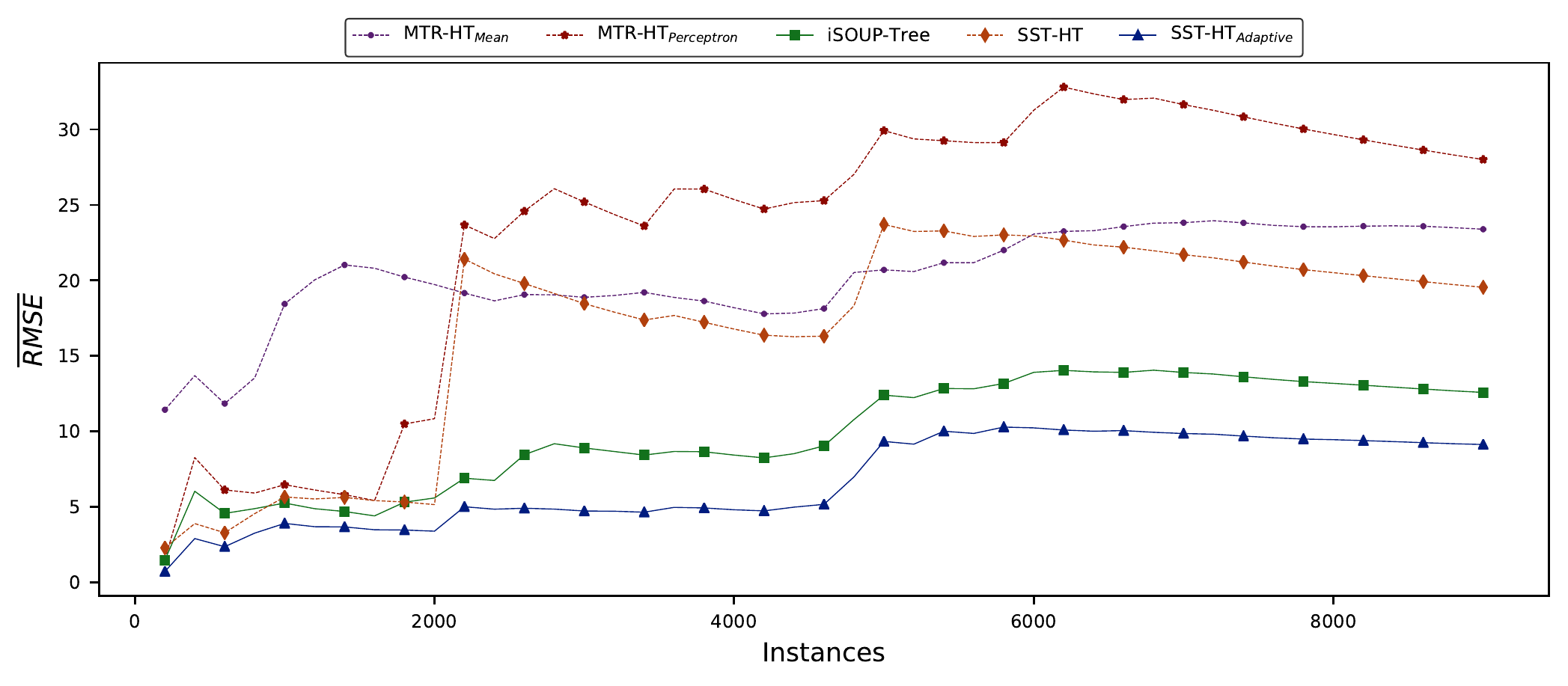}}
    
    \subfloat[RF2]{\includegraphics[width=0.5\textwidth]{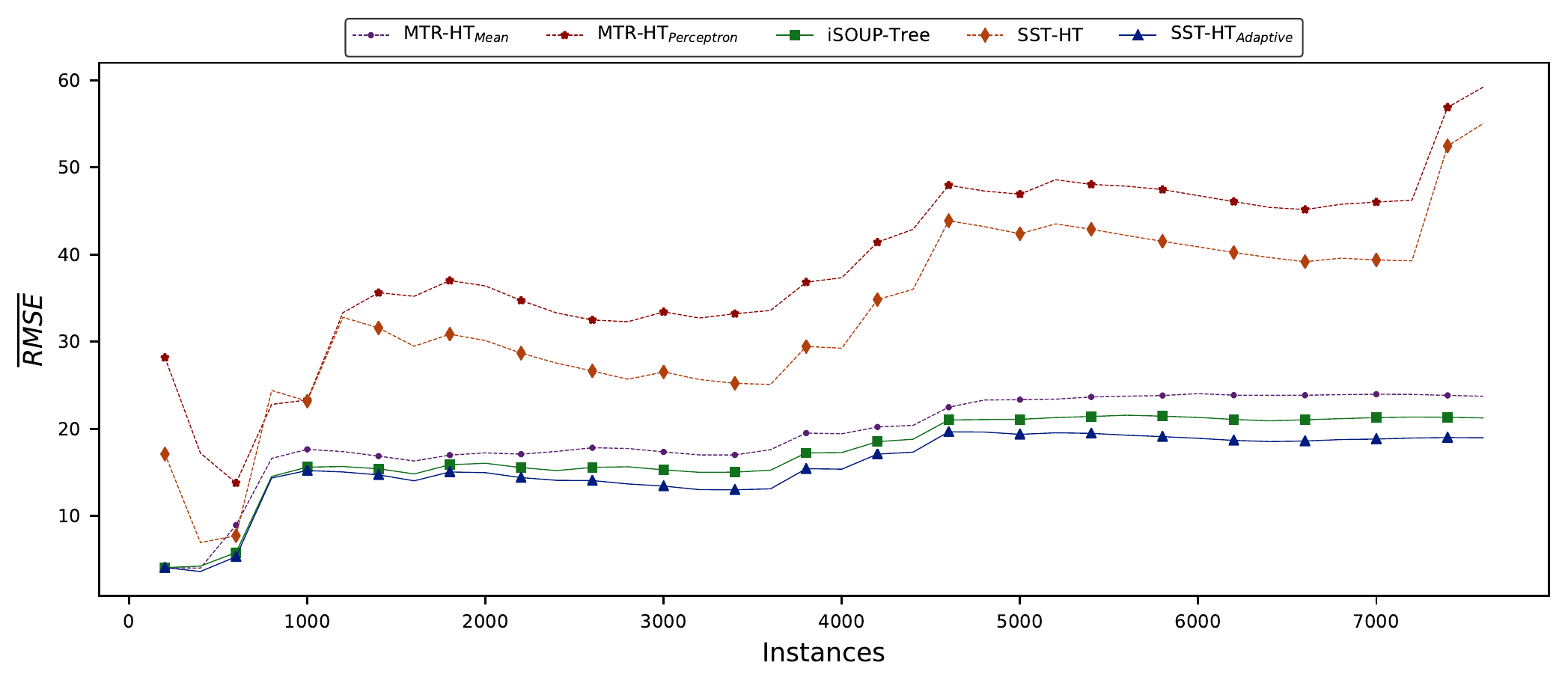}}
    \subfloat[SCFP]{\includegraphics[width=0.5\textwidth]{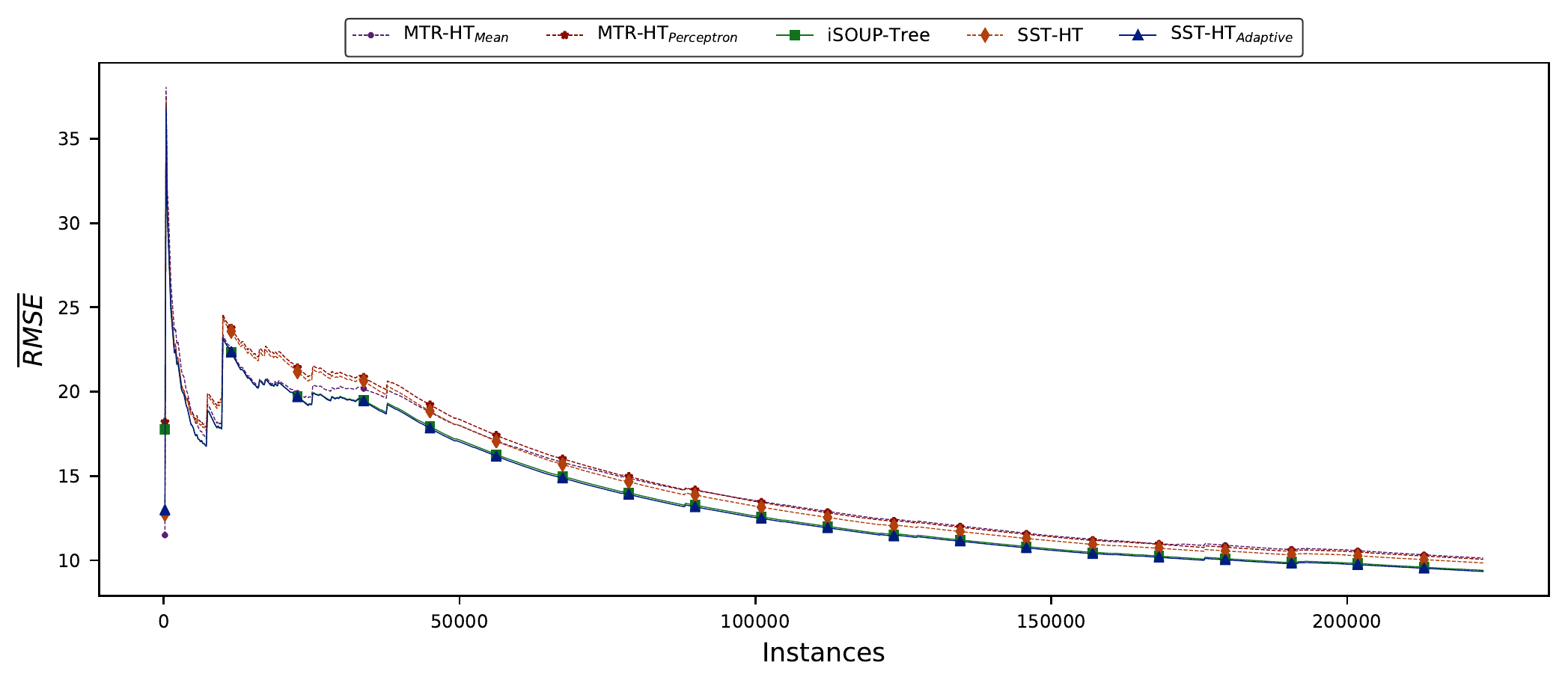}}
    
    \subfloat[SCM1d]{\includegraphics[width=0.5\textwidth]{line_scm1d_mean_armse_M0}}
    \subfloat[SCM20d]{\includegraphics[width=0.5\textwidth]{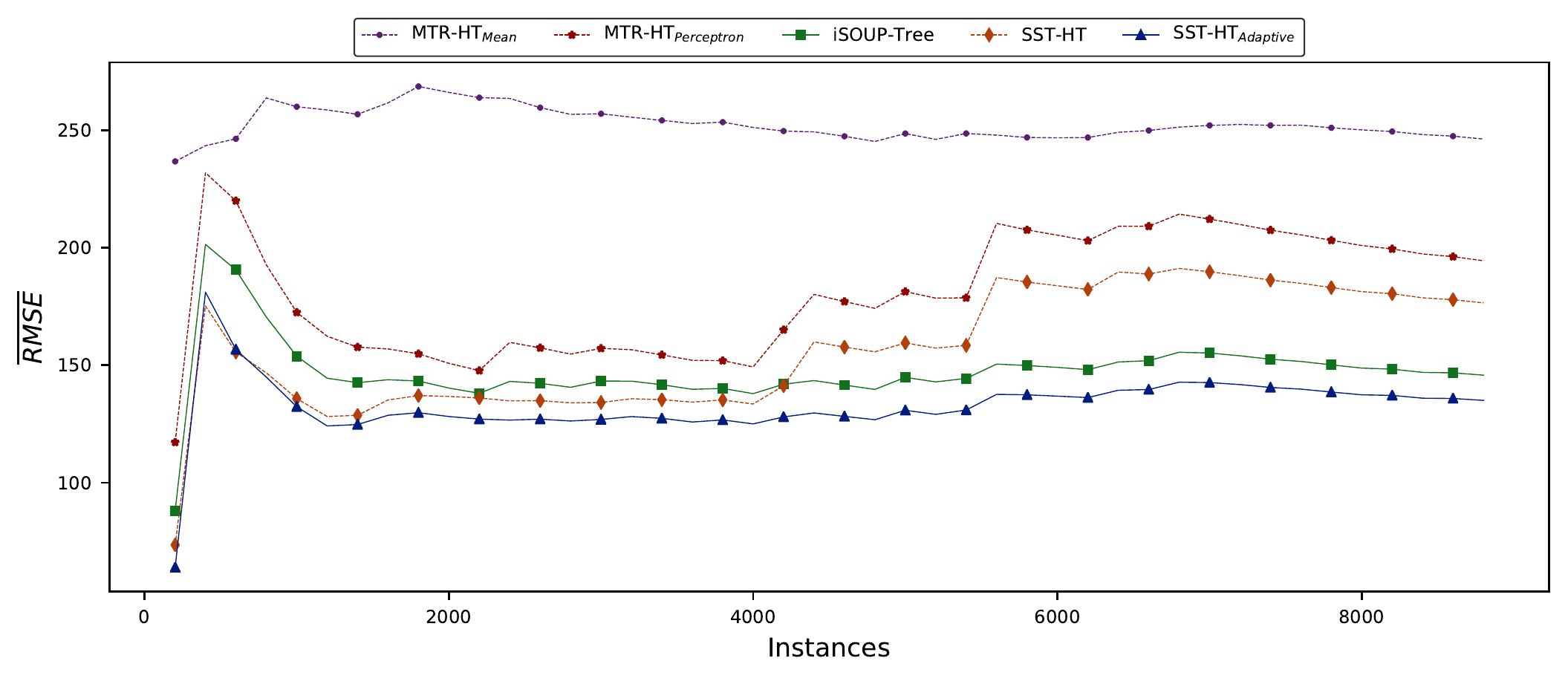}}
    
    \subfloat[Sulfur]{\includegraphics[width=0.5\textwidth]{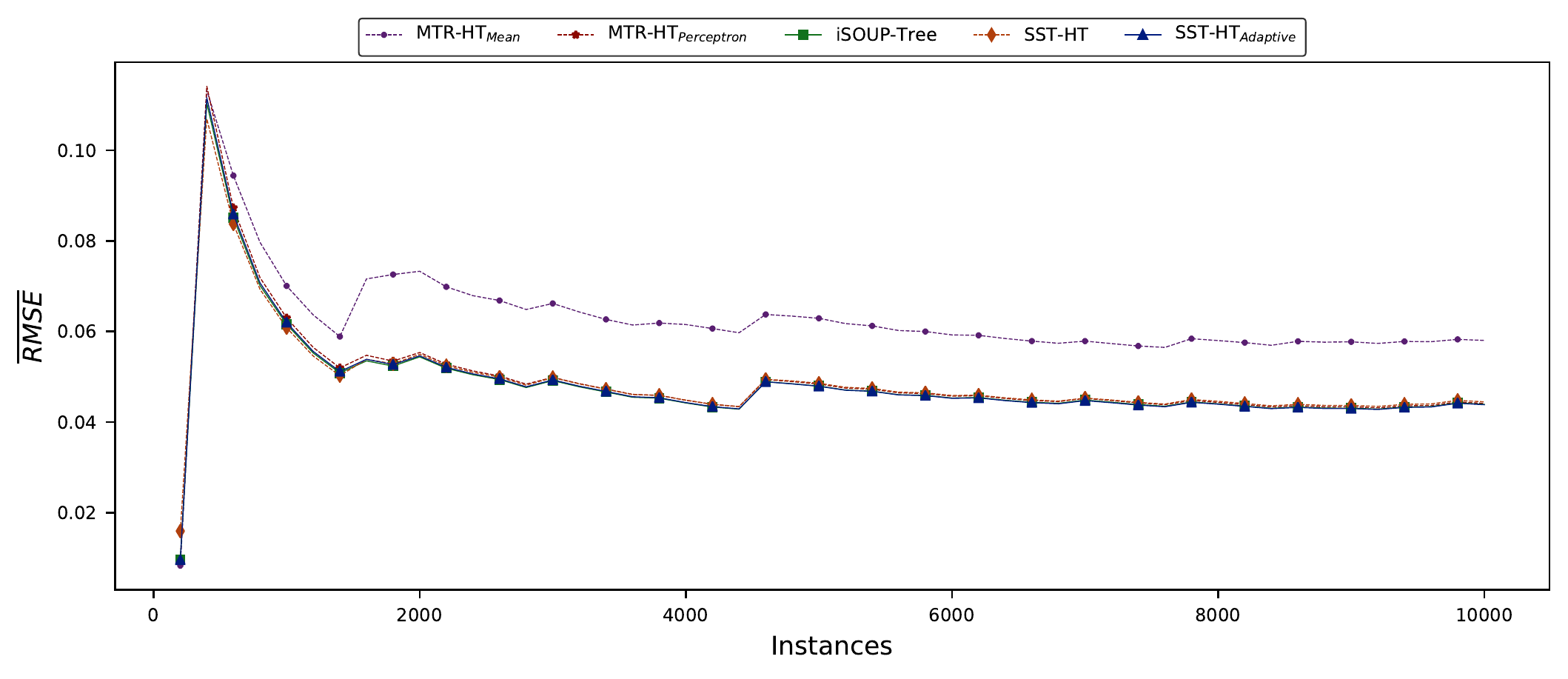}}
    \subfloat[Wine]{\includegraphics[width=0.5\textwidth]{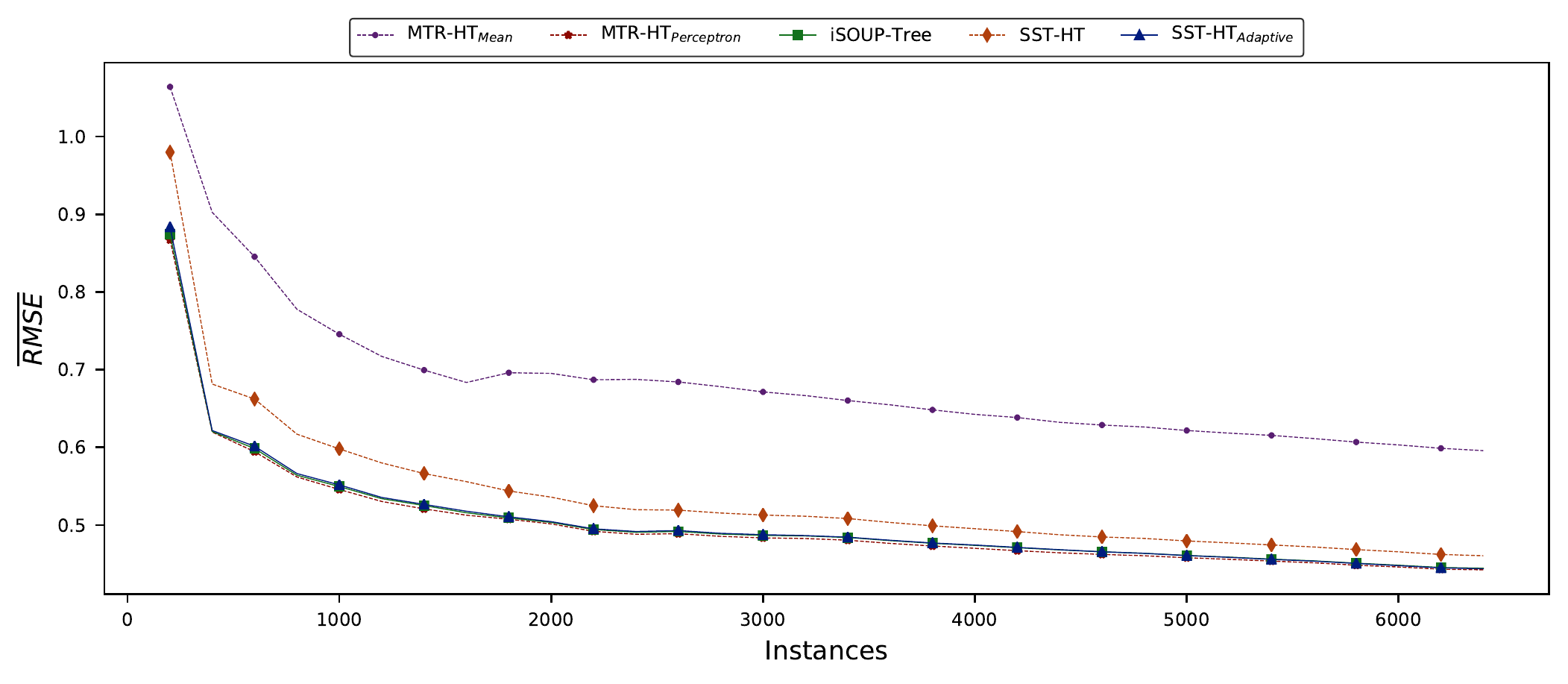}}
    
    \caption{Time varying results for the measured $\overline{\text{RMSE}}$ values (continuation)}
    \label{fig_error_lines}
\end{figure}

\clearpage
\subsection{Running time}\label{sec_appendix_time_plots}

\begin{figure}[!htbp]
    \centering
    \subfloat[2DPlanes]{\includegraphics[width=0.5\textwidth]{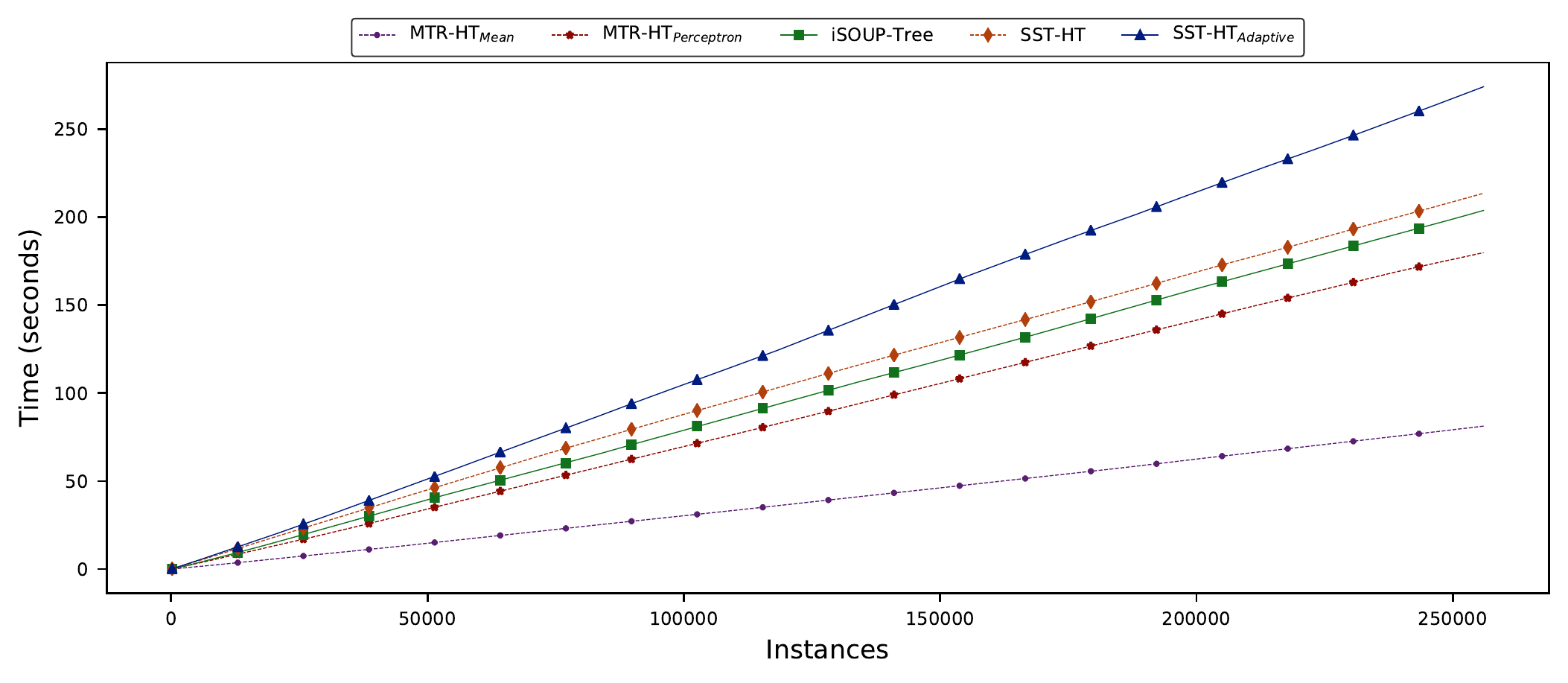}}
    \subfloat[Bicycles]{\includegraphics[width=0.5\textwidth]{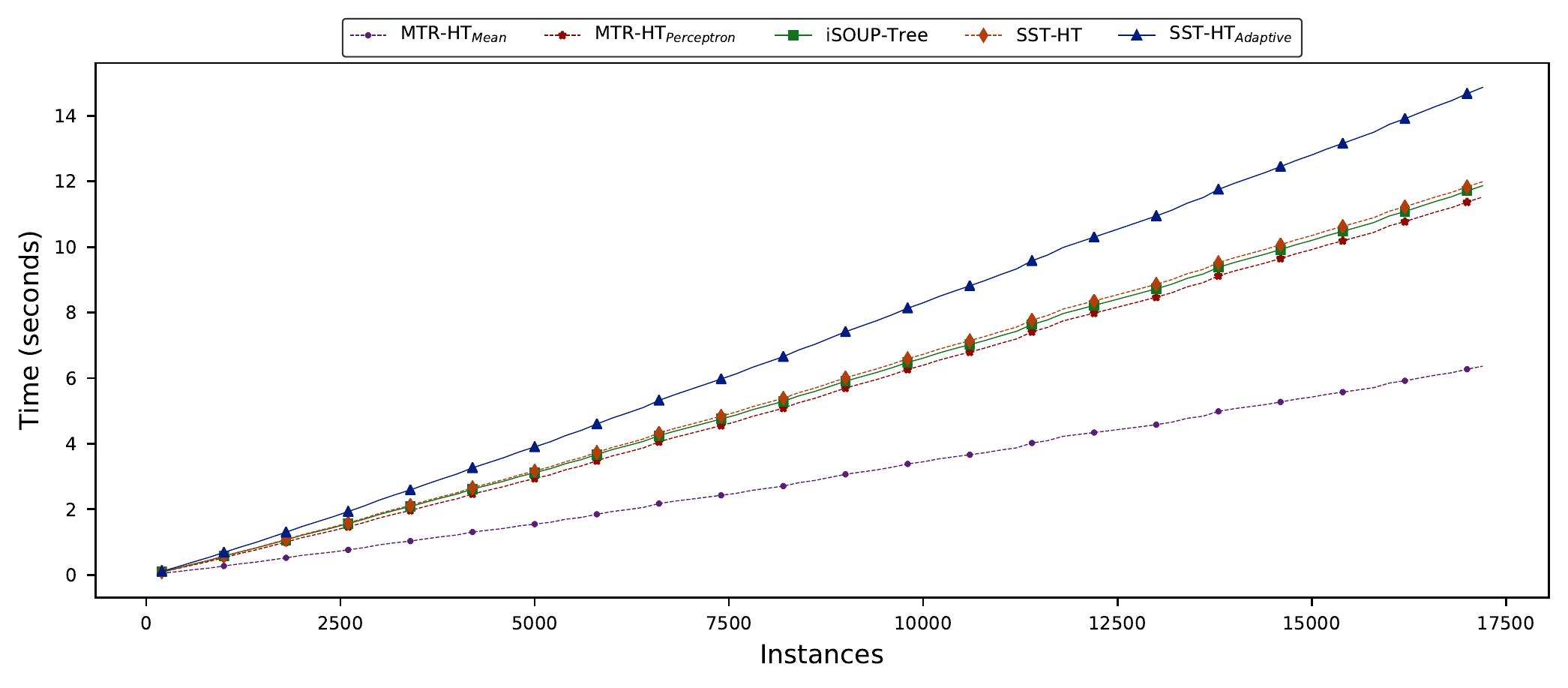}}
    
    \subfloat[CPU]{\includegraphics[width=0.5\textwidth]{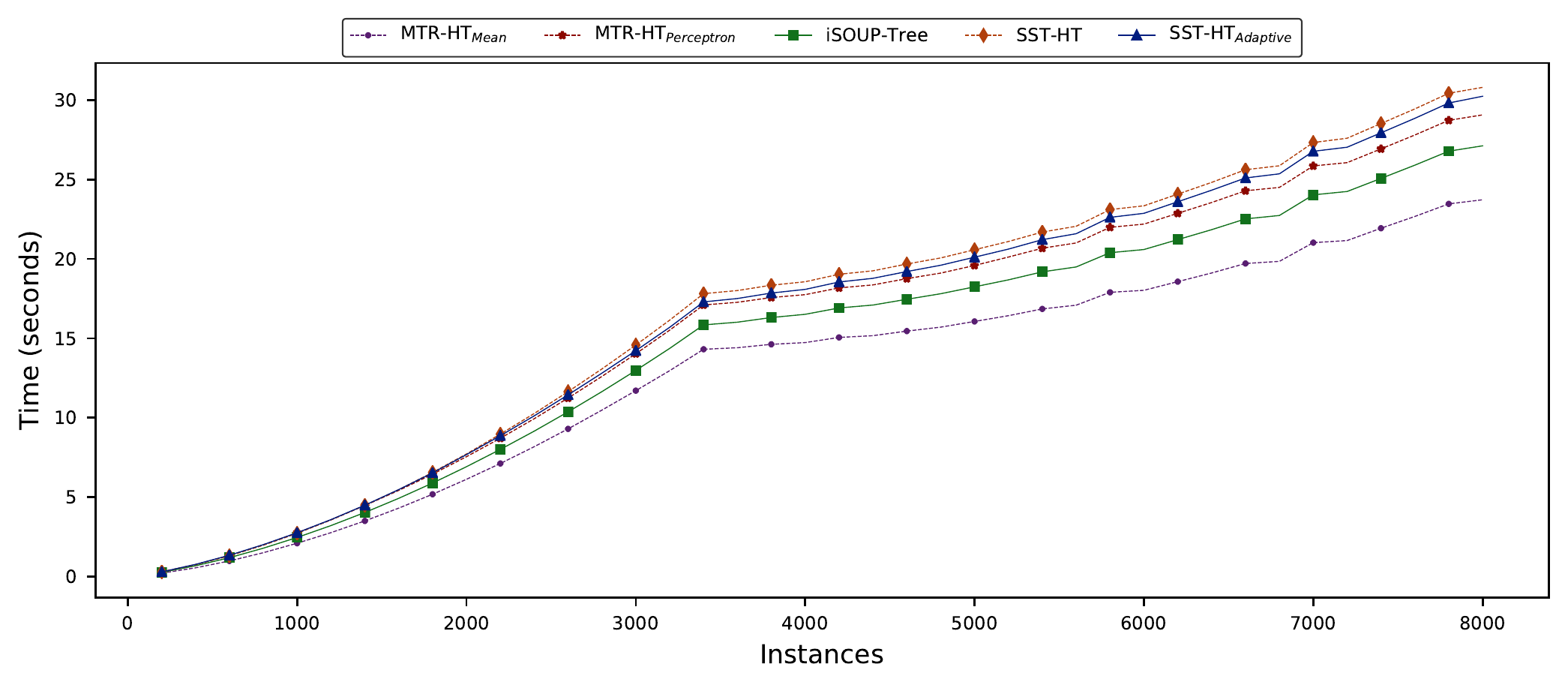}}
    \subfloat[Electricity]{\includegraphics[width=0.5\textwidth]{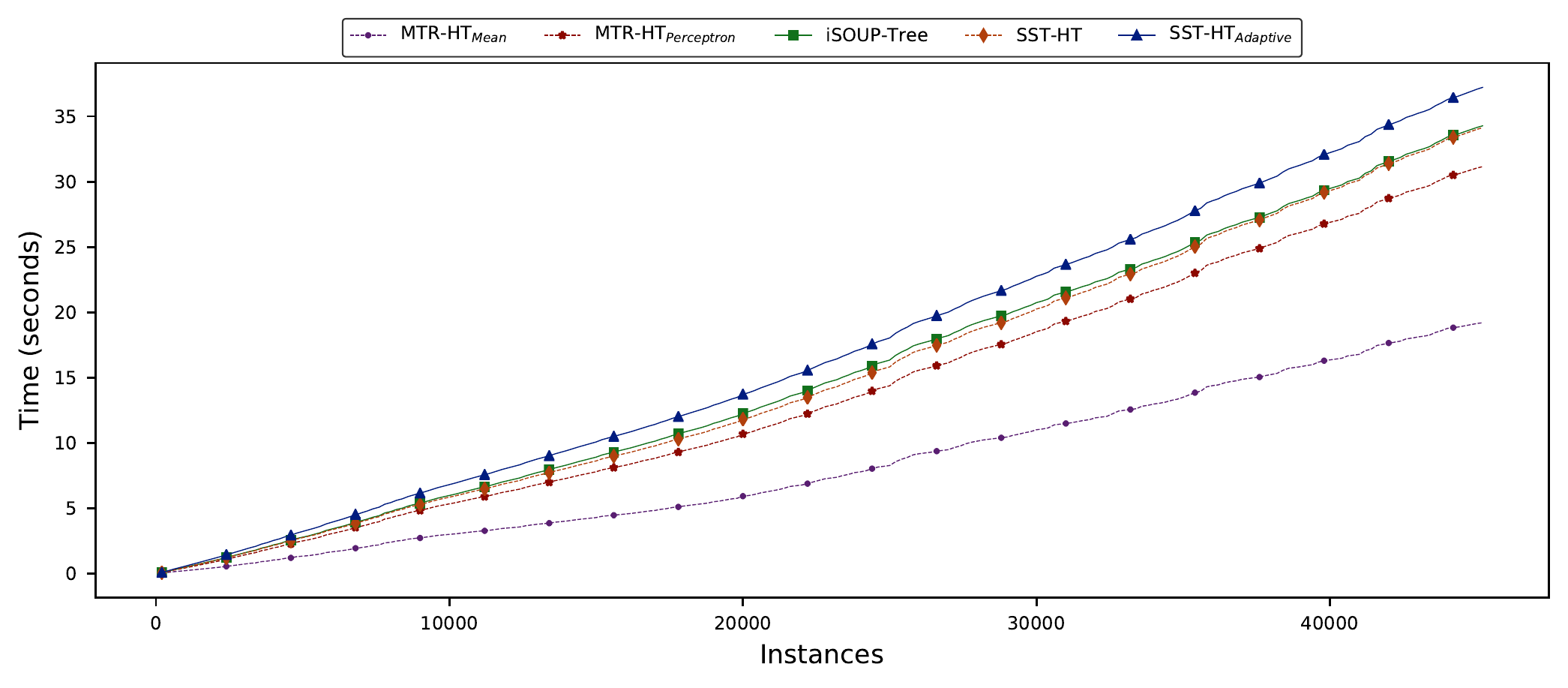}}
    
    \subfloat[Eunite03]{\includegraphics[width=0.5\textwidth]{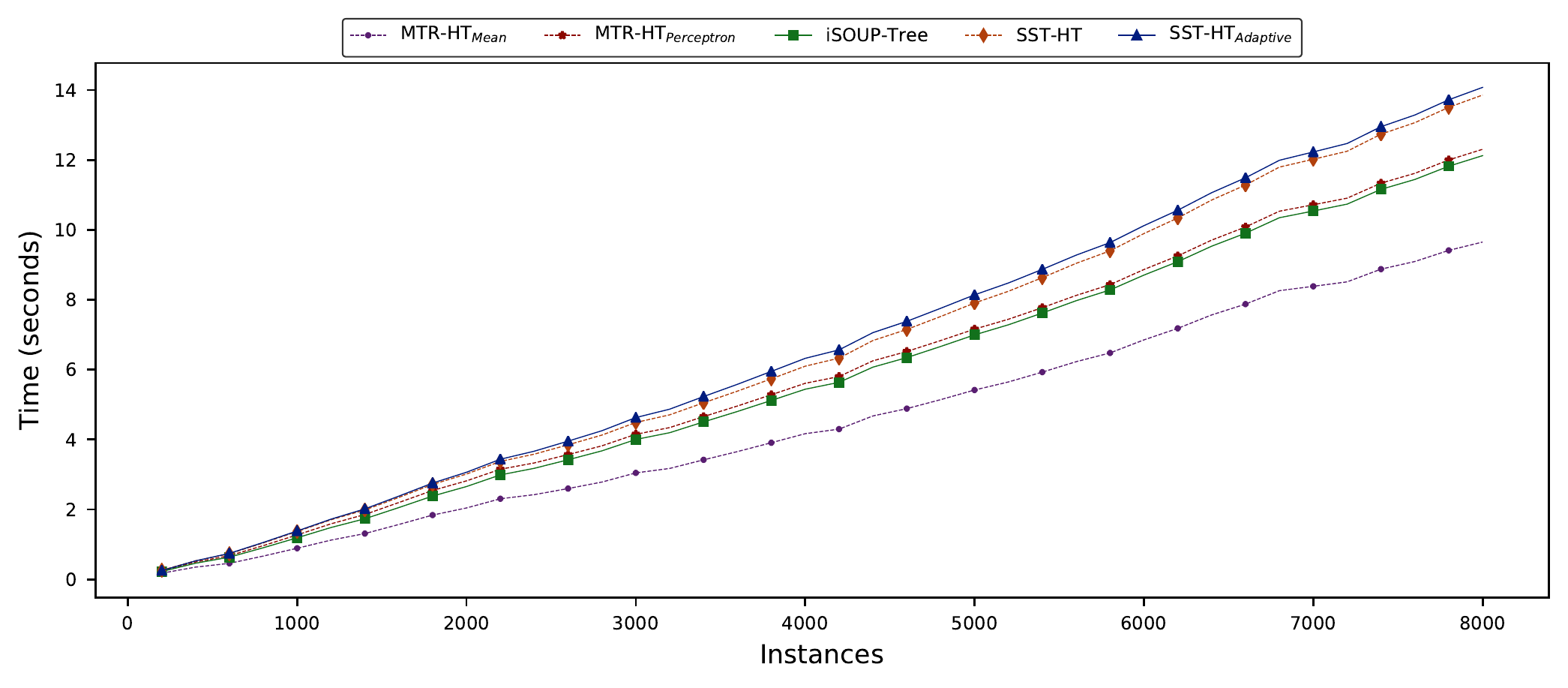}}
    \subfloat[FriedD]{\includegraphics[width=0.5\textwidth]{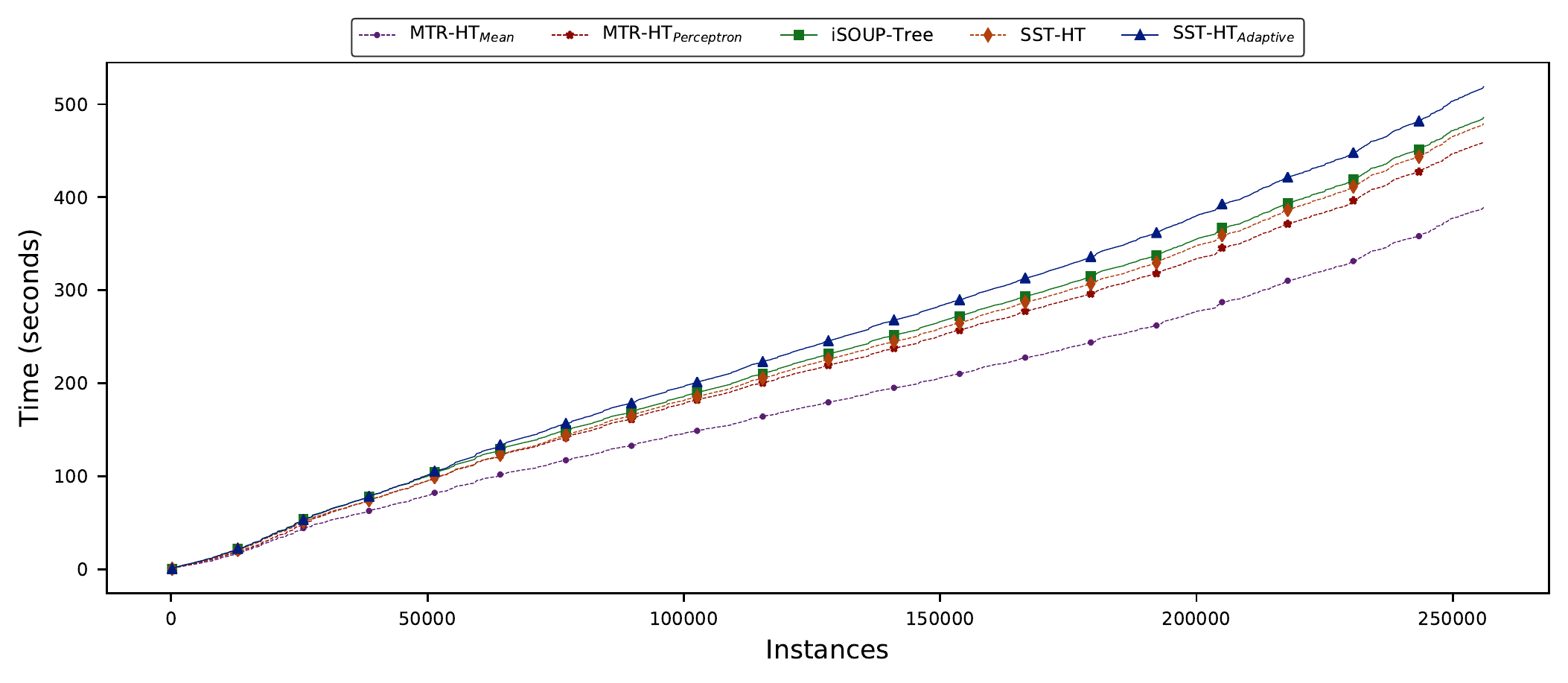}}

    \subfloat[FriedAsyncD]{\includegraphics[width=0.5\textwidth]{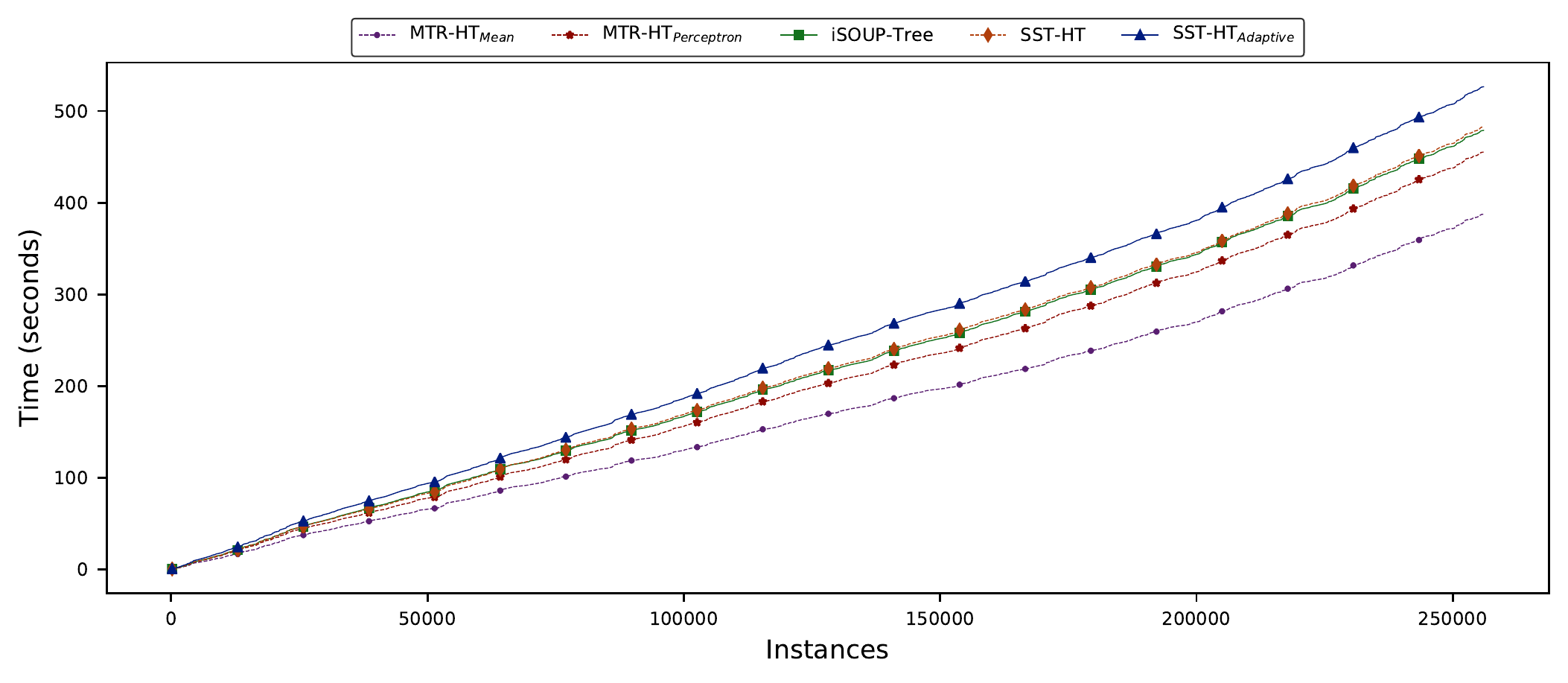}}
    \subfloat[MV]{\includegraphics[width=0.5\textwidth]{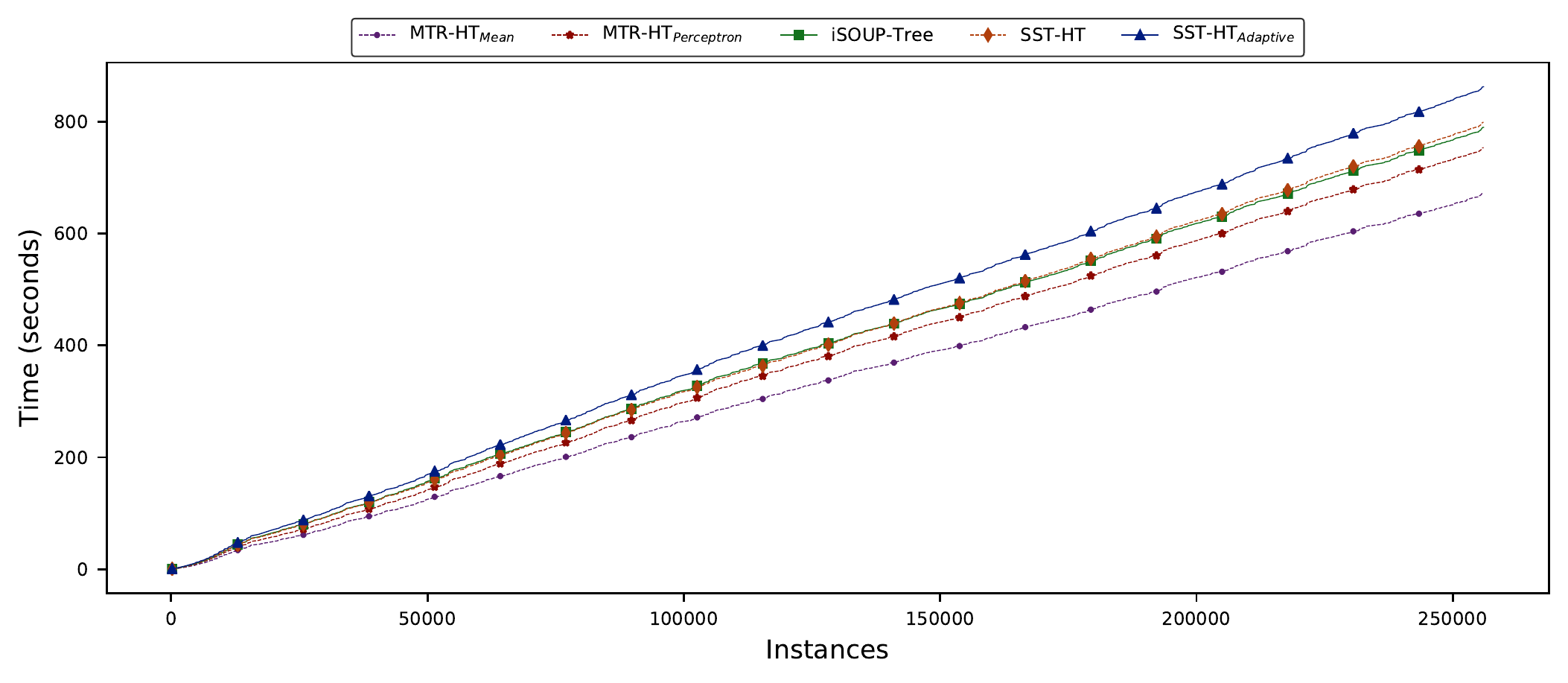}}
    
    \subfloat[NPSDecay]{\includegraphics[width=0.5\textwidth]{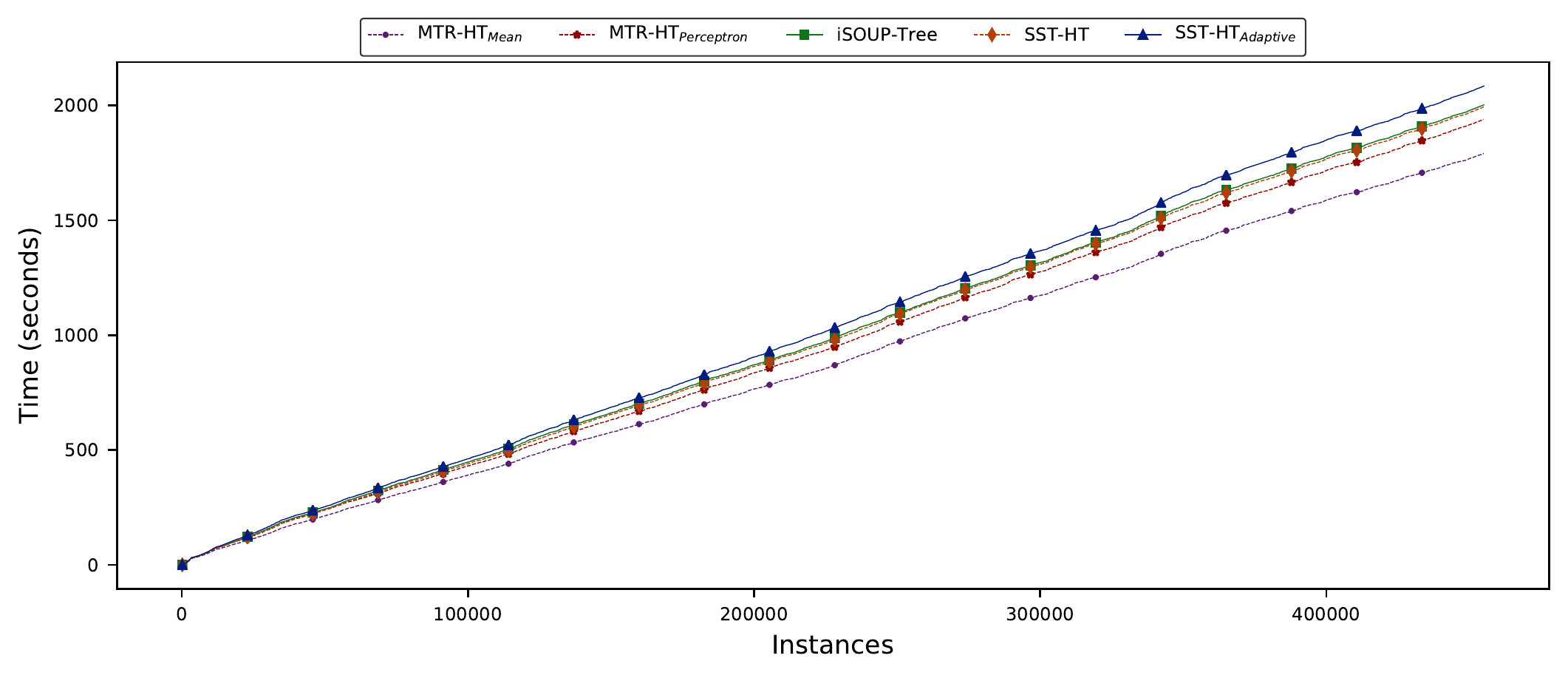}}
    \subfloat[RF1]{\includegraphics[width=0.5\textwidth]{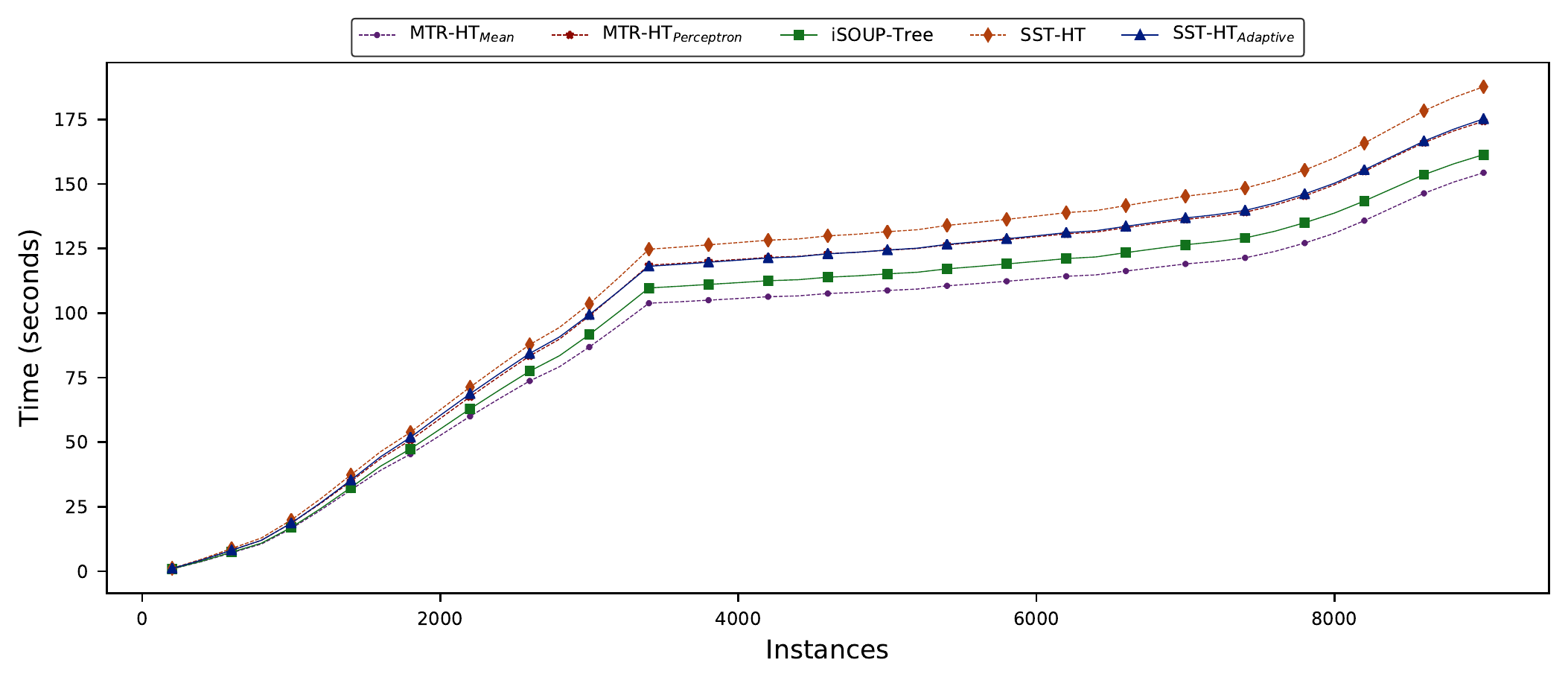}}
    \caption{Accounted running times for all the evaluated datasets}
\end{figure}

\begin{figure}[!htbp]\ContinuedFloat
    \centering
    \subfloat[RF2]{\includegraphics[width=0.5\textwidth]{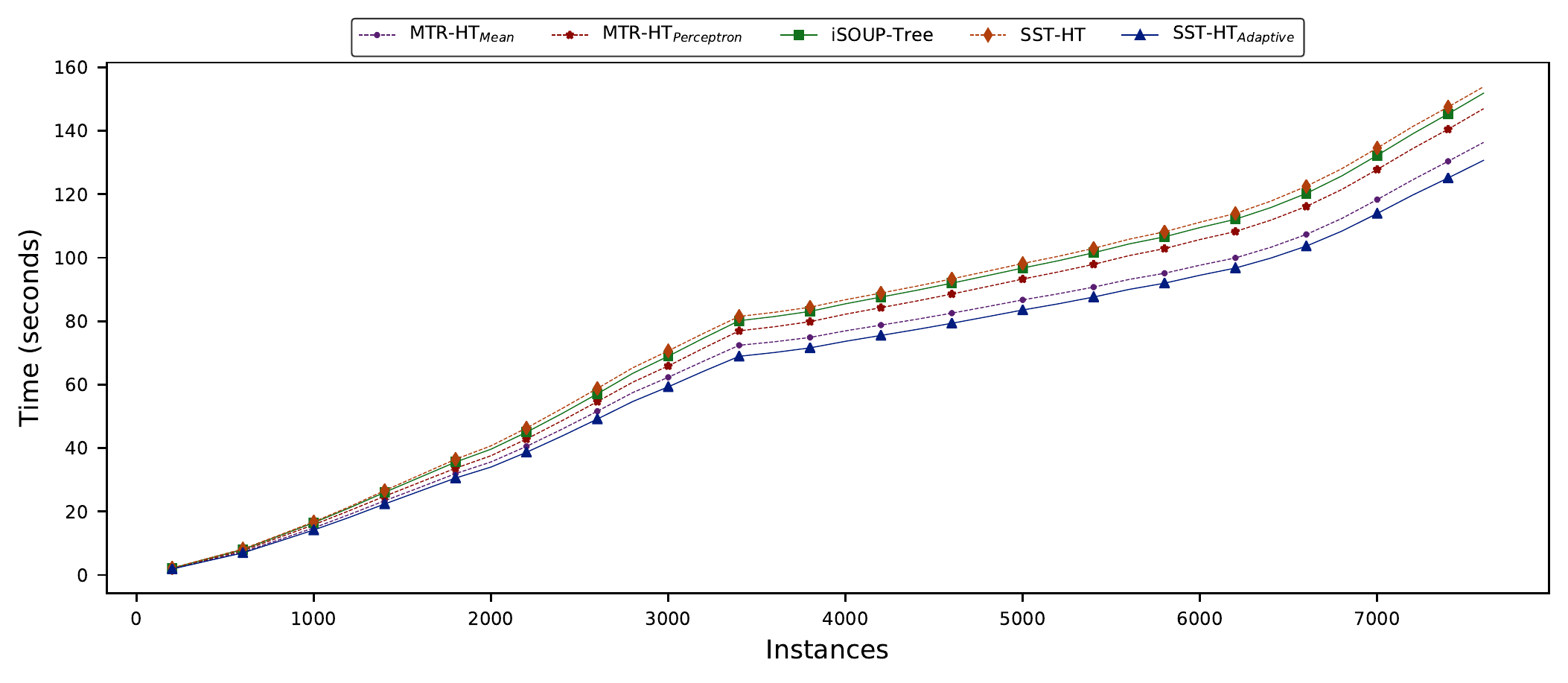}}
    \subfloat[SCFP]{\includegraphics[width=0.5\textwidth]{line_scfp_complete_total_running_time_M0}}

    \subfloat[SCM1d]{\includegraphics[width=0.5\textwidth]{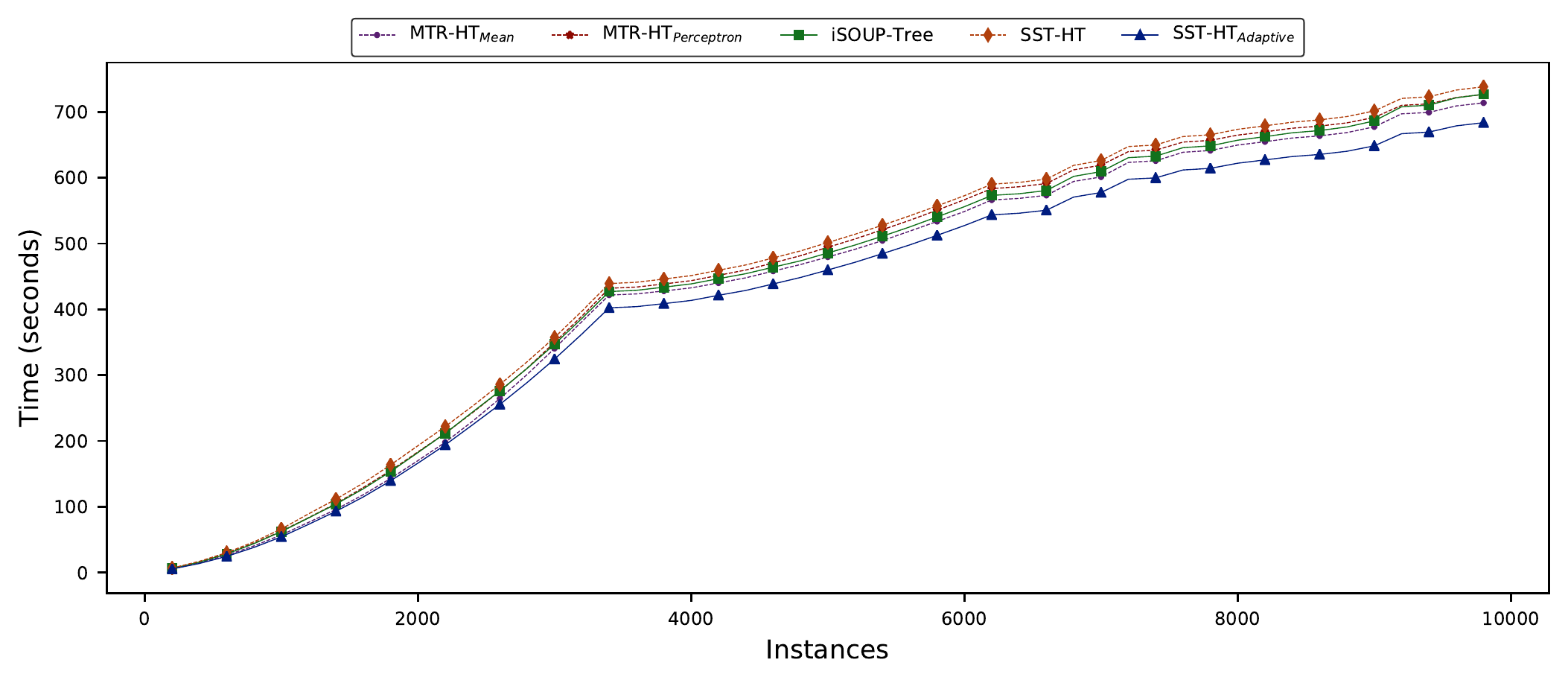}}
    \subfloat[SCM20d]{\includegraphics[width=0.5\textwidth]{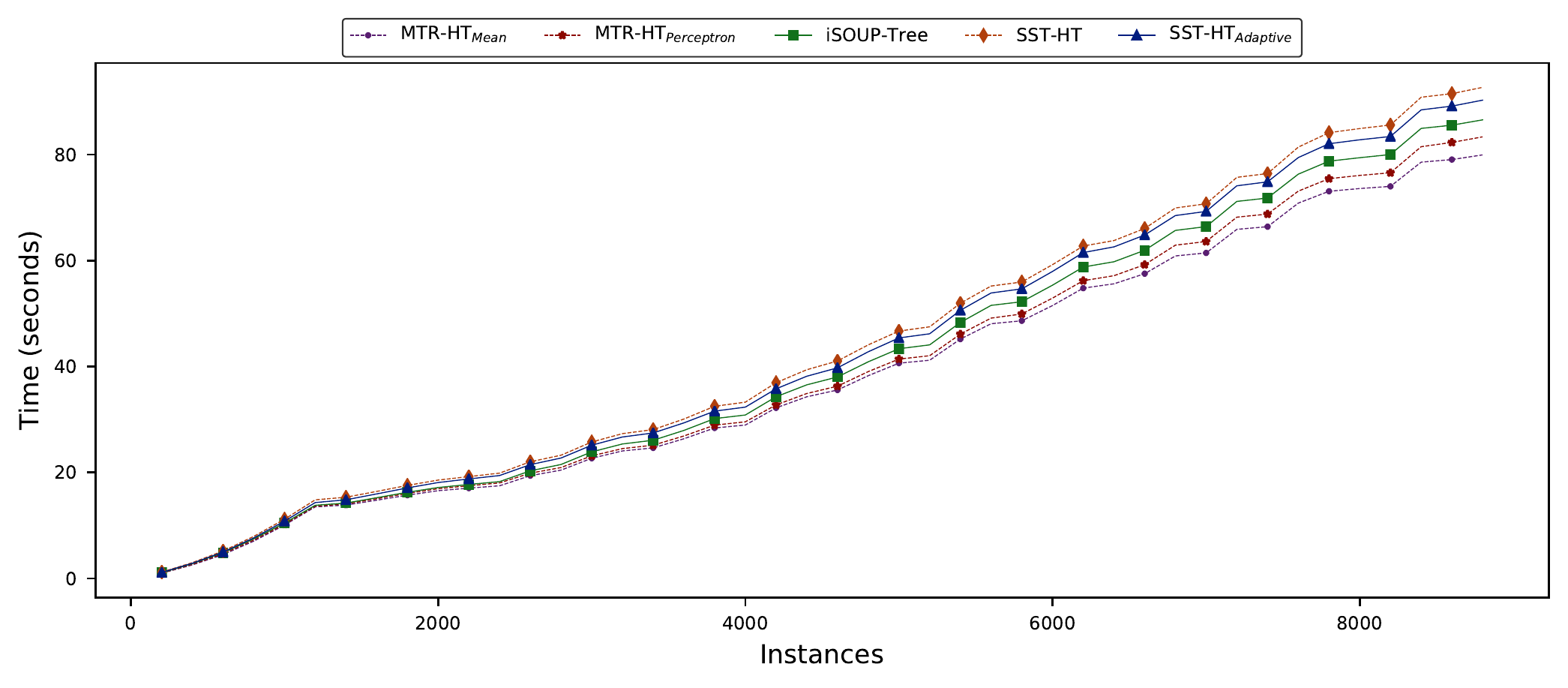}}
    
    \subfloat[Sulfur]{\includegraphics[width=0.5\textwidth]{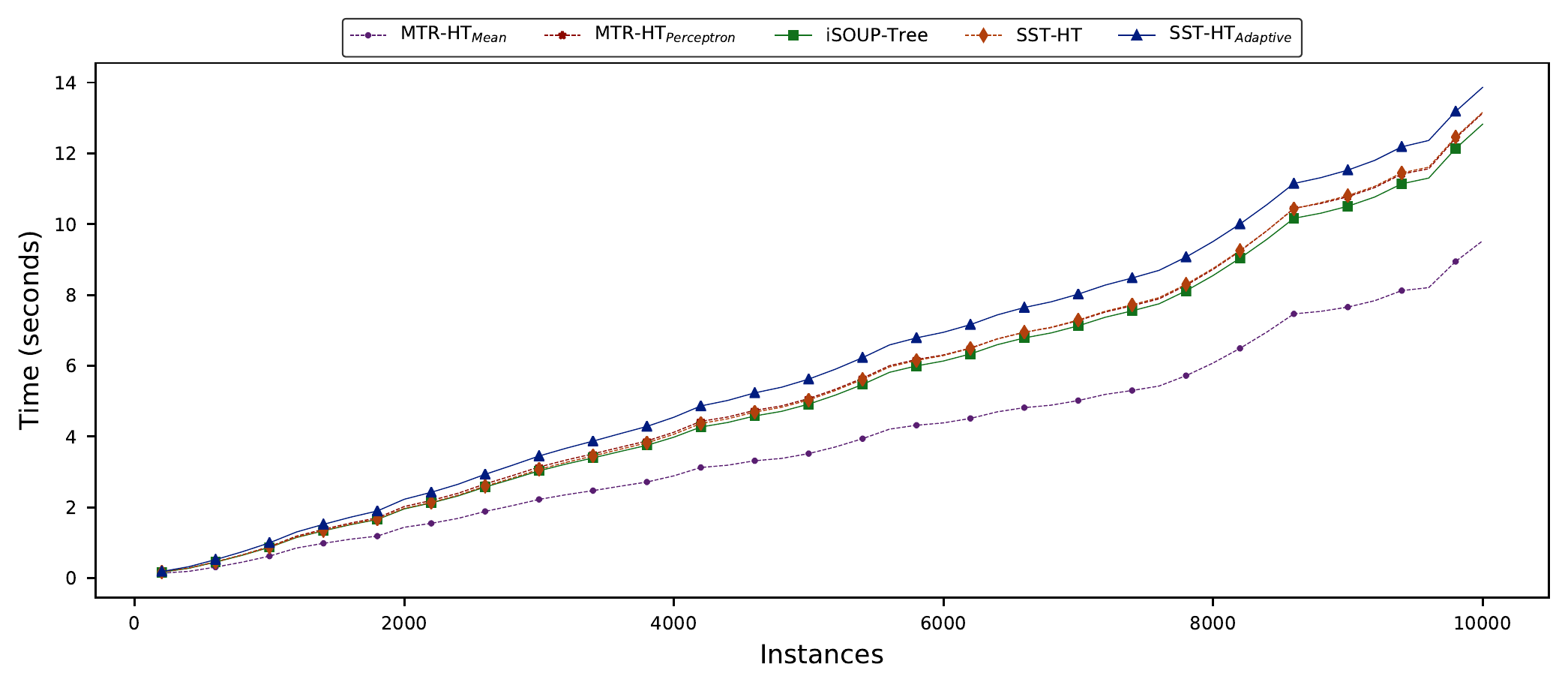}}
    \subfloat[Wine]{\includegraphics[width=0.5\textwidth]{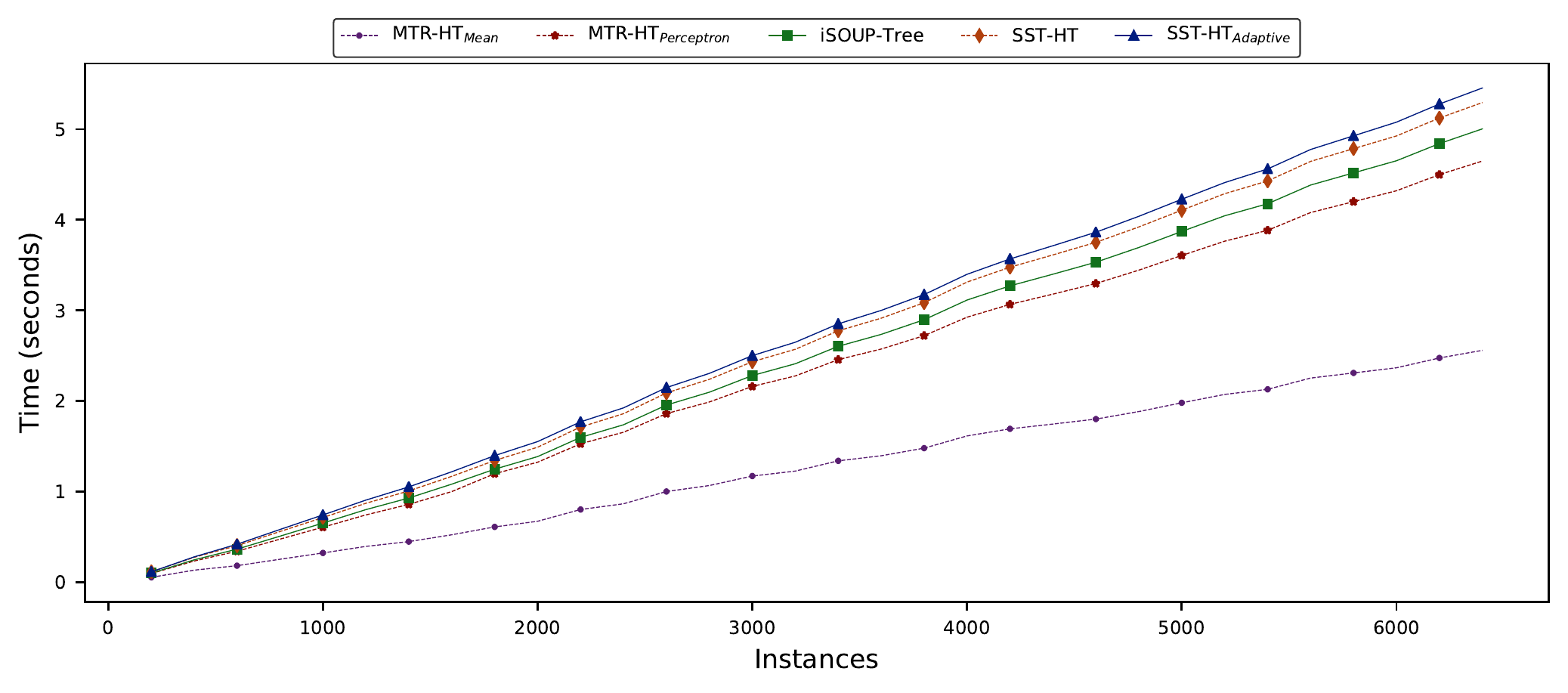}}
    
    \caption{Accounted running times for all the evaluated datasets (continuation)}
    \label{fig_time_lines}
\end{figure}

\clearpage
\subsection{Model size}\label{sec_appendix_memory_plots}

\begin{figure}[!htbp]
    \centering
    \subfloat[2DPlanes]{\includegraphics[width=0.5\textwidth]{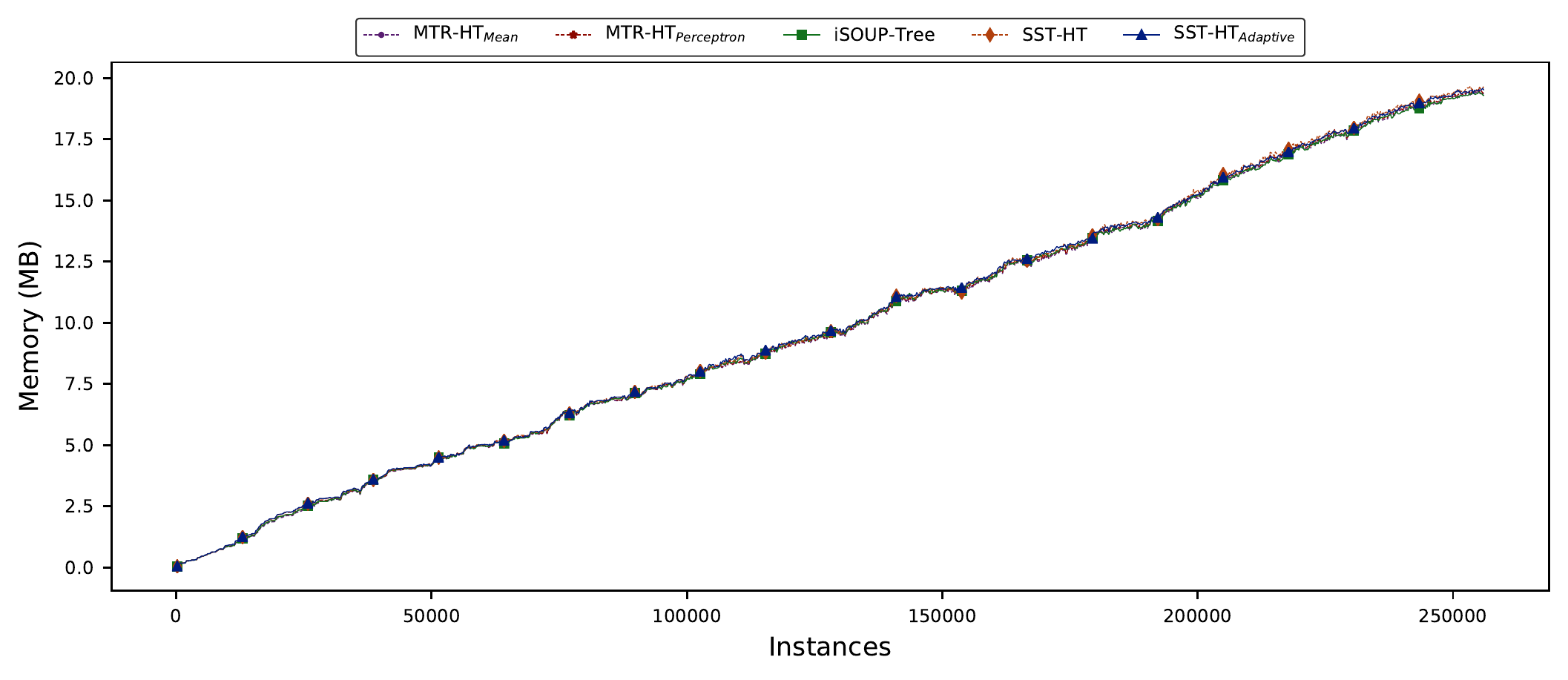}}
    \subfloat[Bicycles]{\includegraphics[width=0.5\textwidth]{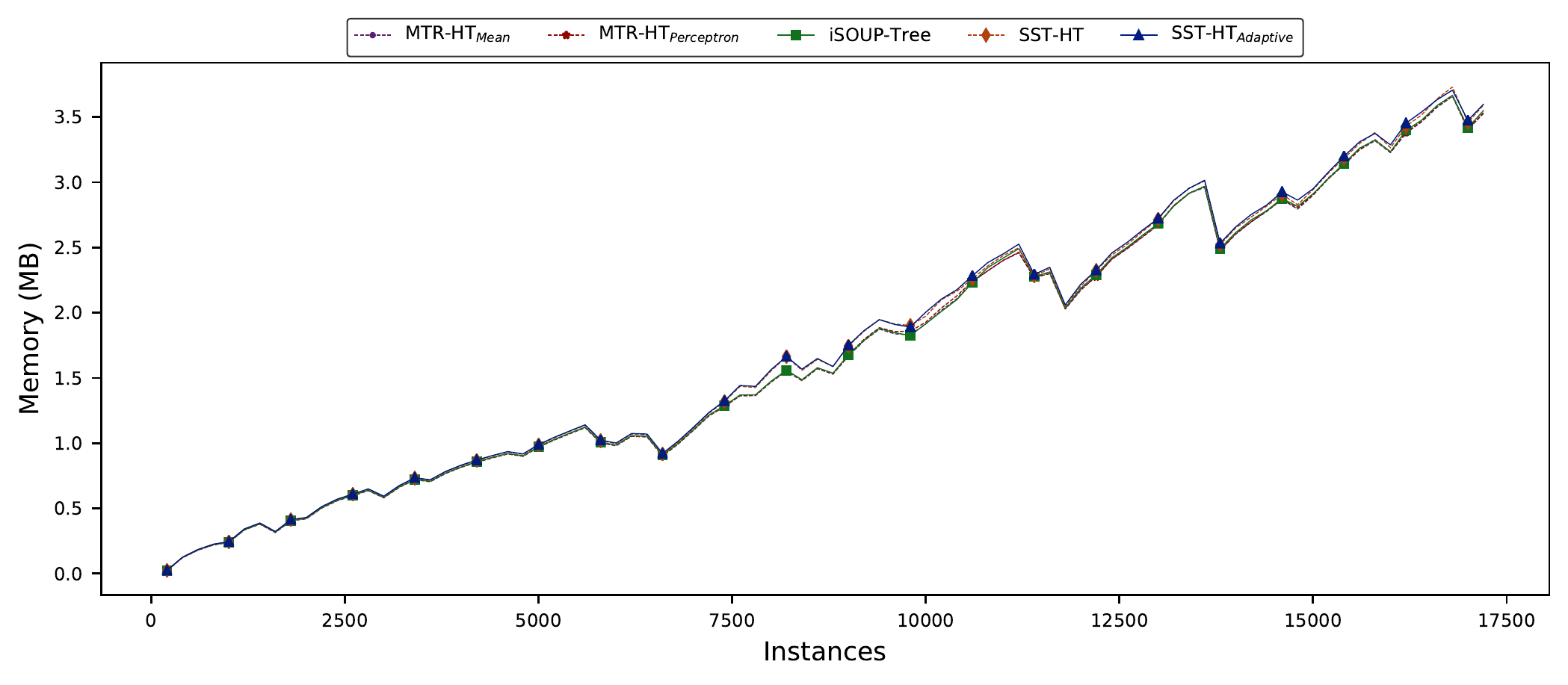}}

    \subfloat[CPU]{\includegraphics[width=0.5\textwidth]{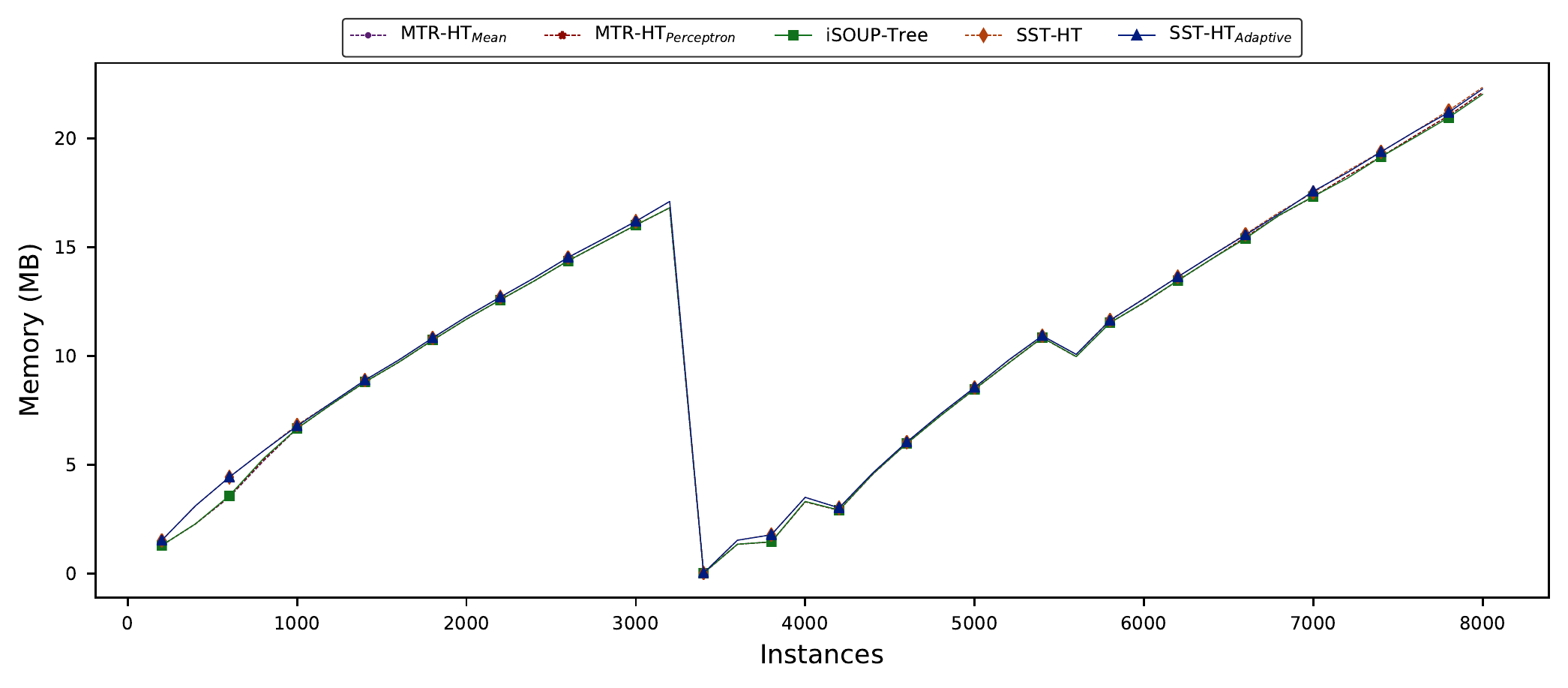}}
    \subfloat[Electricity]{\includegraphics[width=0.5\textwidth]{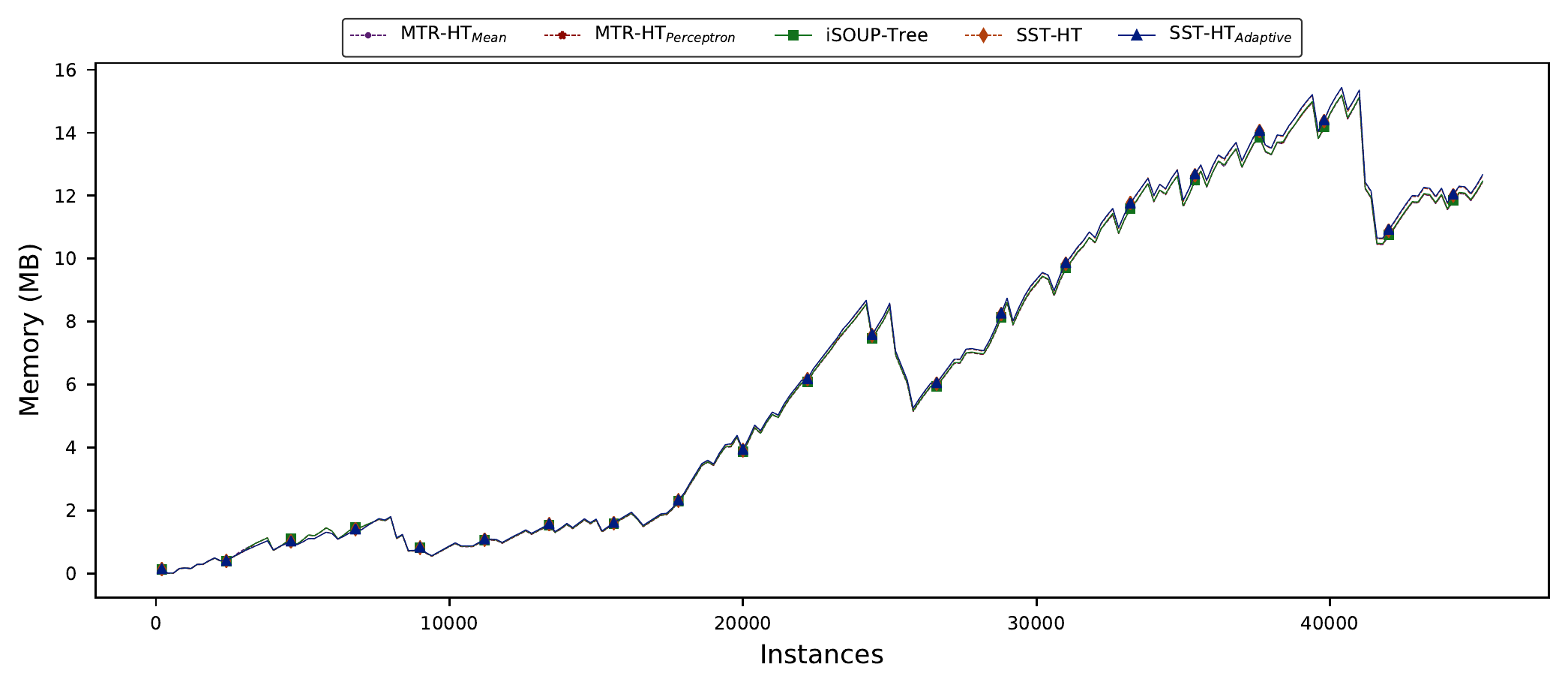}}
    
    \subfloat[Eunite03]{\includegraphics[width=0.5\textwidth]{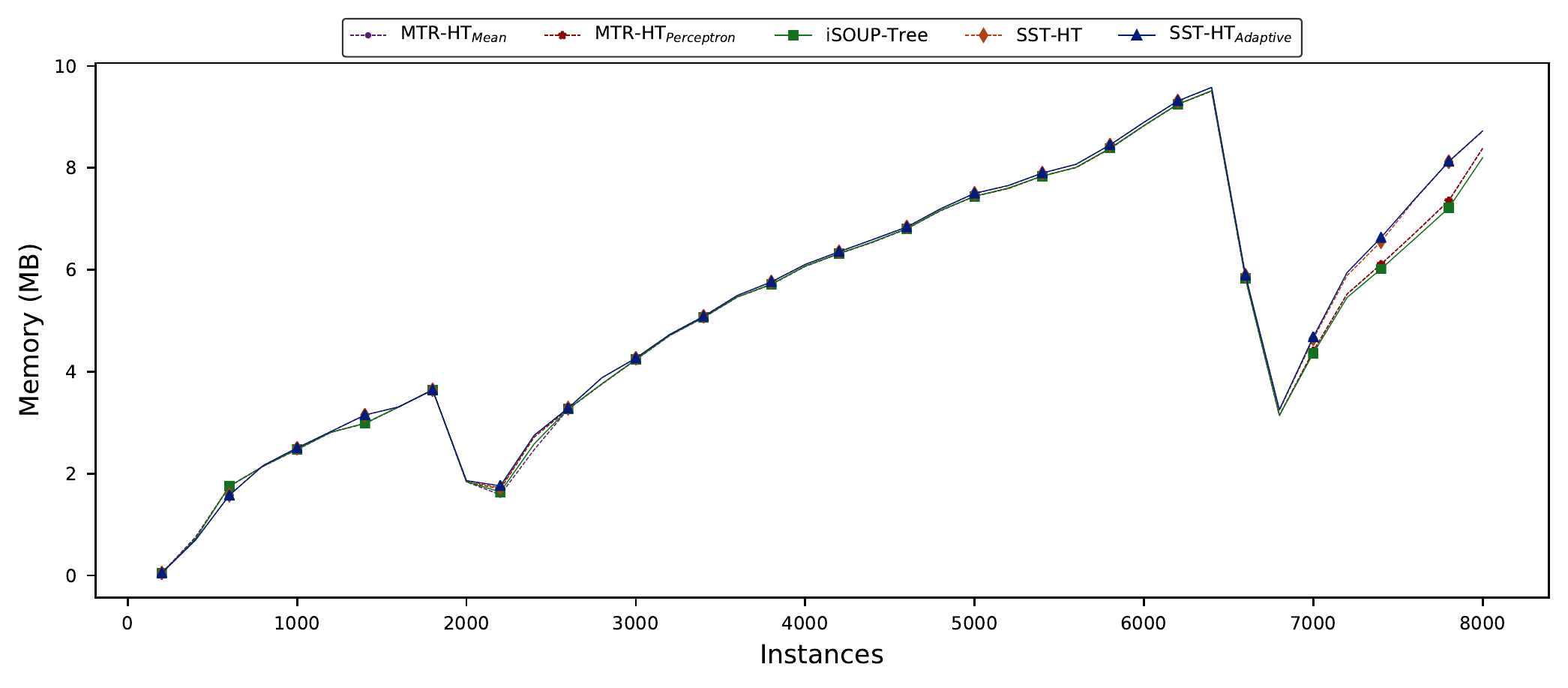}}
    \subfloat[FriedD]{\includegraphics[width=0.5\textwidth]{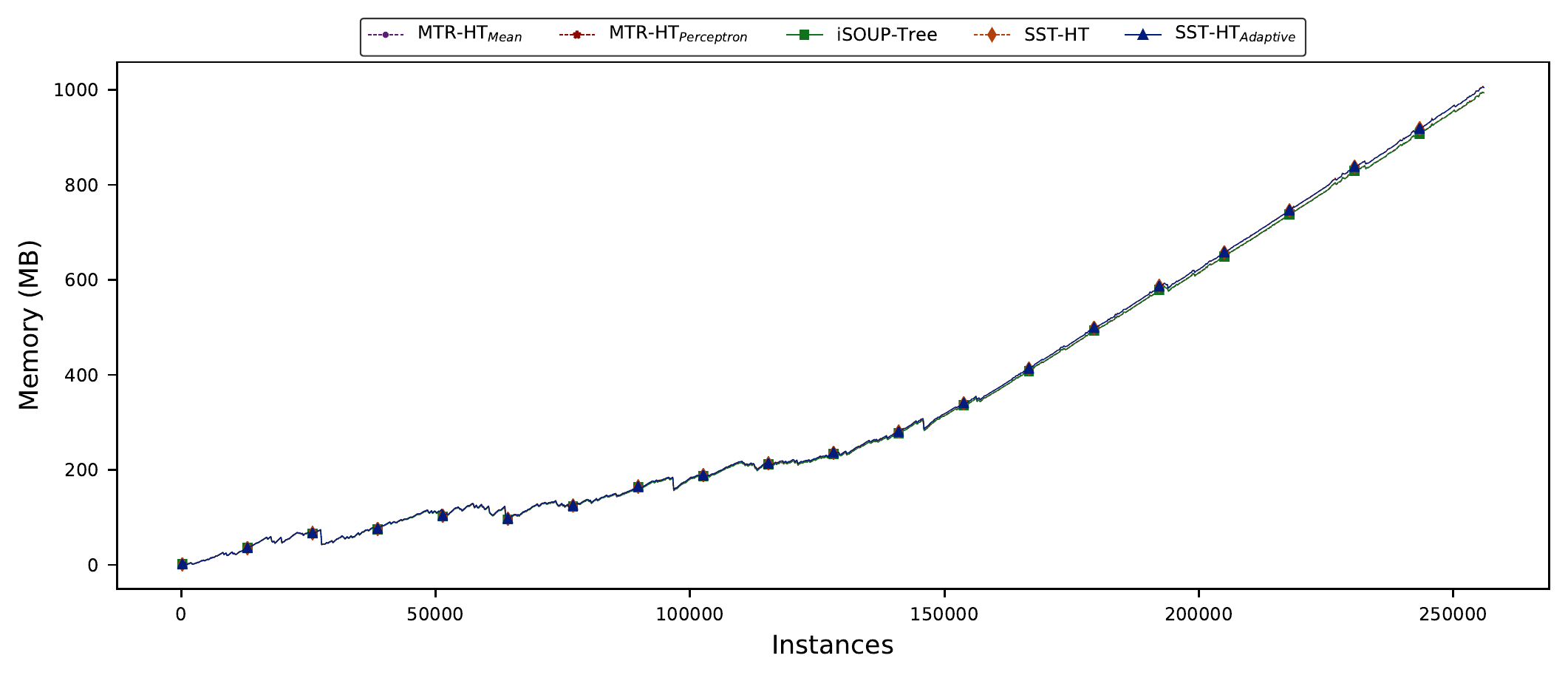}}

    \centering
    \subfloat[FriedAsyncD]{\includegraphics[width=0.5\textwidth]{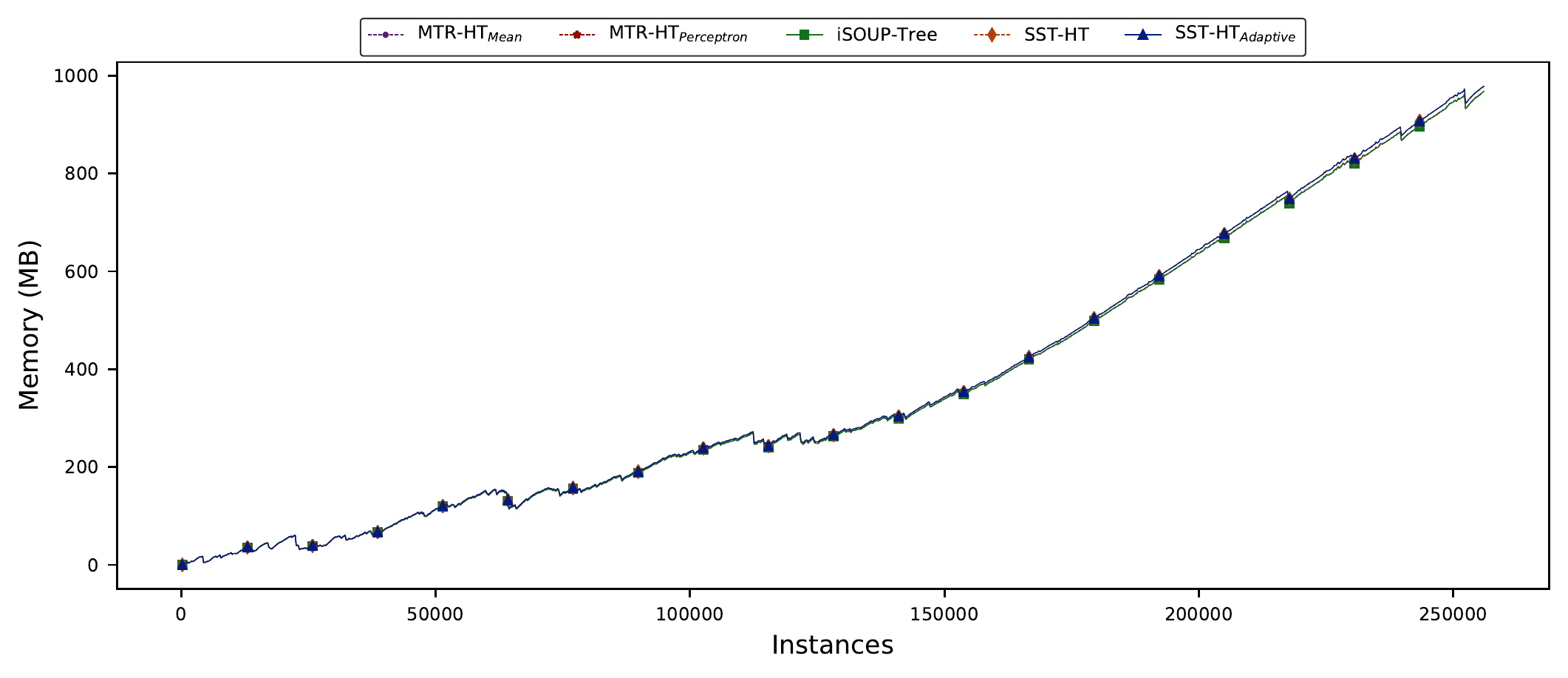}}
    \subfloat[MV]{\includegraphics[width=0.5\textwidth]{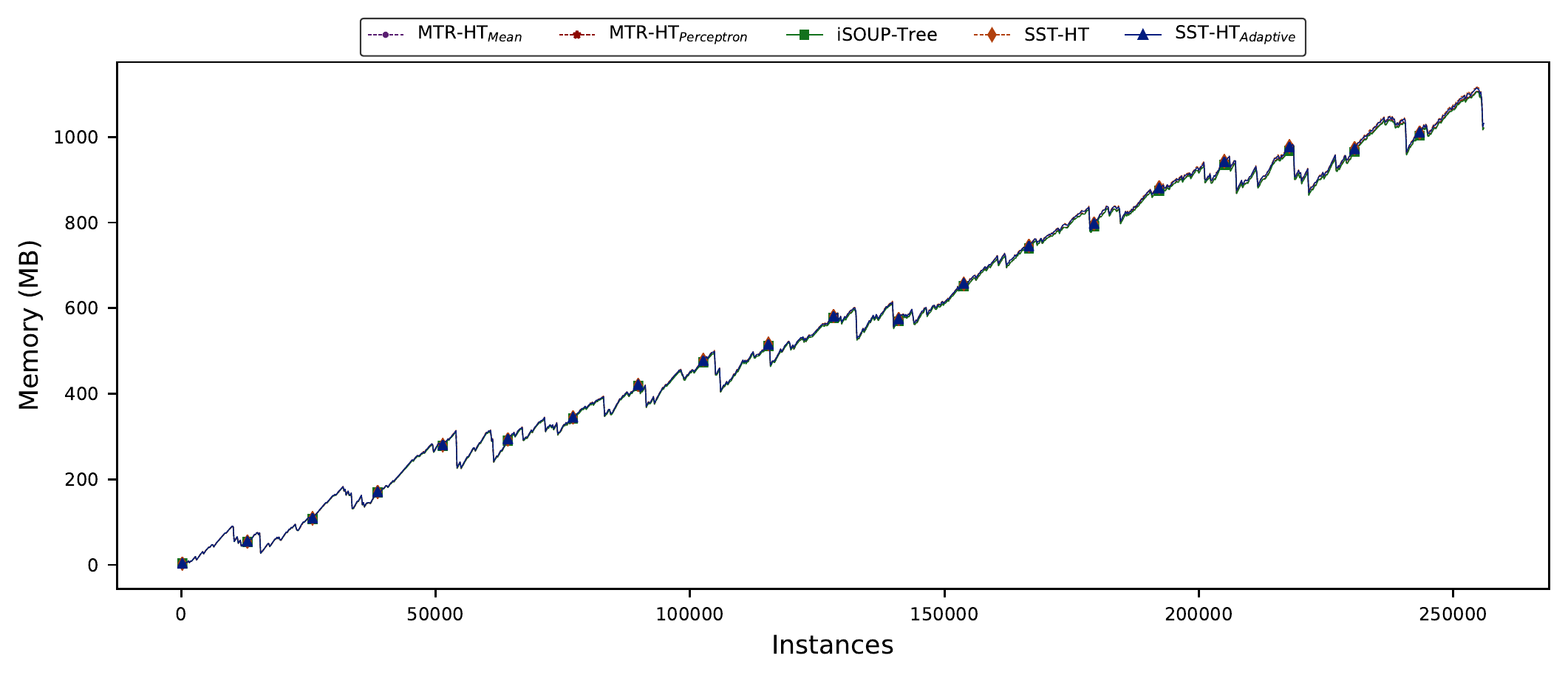}}
    
    \subfloat[NPSDecay]{\includegraphics[width=0.5\textwidth]{line_nps_decay_model_size_M0}}
    \subfloat[RF1]{\includegraphics[width=0.5\textwidth]{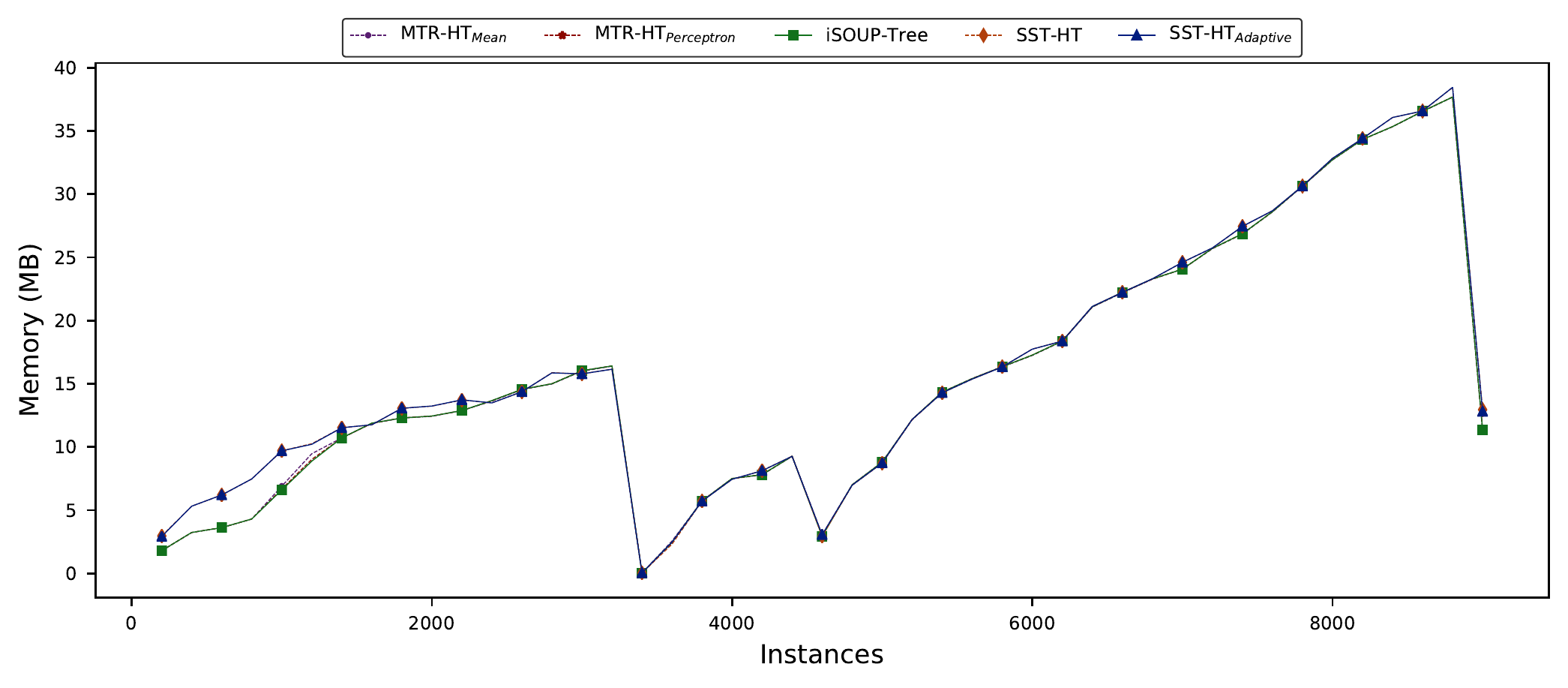}}
    \caption{Time varying model size for all the evaluated datasets}
\end{figure}

\begin{figure}[!htbp]\ContinuedFloat
    \centering
    \subfloat[RF2]{\includegraphics[width=0.5\textwidth]{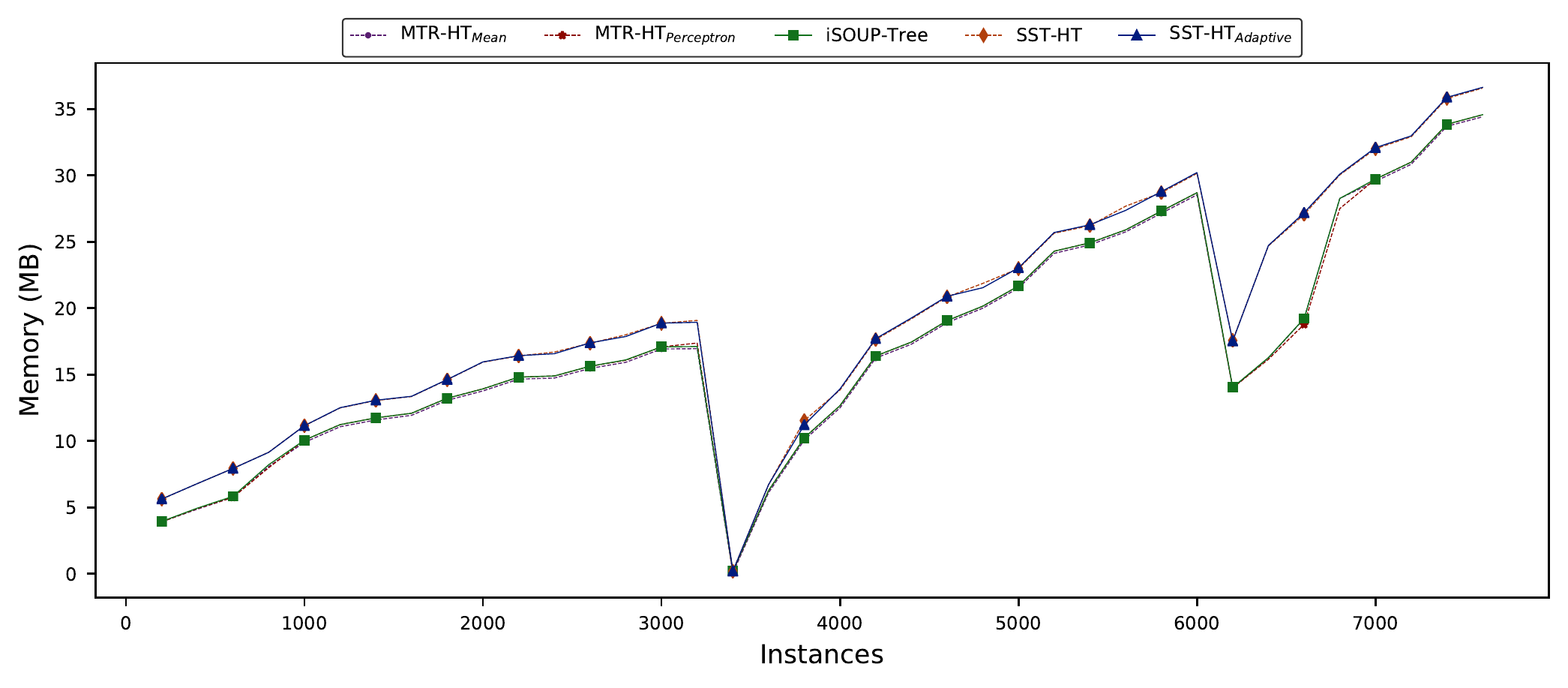}}
    \subfloat[SCFP]{\includegraphics[width=0.5\textwidth]{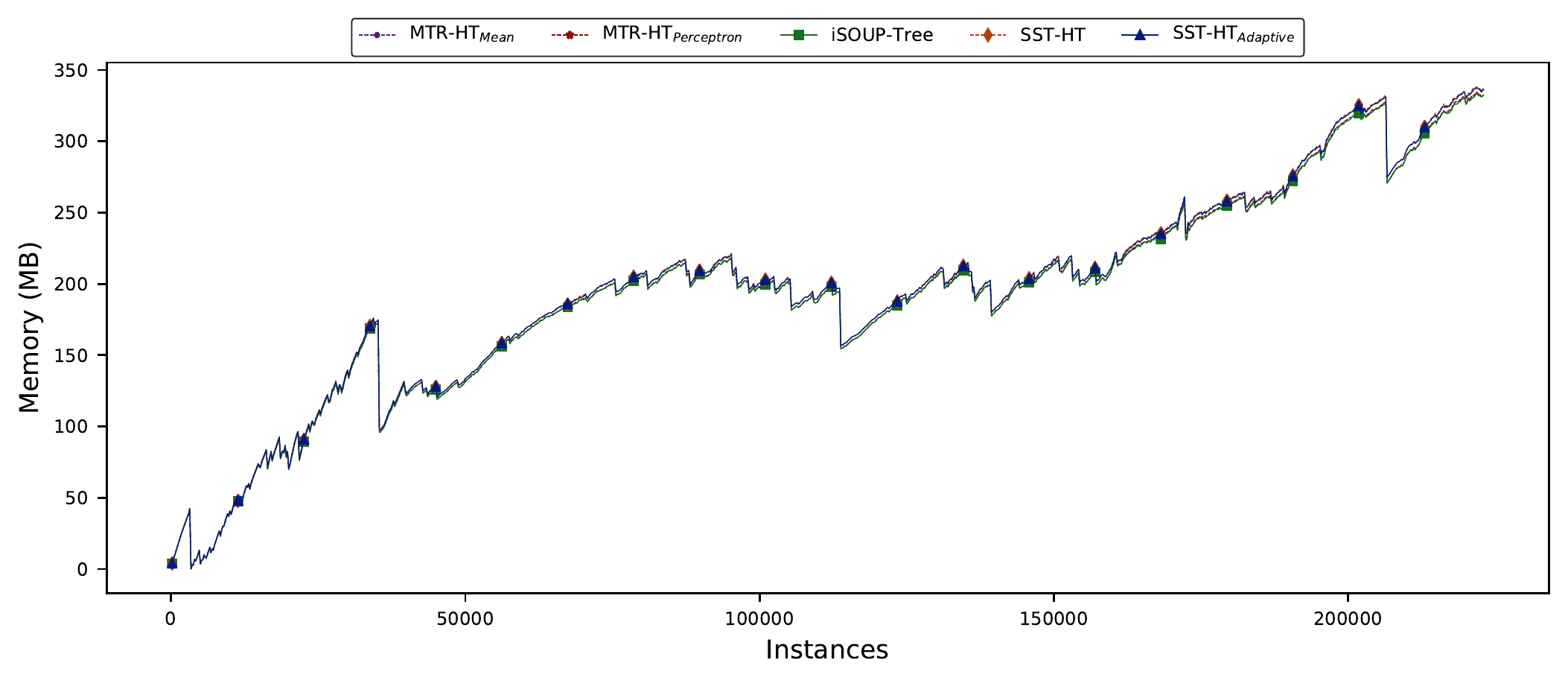}}

    \subfloat[SCM1d]{\includegraphics[width=0.5\textwidth]{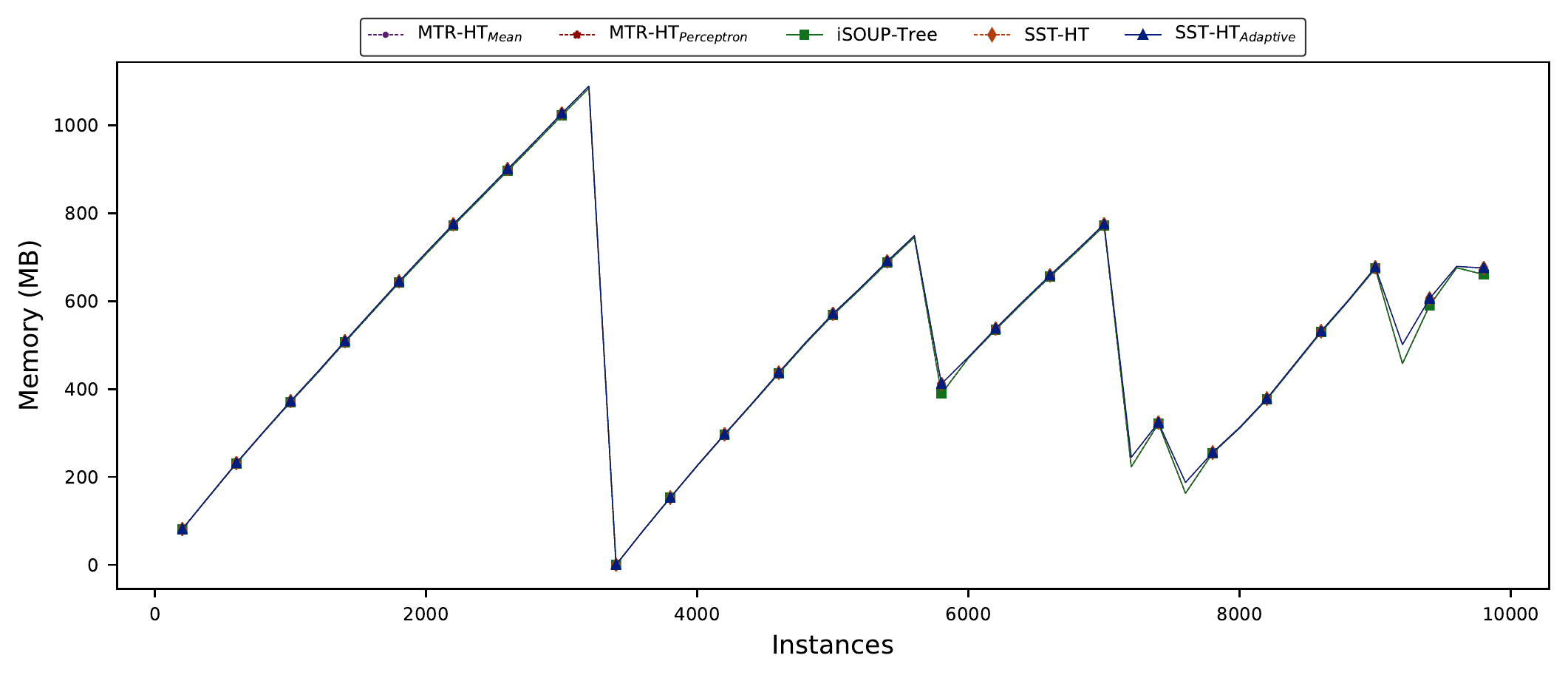}}
    \subfloat[SCM20d]{\includegraphics[width=0.5\textwidth]{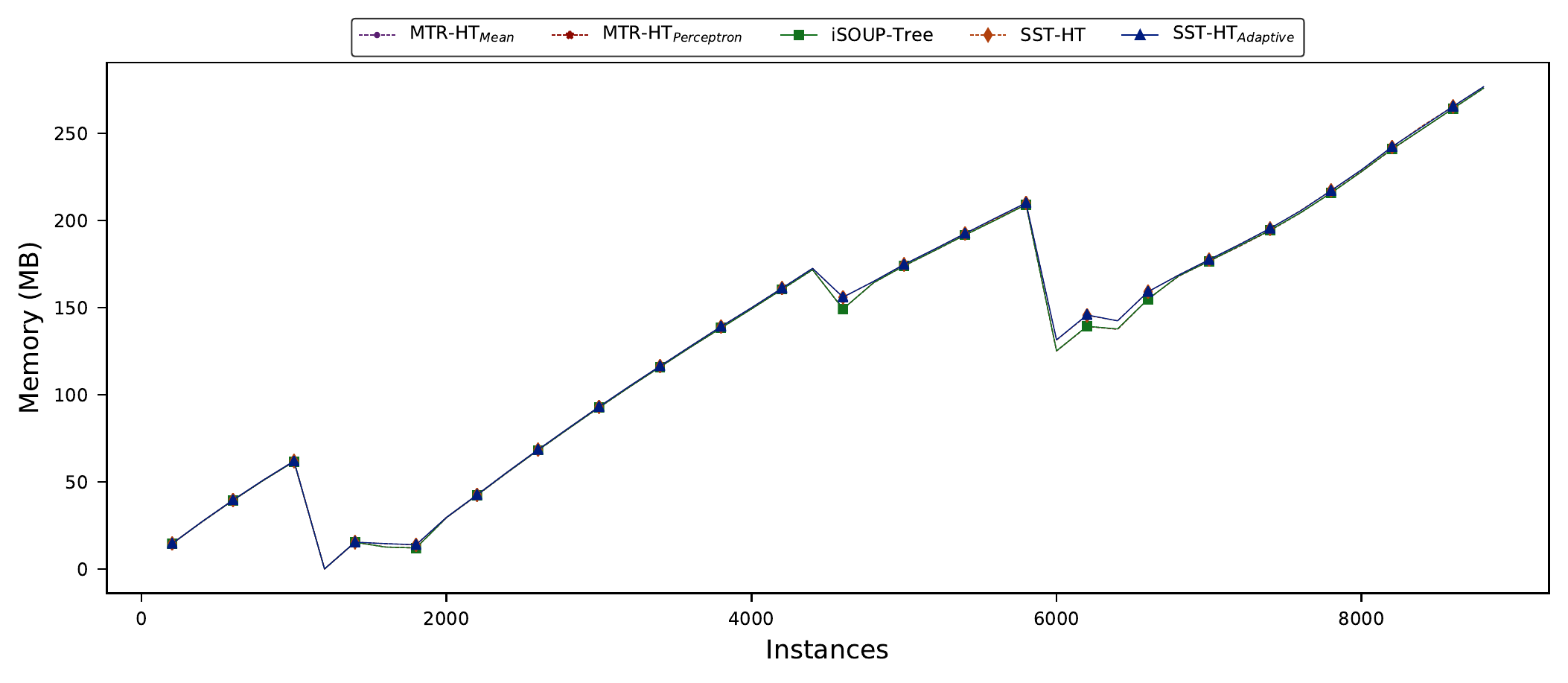}}
    
    \subfloat[Sulfur]{\includegraphics[width=0.5\textwidth]{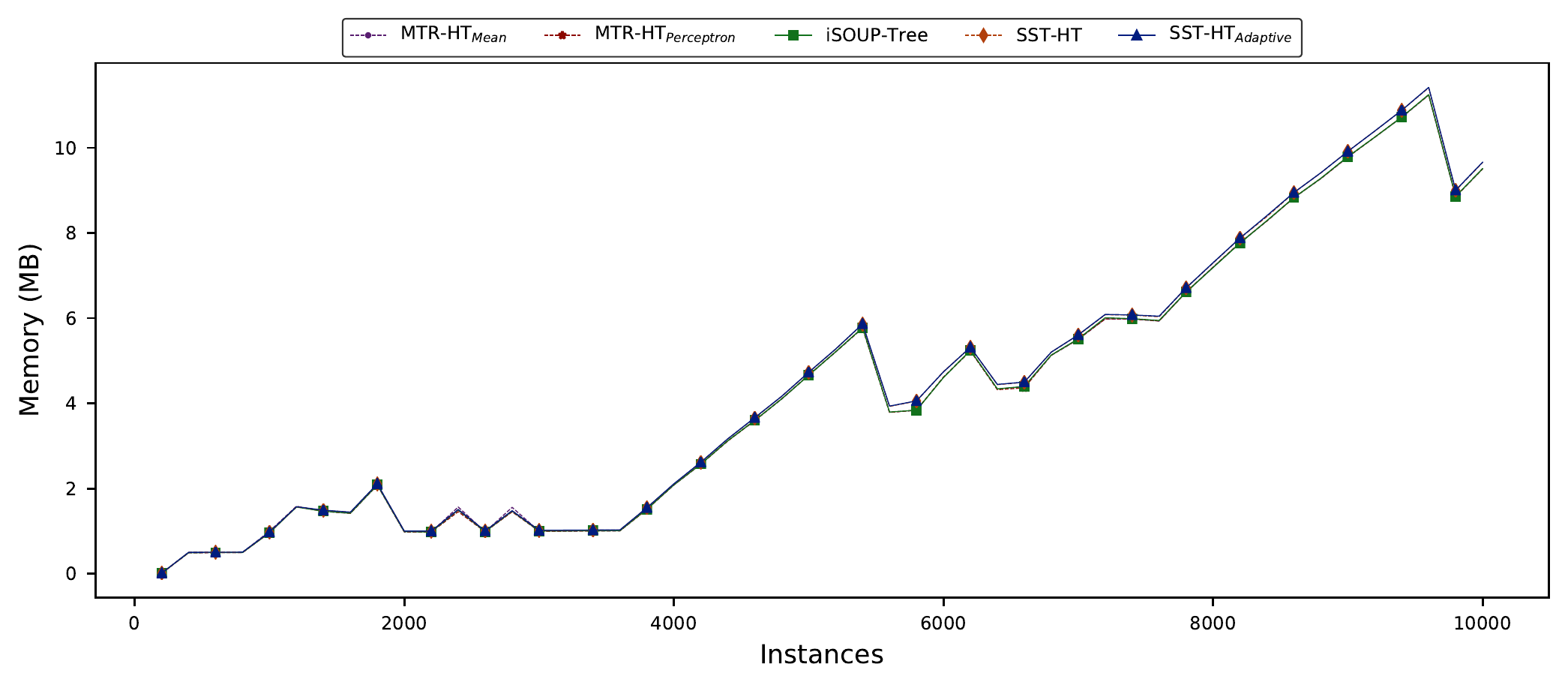}}
    \subfloat[Wine]{\includegraphics[width=0.5\textwidth]{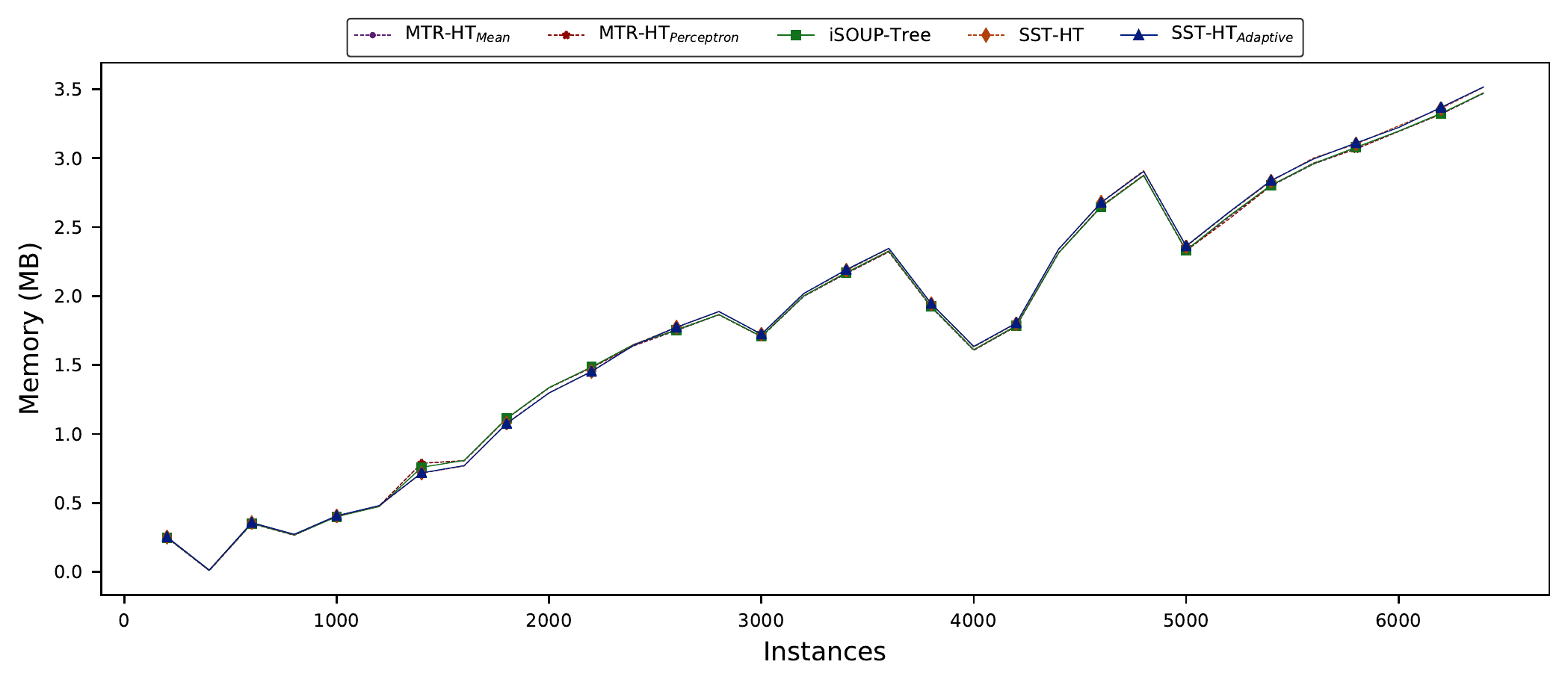}}
    
    \caption{Time varying model size for all the evaluated datasets (continuation)}
    \label{fig_size_lines}
\end{figure}

\end{appendices}

\end{document}